\documentclass[10pt,twocolumn,letterpaper]{article}

\usepackage[pagenumbers]{iccv} 

\definecolor{pinkpurple}{RGB}{219,112,147}

\newcommand{\myparagraph}[1]{\noindent\textbf{#1.}}
\newcommand{\methodname}{MAPLE\xspace}

\newcommand{\secondbest}[1]{{\bf{\textcolor[rgb]{0.145,0.341,0.513}{#1}}}}
\newcommand{\best}[1]{{\bf{\textcolor[rgb]{0.184,0.678,0.149}{#1}}}}

\newcommand{\blsecondbest}[1]{{\bf{\textcolor[rgb]{0.145,0.341,0.513}{#1}}}}
\newcommand{\blbest}[1]{{\bf{\textcolor[rgb]{0.184,0.678,0.149}{#1}}}}

\newcounter{oldtocdepth}

\newcommand{\hidefromtoc}{%
  \setcounter{oldtocdepth}{\value{tocdepth}}%
  \addtocontents{toc}{\protect\setcounter{tocdepth}{-10}}%
}

\newcommand{\unhidefromtoc}{%
  \addtocontents{toc}{\protect\setcounter{tocdepth}{\value{oldtocdepth}}}%
}
\usepackage{stfloats}
\usepackage{bbding}

\newcommand{\noop}[1]{}

\usepackage{tocloft}
\usepackage{amsmath}

\definecolor{iccvblue}{rgb}{0.21,0.49,0.74}
\usepackage[pagebackref,breaklinks,colorlinks,allcolors=iccvblue]{hyperref}

\usepackage{amssymb}
\usepackage{pifont}
\newcommand{\xmark}{\ding{55}}%
\usepackage{arydshln}

\usepackage{amsmath}
\usepackage{amssymb}
\usepackage{multirow}

\def\paperID{0000} 
\def\confName{}
\def\confYear{}

\title{\raisebox{-0.7mm}{\includegraphics[width=0.6cm]{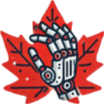}\hspace{0.9mm}}MAPLE: Encoding Dexterous Robotic\\ \underline{Ma}nipulation \underline{P}riors \underline{L}earned From \underline{E}gocentric Videos
\vspace*{-0.5cm}}

\author{
Alexey Gavryushin\textsuperscript{1 \Envelope}
\quad
Xi Wang\textsuperscript{1}
\quad
Robert J.~S.~Malate\textsuperscript{1,2}
\quad
Chenyu Yang\textsuperscript{1}\\
\quad
Davide Liconti\textsuperscript{1}
\quad 
Ren\'{e} Zurbr\"ugg\textsuperscript{1}
\quad 
Robert K.~Katzschmann\textsuperscript{1,2}
\quad 
Marc Pollefeys\textsuperscript{1,3}
\\[8pt]
\textsuperscript{1}ETH Z\"urich, R\"amistrasse 101, 8092 Z\"urich, Switzerland\\
\textsuperscript{2}Mimic Robotics, Andreasstrasse 5, 8050 Z\"urich, Switzerland\\
\textsuperscript{3}Microsoft Research, Seestrasse 356, 8038 Z\"urich, Switzerland
}

\begin{document}

\twocolumn[{%
  \renewcommand\twocolumn[1][]{#1}%
 \maketitle
  \vspace*{-1.2cm}
   \begin{center}
    \centerline{ 
        \captionsetup{type=figure}
        \includegraphics[width=0.95\linewidth]{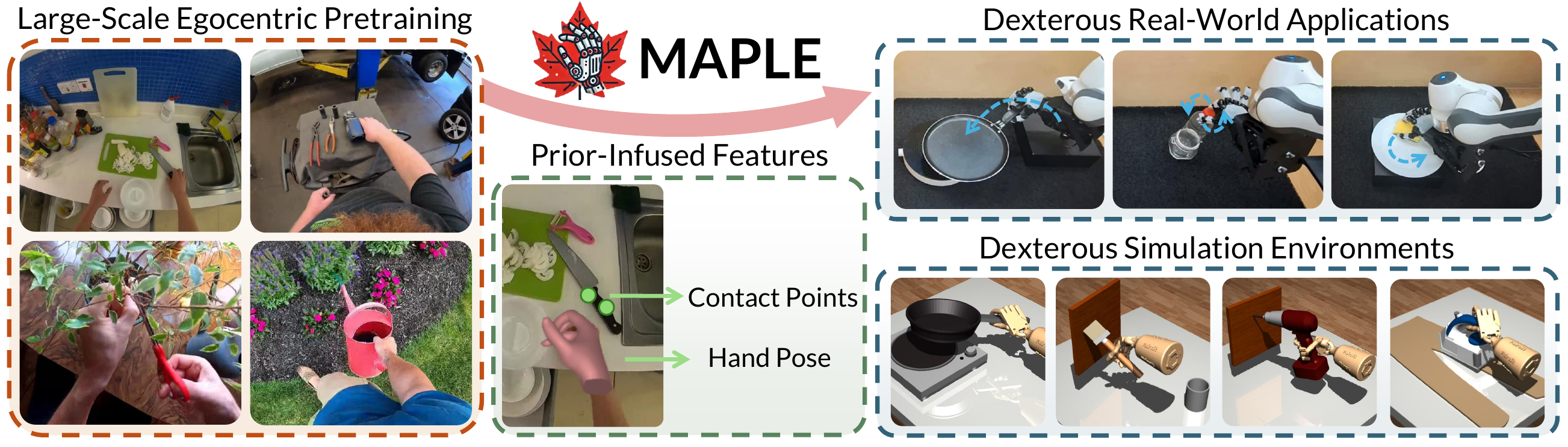}

    }
    
 \vspace*{-2.6em}

 
  \end{center}  
  \begin{center}
\captionsetup{hypcap=false}
\captionof{figure}{\textbf{Encoding Dexterous Robotic Manipulation Priors Learned From Egocentric Videos.} 
We present \methodname, a framework that learns dexterous manipulation priors from egocentric videos and produces features well-suited for downstream dexterous robotic manipulation tasks. 
Experiments in both simulation and real-world settings demonstrate that \methodname enables efficient policy learning and improves generalization across various tasks.  
}
\label{fig:teaser}
  \vspace*{-0.2cm}
\end{center}%
 }]

\maketitle
\hidefromtoc

\def\thefootnote{\Envelope}\footnotetext[1]{\hspace{0.1cm}Correspondence to \texttt{agavryushin@ethz.ch}}
\begin{abstract}

Large-scale egocentric video datasets capture diverse human activities across a wide range of scenarios, offering rich and detailed insights into how humans interact with objects, especially those that require fine-grained dexterous control. 
Such complex, dexterous skills with precise controls are also crucial for many robotic manipulation tasks, which are often insufficiently addressed by traditional data-driven approaches to robotic manipulation. 
To address this gap, we leverage manipulation priors learned from large-scale egocentric videos to improve policy learning for dexterous robotic manipulation tasks. 
We present \methodname, a novel method for dexterous robotic manipulation that exploits rich manipulation priors to enable efficient policy learning and better performance on diverse, complex manipulation tasks. 
Specifically, we predict hand-object contact points and detailed hand poses at the moment of contact and use the learned features to train policies for downstream manipulation tasks. 
Experimental results demonstrate the effectiveness of \methodname across existing simulation benchmarks, as well as a newly designed set of challenging simulation tasks, which require fine-grained object control and complex dexterous skills. Our newly designed simulation environments address the shortage of dexterous manipulation benchmarks in the literature. 
The benefits of \methodname are further highlighted in real-world experiments using a dexterous robotic hand, while simultaneous evaluation across both simulation and real-world experiments has often been underexplored in prior work. 
Our code, trained model, and the new manipulation benchmark suite will be made publicly available on \url{https://algvr.com/maple/}. 
\end{abstract}    
\vspace*{-0.3cm}

\section{Introduction}
\label{sec:intro}

With robotic systems expanding into diverse applications, dexterous manipulation emerges as a critical capability to interact 
 autonomously in dynamic, real-world environments. 
Considering a household robot tasked with preparing a simple meal, 
the robot must interact with various kitchen objects, such as selecting a pan 
and placing it on the stove. 
In such scenarios, the robot encounters a wide variety of everyday objects, many of which require complex dexterous manipulation skills. 
What is, then, the best source of information for teaching robots these skills? 

Large-scale egocentric datasets \cite{Ego4D,EK100, wang2023holoassist, grauman2024ego, li2024egogen} can offer a compelling answer, as they provide a scalable data source and capture rich, natural examples of how humans interact with objects. 
This makes them well-suited for learning visuo-motor manipulation priors directly from real-world natural human-object interactions. 
Several recent studies~\cite{R3M,VC1, singh2024hand, HRP} have explored this direction by pretraining visual representations on large-scale human egocentric videos. 
However, these approaches generally lack the focus on dexterous manipulation and often struggle to generalize, particularly in human-centric environments where fine-grained control is essential~\cite{radosavovic2023robot, burns2023makes, dasari2023unbiased, DAPG}. 

Why, then, is learning generalizable dexterous manipulation priors still so challenging? 
While general-purpose representations pre-trained on large datasets capture rich scene-level details~\cite{CLIP,DINO, DINOv3}, they lack a specific focus on manipulation-relevant cues, making them suboptimal. 
Other self-supervised learning approaches (e.g.~R3M~\cite{R3M}, trained with contrastive losses) struggle to effectively extract dexterous manipulation information from egocentric videos, especially due to the complex and cluttered nature of real-world scenarios depicted in these videos. 
This raises a key question: what specific information should a visual encoder designed for object manipulation learn to extract?
We argue that this encoder should be trained to reason about \textit{where} to interact with an object and \textit{how} to do so. More concretely, we hypothesize that \textit{object contact points} and \textit{grasping hand poses} provide strong priors for downstream dexterous manipulation tasks.
To this end, we introduce \methodname, a novel method that learns \emph{dexterous manipulation priors} from egocentric videos for robotic manipulation.  
We develop a pipeline to automatically extract supervision signals from egocentric videos, enabling large-scale training without manual annotation.
Given a single image of the target object, \methodname learns to predict hand-object contact points and the 3D hand pose at the moment of contact. 
The learned visual representations are then used as input to visuo-motor policies that are trained to control dexterous robotic hands. 

Unlike previous work that predicts coarse affordance~\cite{HRP, kannan2023deft}, we demonstrate that \methodname learns transferable features by \emph{predicting fine-grained hand-object interaction cues}, enabling diverse robotic embodiments to perform a wide range of manipulation tasks that require complex dexterous skills. 
We conduct experiments on established simulated benchmarks specifically designed to evaluate dexterous manipulation skills. 
Additionally, we introduce a new suite of challenging manipulation tasks in simulation that require fine-grained object control and advanced dexterous coordination.  
We first show that the features learned with \methodname transfer beyond policy learning by evaluating them on a contact point prediction task. 
Then, we use them as input for learning visuomotor policies, performing a comprehensive simulation evaluation across four existing and four newly proposed dexterous manipulation tasks, comparing \methodname against general-purpose visual encoders and state-of-the-art methods.
The benefits of \methodname are further highlighted in real-world experiments using a dexterous robotic hand. 
Such an evaluation setting using both simulation and real-world experiments has been largely underexplored in prior work. 
We will publicly release our codebase and data 
to support future research.  

In summary, our contributions are:

\begin{itemize}
    \item We propose \methodname, a visual encoder for dexterous manipulation tasks, learned through a novel pretraining strategy that predicts interaction cues for object manipulation.
    \item We propose a new set of custom-designed dexterous manipulation tasks with new  assets in simulation that require more fine-grained object control.
    \item We demonstrate that \methodname significantly improves performance on downstream dexterous robotic manipulation tasks in both simulation environments and real-world settings as well as an egocentric contact point prediction task, highlighting its improved generalization ability.
\end{itemize}
\section{Related Work}
\label{sec:related}
Our work focuses on dexterous manipulation in robotics. In the following, we review the most relevant research areas.

\myparagraph{Dexterous Manipulation} 
Dexterous manipulation describes the manipulation of objects with multi-fingered hands. Due to the high dimensional action space and complex physical interactions, it is one of the most challenging tasks in robotics.
Traditional approaches in dexterous manipulation typically rely on analytical grasp planning methods that use models~\cite{DGrasp} or optimization~\cite{turpin2023fast,wang2023dexgraspnet}. 
These grasp proposals are then executed in an open loop, feed-forward fashion using a suitable task-space controller. 
While this modular approach allows for more flexibility, it faces challenges when dealing with partially observed environments or when adaptability and the tight coupling of vision and control are necessary. 
Therefore, more recent work has investigated the use of data-driven end-to-end techniques such as reinforcement learning to learn manipulation policies~\cite{zhang2024artigrasp,zhang2024graspxl, qi2023hand}.
Reinforcement learning provides a powerful tool to learn robust dexterous manipulation policies, but its performance heavily relies on simulation accuracy as well as proper reward formulations, which might be challenging to formulate. 
This has sparked research interest that has led to the advent of imitation learning \cite{chi2023diffusion, zhao2023learning, kareer2024egomimic, lin2024learning, nava2025mimiconescalablemodelrecipe, bauer2025latentactiondiffusioncrossembodiment} 
which directly leverages human demonstrations as a supervision signal. 
While most of these methods are able to learn meaningful manipulation policies, their generalization across different tasks is still limited, and they often rely on a large amount of task-specific trajectories in order to be used for new tasks. 

\myparagraph{Learning From Egocentric Videos} 
Large-scale human videos have been widely used to extract embeddings or motion priors for robotic manipulation.
Existing methods can largely by divided into implicit and explicit methods.

\textit{Implicit methods} \cite{R3M, Mandikal22DexVIP, VC1} use large-scale contrastive learning to derive meaningful features from egocentric interaction data. Some works, such as \cite{VC1, xiao2022masked}, leverage large-scale video datasets to train masked autoencoders (MAE). Methods like \cite{R3M} integrate language embeddings by applying a contrastive loss across image and language modalities. However, due to their implicit nature, these methods cannot be directly used for robotic manipulation. Instead, their pre-trained (frozen) embeddings serve as inputs for reinforcement learning \cite{xiao2022masked} or imitation learning \cite{R3M} algorithms.

Conversely, \textit{explicit methods} \cite{nagarajan2019grounded, VRB, Bharadhwaj23ZeroShotRobotManip, zeng2024mpi, Mandikal21GRAFF} aim to directly learn interaction affordances from videos by using human interaction as an explicit supervision signal. These approaches typically predict interaction trajectories \cite{nagarajan2019grounded}, contact heatmaps \cite{Mandikal21GRAFF, VRB, zeng2024mpi}, or future contact points and wrist positions \cite{HRP}. While their output could be directly used within a task-specific controller, most methods use the final predictions or intermediate embeddings in policy learning 
due to the complexity of the manipulation task.
Although explicit supervision helps learn a better-suited latent space, most existing approaches focus on high-level interactions and are primarily designed for simple two-finger manipulation. Furthermore, they often depend on human-annotated data, limiting scalability. In contrast, our method does not rely on any human-annotated data and adopts a more fine-grained formulation to capture detailed manipulation dynamics, well-suitable for dexterous manipulation. 

\myparagraph{Evaluation in Robotic Simulators}
We categorized simulation benchmarks by the type of end-effector used. 
Benchmarks like MetaWorld \cite{MetaWorld}, Franka Kitchen \cite{D4RL,gupta2019relay}, and Habitat 2.0 \cite{Habitat2} focus on parallel jaw grippers, defining tasks such as object manipulation, cooking-related activities, and home environment rearrangement tasks. The DeepMind Control Suite \cite{DMCS} includes basic object manipulation tasks in abstract environments.
For three-finger grippers, TriFinger \cite{TriFinger} provides a benchmark for object manipulation, particularly evaluating toy cube tasks in both real and simulated settings.
Finally, dexterous hand benchmarks offer more complex manipulation challenges. DAPG \cite{DAPG} introduces four environments for a simulated Adroit \cite{AdroitHand} hand, while DexMV \cite{DexMV} includes tasks like pouring, object relocation, and target-based movement, requiring more fine-grained control.
A key limitation across these benchmarks is their overwhelming reliance on low-DoF grippers, leaving researchers few options to evaluate policies for fine-grained manipulation. This motivates us to design a more comprehensive dexterous manipulation benchmark.

\myparagraph{Evaluation in Real World}
In the robotics community, imitation learning is becoming a standard approach to performing dexterous manipulation tasks \cite{chi2023diffusion, zhao2023learning, dasari2024DiTpolicy}, even driving the development of foundation model-style generalist robot policies that can be fine-tuned with target data from small samples \cite{octomodelteam2024Octo, kim2024OpenVLA}. This is because doing so enables learning from task demonstrations without the need for simulation environments or task-specific rewards. However, these models have been primarily trained and deployed on low-DoF grippers. Furthermore, for dexterous manipulators, there are no standard datasets due to the high variation between designs. This served as motivation for our benchmark for dexterous manipulation tasks in the real world. 

\section{\methodname}
\label{sec:method}

\begin{figure*}
    \centering
\includegraphics[width=0.78\linewidth]{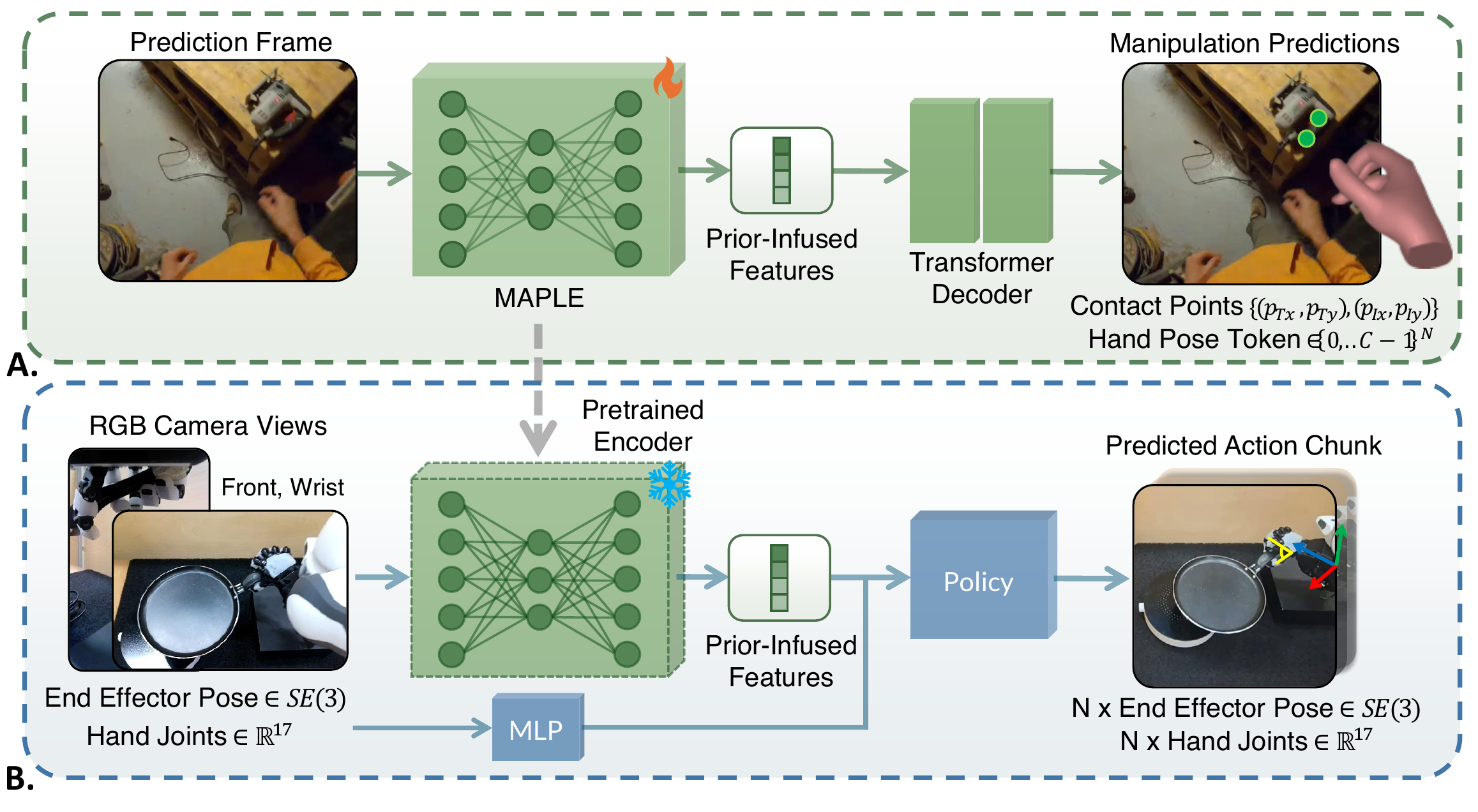}
    \vspace{-0.2cm}
    \caption{\textbf{Overview of \methodname.} 
    Given a single input frame, the encoder is trained to reason about hand-object interactions, specifically predicting contact points and grasping hand poses. 
    This training infuses a \textbf{manipulation prior} into the learned feature representation, making it well-suited for downstream robotic manipulation. 
    Features extracted from the frozen visual encoder, combined with robotic hand positions, are fed into a Transformer-based diffusion policy network to predict dexterous hand action sequences. 
    }
    \label{fig:method}
    \vspace{-0.4cm}
\end{figure*}
We aim to learn object manipulation skills from egocentric videos that capture natural human interactions with diverse everyday objects. To this end, we introduce \methodname, a pipeline for learning manipulation priors. The pipeline addresses two key challenges: modeling manipulation priors (Sec.~\ref{sec:formulation}) by extracting supervision from videos (Sec.~\ref{sec:extract}) and learning these priors for downstream dexterous control (Sec.~\ref{sec:priorlearning}). Sec.~\ref{sec:training} details the training strategy.
See \autoref{fig:method} for our method overview. 

Our fully automated extraction pipeline produces 
approximately 82,100 training samples from the large-scale Ego4D \cite{Ego4D} dataset, which we then use to train our visual encoder model for dexterous tasks.

\begin{figure}[!ht] 
    \centering
    \includegraphics[width=0.6\linewidth]{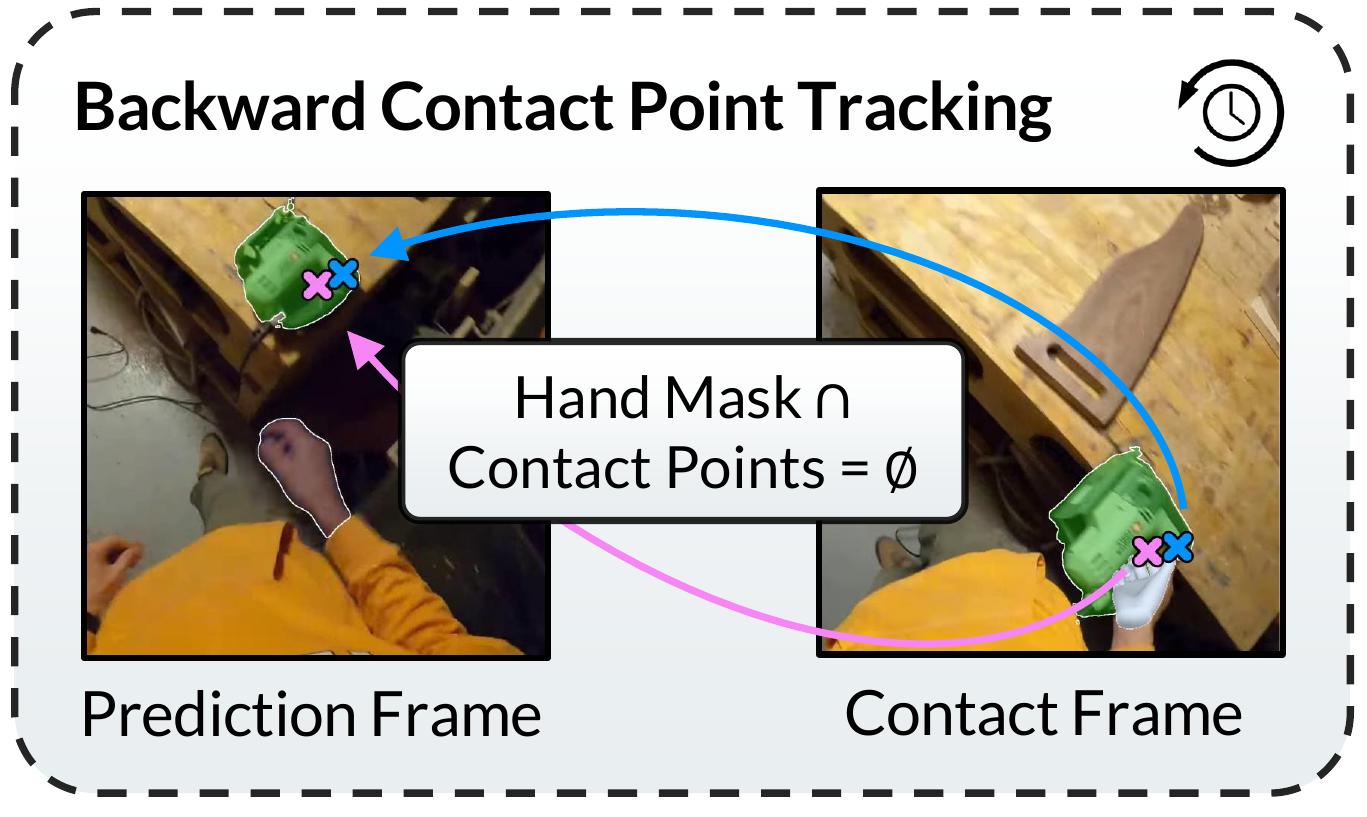}  
        \vspace{-0.2cm}
    \caption{
    \textbf{Manipulation Prior Modeling.} 
    We learn manipulation priors by predicting future contact points and hand poses from an input frame. Training data are extracted by identifying a \textit{contact frame} $f_c$ and a preceding \textit{prediction frame} $f_p$, which is used as input. 
    Contact points and hand poses are extracted from the contact frame, and a point tracker is used to back-project the contact locations onto the prediction frame. 
    } 
    \vspace{-0.6cm}
\label{fig:data_extraction}
\end{figure}

\subsection{Manipulation Prior Formulation}
\label{sec:formulation}
To effectively model manipulation priors, we focus on the moment of contact, which captures the transition from perception to physical interaction. At contact, the hand pose and contact points between hand and objects are jointly constrained, exposing the action's affordance. This physically grounded state provides stable cues that are directly transferable to downstream manipulation learning. 

Therefore, we consider both contact points and hand poses as the manipulation prior and formulate the following learning task. 
Given an input frame, our goal is to predict future contact points and hand poses upon contact. 
Training data are extracted by identifying a \textit{contact frame} $f_c$ and a \textit{prediction frame} $f_p$, which is used as input.
We define $f_c$ as the moment when the hand first contacts the object and $f_p$ as an earlier frame, where the hand is sufficiently distant from the object. 
From $f_c$, we extract the interacting hand pose vector $\mathcal{H}_{c}$ and corresponding contact points $pt_c$, which are then back-projected to $f_p$ via point tracking to obtain future contact points 
$pt_p$ in the prediction frame. 
\autoref{fig:data_extraction} provides a visualization of this formulation. 

\subsection{Manipulation Supervision Extraction}
\label{sec:extract}

\myparagraph{Identifying the Contact Frames}
To locate moments of contact, 
we process all frames using the off-the-shelf hand-object contact segmentation model VISOR-HOS~\cite{EKVisor} and consider the first contact frame in a series as $f_c$.

\myparagraph{Extracting Contact Labels from the Contact Frame}
We define the \textit{contact points} $pt_c$ as the object surface points where the fingertips of the thumb and index finger first touch the object. 
For each contact frame, we use the pre-trained hand pose estimation model HaMeR \cite{HaMeR} to extract the MANO~\cite{MANO} pose vector $\mathcal{H}_c \in \mathbb{R}^{21\times 3}$ of the interacting hand. 
$\mathcal{H}_c$ is forwarded through a MANO layer to obtain 2D fingertip keypoints. 
We then project the thumb and index fingertip keypoints onto the object mask to obtain $pt_c$. 
When calculating the projected fingertip positions, we use a slightly smaller object mask from which a portion of the boundary has been erased by binary erosion \cite{BinaryDilation}.
This also helps when finding the contact points in the prediction frame in the next step. 
Prior work \cite{HRP, VRB} randomly samples contact points from the intersection of the hand bounding boxes and the manipulated object, which is highly error prone, 
such as when points are sampled from the background as the hand is touching an object on its visual boundary. In contrast, we project fingertip locations to the manipulated objects' segmentation masks to extract precise contact information.
\myparagraph{Extracting the Prediction Frame and its Contact Labels}
To identify the prediction frame $f_p$, we run a point tracker \cite{SamPT} to project the contact points backward in  time until reaching the first frame where the hand mask no longer intersects the object mask. 
Using smaller object masks for contact point identification helps prevent the point tracker from tracking the fingers of the manipulating hand instead of the object. 
When checking for the intersection, we expand the object mask using binary dilation \cite{BinaryDilation} to ensure a sufficient distance from the object to the hand. This helps make the contact point regression less trivial than simply selecting the object next to the hand.

\myparagraph{Tokenizing Hand Poses}
Learning to predict hand poses associated with grasping a target object is a challenging task, due to the need to conform to anatomical constraints and reason about a complex kinematic tree with a high number of joints. Furthermore, a large subset of the joint configuration space results in hand poses that exhibit self-penetration. We simplify the prediction task by reframing it as a \textit{classification} task and letting the model select the hand pose among a limited list of ``reasonably frequent" (i.e., observed in videos) and anatomically correct choices rather than regressing the correct hand pose by itself. For that purpose, we train a tokenizer on hand poses extracted by HaMeR \cite{HaMeR} from a subset of the Ego4D dataset.
We use the same Memcodes \cite{mama2021nwt}-based tokenization scheme as the human pose tokenizer proposed in 4M \cite{4m}, employing 8 codebooks of size 1024 each to represent a single MANO \cite{MANO} hand pose. 

In Supp.~Mat., we provide visualizations and an analysis of failure cases in the extracted manipulation supervision, the most common ones being false contact positives and samples with prediction frames still exhibiting contact. These residual errors reflect the large number of false positives generated by \cite{EKVisor}, which are largely eliminated by filtering in our pipeline.
Qualitative reconstruction examples of the hand tokenizer can be found in the Supp.~Mat. 

\subsection{Encoding Manipulation Priors}
\label{sec:priorlearning}

We aim to train a visual encoder tailored for dexterous manipulation by reasoning about human hand-object interactions extracted from egocentric videos. 
We use an encoder-decoder structure as a stepping stone in learning useful representations: the encoder processes input prediction frames into embedding vectors, which the decoder then uses to predict contact points and future hand poses. 
The decoder relies entirely on the encoder's embeddings, which have a much lower dimensionality than the input images, to infer its output. This bottleneck design encourages the encoder to produce information tailored specifically for object manipulation-related predictions, rather than learning to produce ``general-purpose" visual embeddings \cite{CLIP,DINO}.  

\myparagraph{Contact Point Prediction}
We employ the cross-entropy loss to learn to predict contact points $pt_p$ in the prediction frame $f_p$. 
To do so, we discretize the image's width $w$ and height $h$ into $B_x$ resp.~\ $B_y$ bins each and let the decoder predict the respective bin indices into which the contact point should fall. 
We set $B_x=100$ and $B_y=100$ in our experiments.
We formulate the contact loss $\mathcal{L}_{ct, pt}$ for one sample and one contact point $pt$ as the following: 
\vspace{-0.2cm}
\begin{equation*}
  \begin{aligned}
    \mathcal{L}_{ct,pt} = CE({\arg\max}_{j\in \{1,...,B_x\} }\ \tilde{p}_{pt,x,j}, \left\lfloor\frac{\hat{x}}{w} \times B_x\right\rfloor ) +\\
CE({\arg\max}_{j\in \{1,...,B_y\} }\ \tilde{p}_{pt, y,j}, \left\lfloor\frac{\hat{y}}{h} \times B_y\right\rfloor ),
  \end{aligned}
\end{equation*}
where $\tilde{p}_{x,\hat{x}} \in \mathbb{R}^{B_x}$, $\tilde{p}_{y,\hat{y}} \in \mathbb{R}^{B_y}$ are the decoder's corresponding logit vectors. 
$CE(\tilde{c}, \hat{c})$ represents the cross-entropy loss for the predicted class $\tilde{c}$ and the ground-truth class $\hat{c}$.  $\hat{x}$ and $\hat{y}$ are the extracted contact point coordinates. We use $pt \in \{0, 1\}$ for the thumb and index contact points.

\myparagraph{Hand Pose Prediction}
During MAPLE's training, we let the decoder predict a probability distribution for each hand pose token in the token sequence and use the highest-scoring token logit to determine the final token from which to reconstruct the hand pose.
Similarly to the contact point prediction, we also use a cross-entropy loss for this.
Specifically, we process a MANO hand pose $\mathcal{H} \in \mathbb{R}^{21\times 3}$ using a tokenization producedure $$\mathcal{T}: \mathbb{R}^{21\times 3} \mapsto \{0, 1, ..., C-1\}^{N},$$ which converts the pose into a token $t_n$, where $1\leq n \leq N$ and $C$ represents the codebook size of the tokenizer. 
Let $\tilde{p}_{n}$ be the decoder's logit predictions for estimating $t_n$. Then the loss formulation $\mathcal{L}_{hand}$ is given by:
\vspace{-0.2cm}
\begin{equation*}
\mathcal{L}_{hand}  = \sum_{in=1}^N CE({\arg\max}_{j\in \{1,...,C\} }\ \tilde{p}_{n,j}, p_n).
\end{equation*}

The overall loss for a single sample is thus calculated as 
$$
    \mathcal{L} = \sum_{pt \in \{0, 1\}} \mathcal{L}_{ct,pt} + \lambda_{hand}  \mathcal{L}_{hand} 
$$

\subsection{Training Strategy}
\label{sec:training}

\myparagraph{Training Data}
\label{sec:datasets}
We run our extraction pipeline on \textit{Ego4D} \cite{Ego4D}, which consists of more than 3,000 hours of unscripted human activities across diverse scenarios (e.g., gardening, exercise, and cooking).
To ensure a fair comparison, we follow the established protocols as in \cite{R3M, HRP} and select the same subset covering approximately 1,590 hours of video for extracting data. Our pipeline produces approximately 82,100 training samples from this subset. 
See the Supp.~Mat.~for more details and visualizations.

\myparagraph{Implementation Details}
Similar to other ViT \cite{Dosovitskiy20ViT}-derived baselines \cite{HRP,VC1,DINO}, we employ a ViT-B/16 as the encoder's architecture. The encoder is initialized with the weights of \cite{DINO} and produces visual features of dimension $d=768$.
To ensure fair comparisons and evaluate the performance of our supervised training procedure, rather than the choice of architecture, we do not opt for a larger or more intricate backbone. 
We use a Transformer \cite{Vaswani17Transformer} decoder model with 2 Transformer layers and 4 attention heads per layer. The encoder and decoder are jointly trained using the AdamW \cite{AdamW}  optimizer with a learning rate of $5\times 10^{-5}$. We use a batch size of 128 when training MAPLE.

\section{Experiments}
\label{sec:exp}

To demonstrate the generality of \methodname, we evaluate on both perception-oriented contact prediction tasks (Sec.~\ref{sec:contact}) and robotic dexterous manipulation tasks, which are tested in simulation (Sec.~\ref{sec:evaluation_setup}) and a real-world setup (Sec.~\ref{sec:evaluation_realworld}). 

\begin{figure*}
    \centering
    \begin{subfigure}[t]{0.12\linewidth}    
     \centering
     \includegraphics[width=1\textwidth]{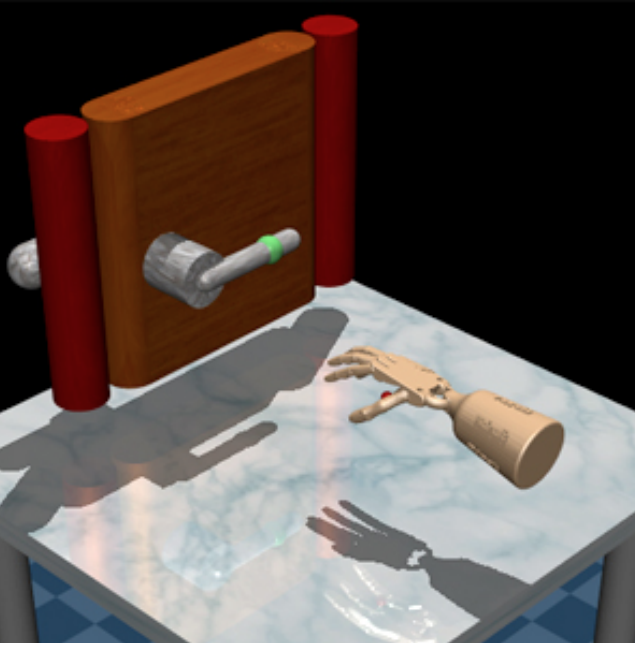} 
    \caption{Door}
    \end{subfigure}   
    \begin{subfigure}[t]{.12\linewidth}    
     \centering
     \includegraphics[width=1\textwidth]{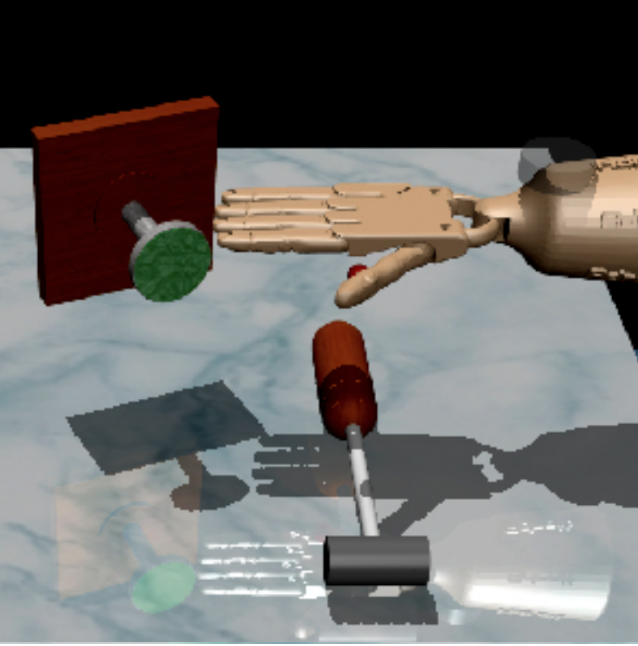} 
    \caption{Hammer}
    \end{subfigure}   
    \begin{subfigure}[t]{.12\linewidth}    
     \centering
     \includegraphics[width=1\textwidth]{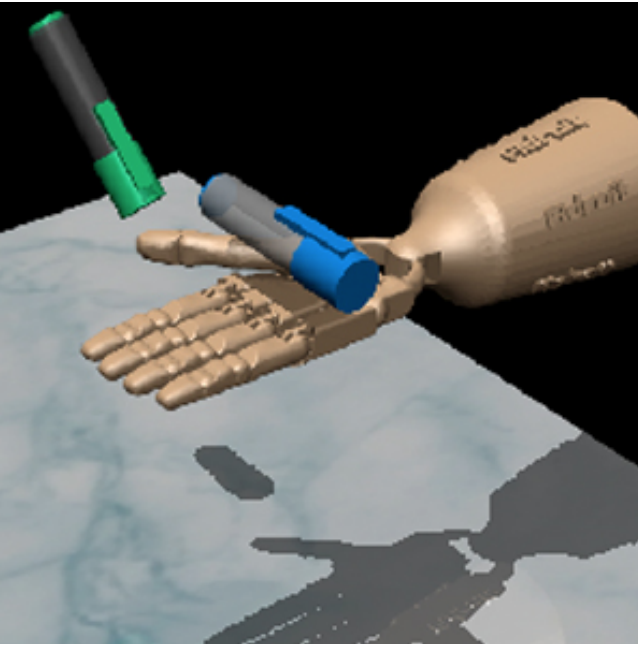} 
    \caption{Pen}
    \end{subfigure}   
    \begin{subfigure}[t]{.12\linewidth}    
     \centering
     \includegraphics[width=1\textwidth]{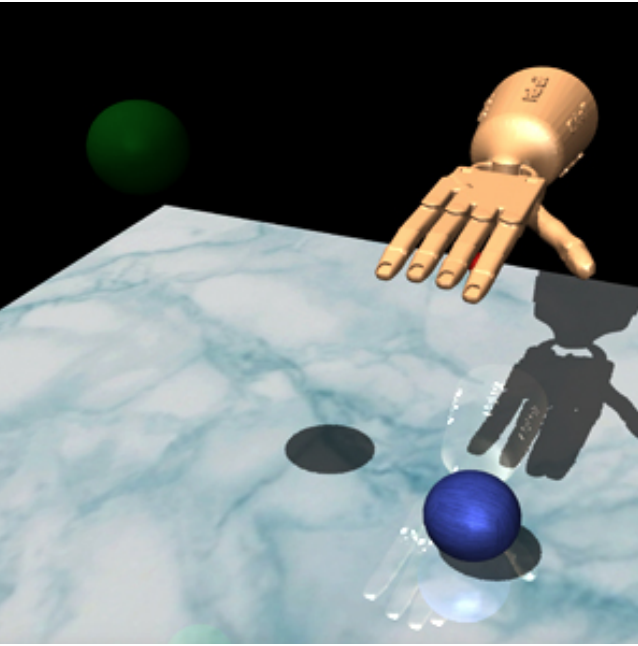} 
    \caption{Relocate}
    \end{subfigure}   
     \begin{subfigure}[t]{.12\linewidth}    
     \centering
     \includegraphics[width=1\textwidth]{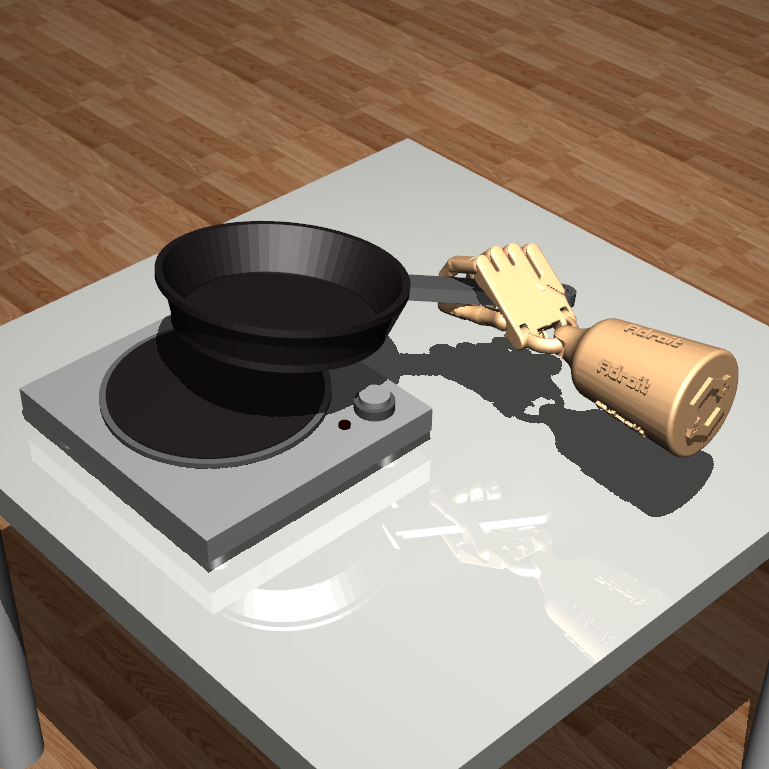} 
    \caption{Pan}
    \end{subfigure}   
    \begin{subfigure}[t]{.12\linewidth}    
     \centering
     \includegraphics[width=1\textwidth]{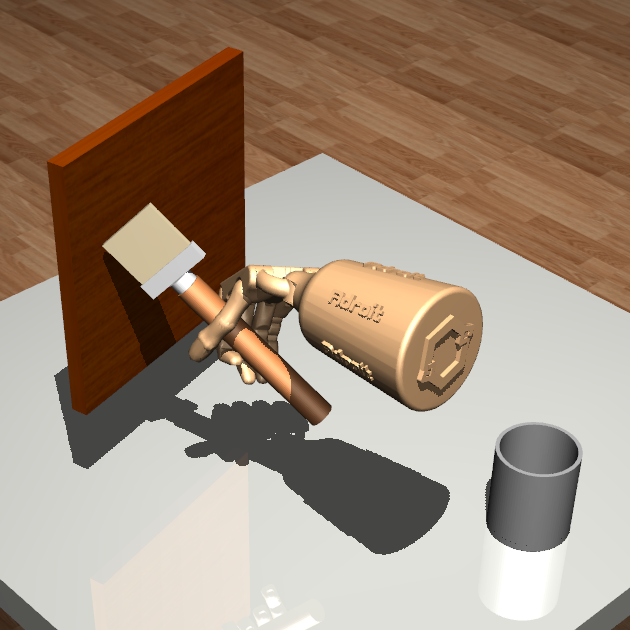} 
    \caption{Brush}
    \end{subfigure}   
    \begin{subfigure}[t]{.12\linewidth}    
     \centering
     \includegraphics[width=1\textwidth]{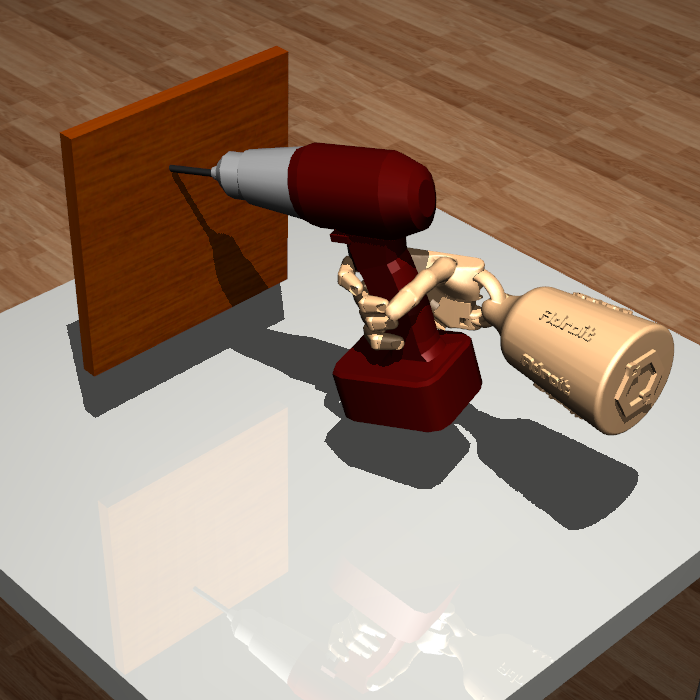} 
    \caption{Drill}
    \end{subfigure}   
    \begin{subfigure}[t]{.12\linewidth}    
     \centering
     \includegraphics[width=1\textwidth]{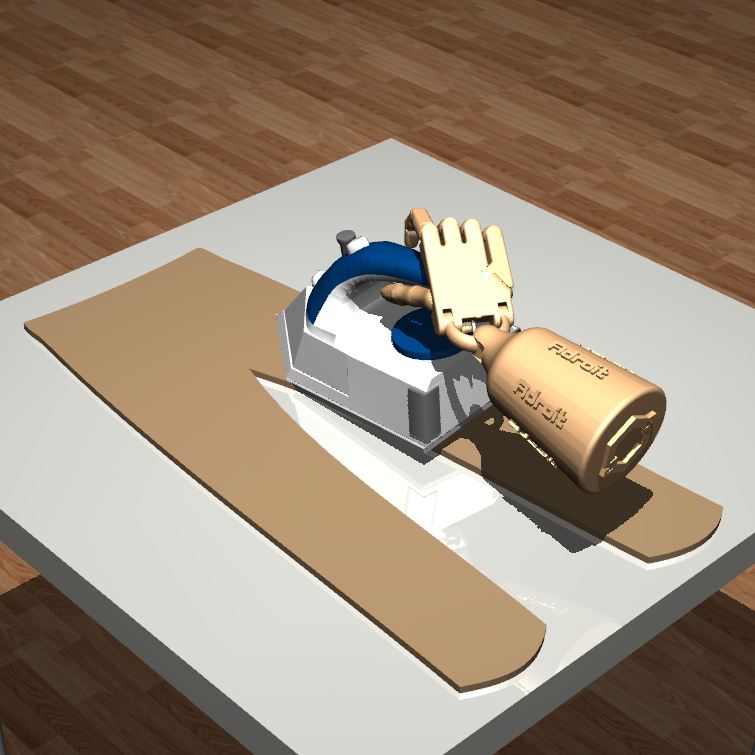} 
    \caption{Iron}
    \end{subfigure}

    \vspace{-0.2cm}
    \caption{\textbf{Simulated Evaluation Environments.} We evaluate our method on four environments from DAPG \cite{DAPG} (a-d) and propose four new robotic environments (e-h). Our new environments aim to evaluate the manipulation capabilities of a set of objects commonly used by humans, namely a pan, a brush, a drill, and a clothes iron. 
     }
    \label{fig:env_vis}
    \vspace{-0.5cm}
\end{figure*}

\makeatletter\def\Hy@Warning#1{}\makeatother

\def\thefootnote{1}\footnotetext{\hspace{0.05cm}We compare with open-source models and do not include models that do not have open access.}

\myparagraph{Baseline Encoders}
\label{sec:baselines}
We evaluate against both the commonly used, general-purpose visual encoders CLIP \cite{CLIP} and DINOv1/v2/v3~\cite{DINO,DINOv2,DINOv3}, as well as three state-of-the-art (SotA) encoders designed specifically for robotic manipulation. The baselines can be described as follows\textsuperscript{ 1}:
\begin{itemize}
    \item \textbf{CLIP \cite{CLIP}}: A visual encoder trained jointly with a language encoder in a representation alignment task on a large number of text-image pairs from the internet.
    \item \textbf{DINO \cite{DINO}}: A self-supervised visual encoder trained with a contrastive loss on images and their extracted patches. We evaluate on the three released versions DINO \cite{DINO}, DINOv2 \cite{DINOv2} and DINOv3 \cite{DINOv3}.
    \item \textbf{R3M \cite{R3M}}: A robotic manipulation baseline using a temporal contrastive loss on images and another contrastive loss estimating task completion based on video summaries matched with visual features for pairs of images.
    \item \textbf{HRP \cite{HRP}}: A visual encoder trained for robotic manipulation settings using contact points extracted from hand-object bounding box intersections and wrist trajectories extracted using FrankMocap \cite{FrankMocap}.
    \item \textbf{VC-1 \cite{VC1}}: A visual encoder targeting a variety of embodied intelligence tasks, trained on large amounts of images using the Masked Auto-Encoder \cite{MAE} paradigm.
\end{itemize}
For the methods based on ViT \cite{Dosovitskiy20ViT}, namely CLIP, DINO, DINOv3, HRP, VC-1 and \methodname, we employ a ViT-B/16 \cite{Dosovitskiy20ViT} backbone, with DINOv2 using ViT-B/14 by its authors' design. R3M uses a ResNet50 \cite{ResNet} backbone. We initialize all baselines with their publicly available weights.

\myparagraph{Result Summary} We evaluate \methodname on (1) contact anticipation, (2) simulated dexterous manipulation across 8 tasks including 4 new benchmarks, (3) real-world dexterous tasks using an ORCA \cite{ORCA} hand. 
Our results show that \methodname consistently improves generalization and fine-grained control. Specifically, \methodname achieves more accurate contact localization, leading to improved first contact and stronger implicit grasp priors, which yield greater success of in-hand manipulation. Its prior-infused features consistently outperform generic representations. 

\subsection{Evaluation of Contact Point Prediction} 
\label{sec:contact}

We first compare \methodname with other baseline encoders on the contact prediction task.
Similar to the setup in~\cite{liu2022joint}, we predict future contact points given a single input frame, whereas the original task formulation takes video sequences as input. 
This task has applications in robotics \cite{RoboABC,DenseMatcher,Anygrasp} and object interaction anticipation \cite{ContextAwareAnticipation,TransFusion,Ego4D}. 
We train cVAE-based contact prediction heads on features extracted from all baselines and our encoder. Following the original protocol, we evaluate Affordance Map Similarity (SIM) and Normalized Scanpath Saliency (NSS) on the dataset introduced in \cite{liu2022joint}. 
This dataset is derived from various egocentric cooking datasets \cite{EK55,EK100,EGTEAGazePlus} and provides 34.5K interaction hotspots, which we use as contact prediction ground-truth points.
As evident from \autoref{tab:ek100_contact_perf}, \methodname outperforms all baselines in both metrics, confirming the usefulness of its learned features. 
Note that these frame-based predictions are different from the results reported in the original paper, which shows video-based predictions. 
See Supp.~Mat.~for metric definitions and qualitative examples.
\begin{table}[t!]
\footnotesize

\centering
\caption{\textbf{Contact Prediction Evaluation.} Comparison of various vision encoders on the dataset introduced in \cite{liu2022joint}. MAPLE achieves the best performance on both SIM and NSS metrics.}
\label{tab:ek100_contact_perf}
\vspace{-0.2cm}
\setlength{\tabcolsep}{2.2pt}
\begin{tabular}{l@{\hskip 2pt}cccccccc}
\toprule
Metric & CLIP & DINO & DINOv2 & DINOv3 & HRP & VC-1 & R3M & MAPLE \\
\midrule
SIM $\uparrow$ & 0.051 & 0.061 & 0.061 & 0.059 & 0.061 & 0.064 & 0.068 & \textbf{0.068} \\
NSS $\uparrow$ & 0.561 & 0.577 & 0.647 & 0.615 & 0.602 & 0.637 & 0.651 & \textbf{0.654} \\
\bottomrule
\end{tabular}
\vspace{-0.5cm}
\end{table}

\subsection{Evaluation in Simulation Environments}
\label{sec:evaluation_setup}

\myparagraph{Existing Simulation Environments and Tasks}
We first evaluate on the four dexterous tasks from the established DAPG \cite{DAPG} benchmark suite, 
including \textit{open door}, \textit{use hammer}, \textit{rotate pen}, and \textit{relocate ball} (see \autoref{fig:env_vis}).

\myparagraph{New Environment and Task Designs}
We further introduce a new set of four custom-designed, challenging evaluation tasks focused on dexterous manipulation that require more fine-grained control. 
Specifically, we design four tasks in the MuJoCo simulator~\cite{MuJoCo} that require complex dexterous manipulation of tools using an Adroit hand \cite{AdroitHand}. 
All four tasks feature objects standing on a table, with the Adroit hand hovering above them and able to move freely (see \autoref{fig:env_vis} for visualization).
These tasks are %
    (1) \textbf{Pan}: Lift a pan 
    and place it on the induction plate of the nearby stove;
    (2) \textbf{Brush}: Retrieve the brush from a cup, then stroke it onto the nearby canvas; 
    (3) \textbf{Drill}: Lift an electric drill and align the drill bit to strike a wooden board while keeping the finger on the drill's trigger switch;
    and (4) \textbf{Iron}: Grasp a clothes iron on a table by the handle and slide it along the leg of the pants positioned nearby. 
Further details and visualizations of the tasks are in Supp.~Mat.~%

\myparagraph{Collecting Demonstrations for the New Tasks}
To conform with the established evaluation standard in literature \cite{R3M}, we record 25 expert demonstrations for each of our tasks using Rokoko Smartgloves \cite{Smartglove}, employing the algorithm from \cite{DexMV} to retarget the human hand to the Adroit hand in the simulation. Please refer to the Supp.~Mat.~for the details and goal definitions of each task. 

\newcommand{\betweenScriptAndTiny}{\fontsize{6pt}{7pt}\selectfont}
\newcommand{\lowersup}[1]{\raisebox{0.5ex}{\betweenScriptAndTiny #1}}

\begin{table*} 
\small
\centering
{\small

\newcommand{\ironDrill}[1]{#1}
\newcommand{\drill}[1]{#1}
\newcommand{\iron}[1]{#1}

\newcommand{\perfDinoDoor}{29.3\lowersup{$\pm \text{3.8}$}} 
\newcommand{\perfDinoHammer}{14.0\lowersup{$\pm \text{1.1}$}} 
\newcommand{\perfDinoBanana}{}
\newcommand{\perfDinoMug}{}
\newcommand{\perfDinoPen}{71.8\lowersup{$\pm \text{1.7}$}} 
\newcommand{\perfDinoRelocate}{32.9\lowersup{$\pm \text{0.8}$}} 
\newcommand{\perfDinoBrush}{18.0\lowersup{$\pm \text{2.5}$}} 
\newcommand{\perfDinoDrill}{24.7\lowersup{$\pm \text{1.4}$}}
\newcommand{\perfDinoIron}{13.8\lowersup{$\pm \text{1.8}$}} 
\newcommand{\perfDinoPan}{25.8\lowersup{$\pm \text{0.8}$}} 
\newcommand{\perfDinoMean}{28.8\lowersup{$\pm \text{1.2}$}}

\newcommand{\perfDinoNDVFDoor}{28.0\lowersup{$\pm \text{4.7}$}} 
\newcommand{\perfDinoNDVFHammer}{11.8\lowersup{$\pm \text{0.3}$}} 
\newcommand{\perfDinoNDVFBanana}{}
\newcommand{\perfDinoNDVFMug}{}
\newcommand{\perfDinoNDVFPen}{64.2\lowersup{$\pm \text{2.1}$}} 
\newcommand{\perfDinoNDVFRelocate}{36.9\lowersup{$\pm \text{1.3}$}} 
\newcommand{\perfDinoNDVFBrush}{17.3\lowersup{$\pm \text{3.9}$}} 
\newcommand{\perfDinoNDVFDrill}{25.3\lowersup{$\pm \text{0.5}$}}
\newcommand{\perfDinoNDVFIron}{15.1\lowersup{$\pm \text{??}$}} 
\newcommand{\perfDinoNDVFPan}{24.0\lowersup{$\pm \text{??}$}} 
\newcommand{\perfDinoNDVFMean}{29.4}

\newcommand{\perfDinoVTwoDoor}{32.0\lowersup{$\pm \text{3.4}$}} 
\newcommand{\perfDinoVTwoHammer}{10.9\lowersup{$\pm \text{0.3}$}} 
\newcommand{\perfDinoVTwoBanana}{}
\newcommand{\perfDinoVTwoMug}{}
\newcommand{\perfDinoVTwoPen}{59.8\lowersup{$\pm \text{3.0}$}} 
\newcommand{\perfDinoVTwoRelocate}{34.0\lowersup{$\pm \text{1.9}$}} 
\newcommand{\perfDinoVTwoBrush}{14.9\lowersup{$\pm \text{0.6}$}} 
\newcommand{\perfDinoVTwoDrill}{22.4\lowersup{$\pm \text{2.2}$}}
\newcommand{\perfDinoVTwoIron}{18.4\lowersup{$\pm \text{2.7}$}} 
\newcommand{\perfDinoVTwoPan}{30.4\lowersup{$\pm \text{2.2}$}} 
\newcommand{\perfDinoVTwoMean}{27.9\lowersup{$\pm \text{0.6}$}}

\newcommand{\perfDinoVThreeDoor}{34.4\lowersup{$\pm \text{3.2}$}} 
\newcommand{\perfDinoVThreeHammer}{12.2\lowersup{$\pm \text{1.0}$}} 
\newcommand{\perfDinoVThreeBanana}{}
\newcommand{\perfDinoVThreeMug}{}
\newcommand{\perfDinoVThreePen}{59.3\lowersup{$\pm \text{3.1}$}} 
\newcommand{\perfDinoVThreeRelocate}{29.3\lowersup{$\pm \text{1.2}$}} 
\newcommand{\perfDinoVThreeBrush}{18.4\lowersup{$\pm \text{3.2}$}} 
\newcommand{\perfDinoVThreeDrill}{25.3\lowersup{$\pm \text{0.5}$}}
\newcommand{\perfDinoVThreeIron}{14.0\lowersup{$\pm \text{2.0}$}} 
\newcommand{\perfDinoVThreePan}{29.6\lowersup{$\pm \text{4.1}$}} 
\newcommand{\perfDinoVThreeMean}{27.8\lowersup{$\pm \text{0.4}$}}

\newcommand{\perfClipDoor}{31.8\lowersup{$\pm \text{4.5}$}} 
\newcommand{\perfClipHammer}{13.3\lowersup{$\pm \text{1.6}$}} 
\newcommand{\perfClipBanana}{}
\newcommand{\perfClipMug}{}
\newcommand{\perfClipPen}{58.0\lowersup{$\pm \text{2.2}$}} 
\newcommand{\perfClipRelocate}{28.7\lowersup{$\pm \text{1.1}$}} 
\newcommand{\perfClipBrush}{15.3\lowersup{$\pm \text{2.2}$}} 
\newcommand{\perfClipDrill}{18.7\lowersup{$\pm \text{1.6}$}}
\newcommand{\perfClipIron}{22.0\lowersup{$\pm \text{3.8}$}} 
\newcommand{\perfClipPan}{31.8\lowersup{$\pm \text{7.6}$}} 
\newcommand{\perfClipMean}{27.4\lowersup{$\pm \text{1.9}$}}

\newcommand{\perfResNetDoor}{21.6\lowersup{$\pm \text{--}$}} 
\newcommand{\perfResNetHammer}{18.4\lowersup{$\pm \text{--}$}} 
\newcommand{\perfResNetBanana}{}
\newcommand{\perfResNetMug}{}
\newcommand{\perfResNetPen}{52.7\lowersup{$\pm \text{--}$}} 
\newcommand{\perfResNetRelocate}{31.3\lowersup{$\pm \text{--}$}} 
\newcommand{\perfResNetBrush}{22.2\lowersup{$\pm \text{--}$}} 
\newcommand{\perfResNetDrill}{}
\newcommand{\perfResNetIron}{23.6\lowersup{$\pm \text{--}$}} 
\newcommand{\perfResNetPan}{28.7\lowersup{$\pm \text{--}$}} 
\newcommand{\perfResNetMean}{--}

\newcommand{\perfHRPDoor}{23.6\lowersup{$\pm \text{2.6}$}} 
\newcommand{\perfHRPHammer}{11.6\lowersup{$\pm \text{2.3}$}}  
\newcommand{\perfHRPBanana}{}
\newcommand{\perfHRPMug}{}
\newcommand{\perfHRPPen}{52.7\lowersup{$\pm \text{1.8}$}} 
\newcommand{\perfHRPRelocate}{39.6\lowersup{$\pm \text{1.9}$}}  
\newcommand{\perfHRPBrush}{17.8\lowersup{$\pm \text{4.1}$}}  
\newcommand{\perfHRPDrill}{20.0\lowersup{$\pm \text{1.6}$}}
\newcommand{\perfHRPIron}{20.0\lowersup{$\pm \text{5.5}$}} 
\newcommand{\perfHRPPan}{19.3\lowersup{$\pm \text{2.2}$}}  
\newcommand{\perfHRPMean}{25.6\lowersup{$\pm \text{1.3}$}}

\newcommand{\perfVCODoor}{18.0\lowersup{$\pm \text{2.8}$}} 
\newcommand{\perfVCOHammer}{8.9\lowersup{$\pm \text{2.8}$}}  
\newcommand{\perfVCOBanana}{}
\newcommand{\perfVCOMug}{}
\newcommand{\perfVCOPen}{69.6\lowersup{$\pm \text{1.7}$}} 
\newcommand{\perfVCORelocate}{32.4\lowersup{$\pm \text{2.5}$}}  
\newcommand{\perfVCOBrush}{15.6\lowersup{$\pm \text{3.0}$}}  
\newcommand{\perfVCODrill}{18.7\lowersup{$\pm \text{0.5}$}}
\newcommand{\perfVCOIron}{18.7\lowersup{$\pm \text{6.8}$}} 
\newcommand{\perfVCOPan}{17.8\lowersup{$\pm \text{1.4}$}}  
\newcommand{\perfVCOMean}{24.9\lowersup{$\pm \text{1.6}$}}

\newcommand{\perfRTMDoor}{21.1\lowersup{$\pm \text{1.6}$}} 
\newcommand{\perfRTMHammer}{6.9\lowersup{$\pm \text{1.6}$}}  
\newcommand{\perfRTMBanana}{}
\newcommand{\perfRTMMug}{}
\newcommand{\perfRTMPen}{68.9\lowersup{$\pm \text{1.7}$}}  
\newcommand{\perfRTMRelocate}{31.8\lowersup{$\pm \text{2.7}$}}  
\newcommand{\perfRTMBrush}{12.4\lowersup{$\pm \text{1.6}$}}  
\newcommand{\perfRTMDrill}{18.7\lowersup{$\pm \text{2.0}$}}
\newcommand{\perfRTMIron}{14.2\lowersup{$\pm \text{1.9}$}}  
\newcommand{\perfRTMPan}{17.8\lowersup{$\pm \text{1.8}$}}  
\newcommand{\perfRTMMean}{24.0\lowersup{$\pm \text{0.5}$}}

\newcommand{\perfOursJLDoor}{31.8\lowersup{$\pm \text{6.5}$}} 
\newcommand{\perfOursJLHammer}{13.3\lowersup{$\pm \text{1.1}$}}  
\newcommand{\perfOursJLBanana}{}
\newcommand{\perfOursJLMug}{}
\newcommand{\perfOursJLPen}{75.1\lowersup{$\pm \text{1.0}$}}
\newcommand{\perfOursJLRelocate}{32.9\lowersup{$\pm \text{1.0}$}}
\newcommand{\perfOursJLBrush}{22.4\lowersup{$\pm \text{0.8}$}} 
\newcommand{\perfOursJLDrill}{25.8\lowersup{$\pm \text{0.8}$}}
\newcommand{\perfOursJLIron}{25.3\lowersup{$\pm \text{2.0}$}} 
\newcommand{\perfOursJLPan}{28.7\lowersup{$\pm \text{3.9}$}} 
\newcommand{\perfOursJLMean}{31.9\lowersup{$\pm \text{1.2}$}}

\newcommand{\perfOursGDDoor}{}
\newcommand{\perfOursGDHammer}{}
\newcommand{\perfOursGDBanana}{}
\newcommand{\perfOursGDMug}{}
\newcommand{\perfOursGDPen}{}
\newcommand{\perfOursGDRelocate}{}
\newcommand{\perfOursGDBrush}{}
\newcommand{\perfOursGDDrill}{}
\newcommand{\perfOursGDIron}{}
\newcommand{\perfOursGDPan}{}
\newcommand{\perfOursGDMean}{}

\newcommand{\perfOursHDDoor}{}
\newcommand{\perfOursHDHammer}{}
\newcommand{\perfOursHDBanana}{}
\newcommand{\perfOursHDMug}{}
\newcommand{\perfOursHDPen}{}
\newcommand{\perfOursHDRelocate}{}
\newcommand{\perfOursHDBrush}{}
\newcommand{\perfOursHDDrill}{}
\newcommand{\perfOursHDIron}{}
\newcommand{\perfOursHDPan}{}
\newcommand{\perfOursHDMean}{}

\newcommand{\oursJollyLion}{{\perfOursJLDoor} & \secondbest{\perfOursJLHammer} &  \best{\perfOursJLPen} & {\perfOursJLRelocate}    & \best{\perfOursJLBrush}  & \drill{ \best{\perfOursJLDrill} &}  \iron{\best{\perfOursJLIron} & } {\perfOursJLPan} & \best{\perfOursJLMean}  }

\newcommand{\oursJollyLionDVThree}{\secondbest{\perfOursJLDoor} & \best{\perfOursJLHammer} &  \secondbest{\perfOursJLPen} & \best{\perfOursJLRelocate}    & \secondbest{\perfOursJLBrush}  & \drill{ \perfOursJLDrill &}  \iron{\best{\perfOursJLIron} & } {\perfOursJLPan} & \best{\perfOursJLMean}  }

\newcommand{\oursGlamorousDew}{\secondbest{\perfOursGDDoor} & \secondbest{\perfOursGDHammer} &  \secondbest{\perfOursGDPen} & \perfOursGDRelocate    & \secondbest{\perfOursGDBrush}  & \drill{ \perfOursGDDrill &}  \iron{\perfOursGDIron & } \secondbest{\perfOursGDPan} & \best{\perfOursGDMean}  }

\newcommand{\oursHonestDew}{\best{\perfOursHDDoor} & \perfOursHDHammer  &  {\perfOursHDPen} & \perfOursHDRelocate   & \perfOursHDBrush  & \drill{\perfOursHDDrill &} \iron{\perfOursHDIron &} \perfOursHDPan  & \perfOursHDMean  }

\caption{\textbf{Results on Simulation Environments.} We report the task completion success rate for eight simulation tasks. 
\best{Green} indicates the best performance, and \secondbest{blue} the second best. We report the standard deviations across randomization seeds. \methodname reaches the best mean performance on the evaluation suite with variance reduced across tasks with higher complexity. 
}
\label{tab:eval_dexterous}
\vspace{-0.2cm}
\begin{tabular}{lccccccccc}
\toprule
Model & Door & Hammer & Pen & Relocate & Brush & \drill{Drill &} \iron{Iron &} Pan & Mean  \\ \midrule
CLIP \cite{CLIP} & \perfClipDoor & \secondbest{\perfClipHammer}  &  \perfClipPen &  \perfClipRelocate    & \perfClipBrush & \perfClipDrill & \perfClipIron & \blbest{\perfClipPan}  & \perfClipMean \\
DINO \cite{DINO} & {\perfDinoDoor} & \blbest{\perfDinoHammer}  &  \secondbest{\perfDinoPen} &  {\perfDinoRelocate}    & {\perfDinoBrush}  & \drill{\perfDinoDrill &} \iron{{\perfDinoIron} &} {\perfDinoPan} & \blsecondbest{\perfDinoMean}      \\
DINOv2 \cite{DINO} & \secondbest{\perfDinoVTwoDoor} & {\perfDinoVTwoHammer}  &  \perfDinoVTwoPen &  \blsecondbest{\perfDinoVTwoRelocate}    & {\perfDinoVTwoBrush}  & \drill{\perfDinoVTwoDrill &} \iron{{\perfDinoVTwoIron} &} \blsecondbest{\perfDinoVTwoPan} & {\perfDinoVTwoMean}      \\
DINOv3 \cite{DINO} & \best{\perfDinoVThreeDoor} & {\perfDinoVThreeHammer}  &  \perfDinoVThreePen &  {\perfDinoVThreeRelocate}   & \blsecondbest{\perfDinoVThreeBrush}  & \drill{\blsecondbest{\perfDinoVThreeDrill} &} \iron{{\perfDinoVThreeIron} &} {\perfDinoVThreePan} & {\perfDinoVThreeMean}      \\
HRP \cite{HRP} & {\perfHRPDoor} & \perfHRPHammer  &  \perfHRPPen & \blbest{\perfHRPRelocate}   & \perfHRPBrush  & \drill{\perfHRPDrill &} \iron{\blsecondbest{\perfHRPIron} &} \perfHRPPan  & \perfHRPMean      \\
VC-1 \cite{VC1} & \perfVCODoor & {\perfVCOHammer}  &  \perfVCOPen & \perfVCORelocate    & \perfVCOBrush  & \drill{\perfVCODrill &} \iron{\perfVCOIron &} \perfVCOPan & \perfVCOMean \\
R3M \cite{R3M} & \perfRTMDoor & \perfRTMHammer  &  {\perfRTMPen}  & {\perfRTMRelocate}   & {\perfRTMBrush}  & \drill{\perfRTMDrill &} \iron{{\perfRTMIron} &} \perfRTMPan  & {\perfRTMMean}      \\
\cmidrule(lr){1-10}
MAPLE (ours) & \oursJollyLion \\
\bottomrule
\end{tabular}
}
\vspace{-0.3cm}

\end{table*}

\myparagraph{Experiment Protocol}
Our evaluation uses both existing tasks and newly designed ones. To evaluate a given visual encoder on a given task, MLP-based policies \cite{R3M} are trained to control a robot hand in simulation given proprioceptive input and features from the visual encoder. Policies are evaluated every 1,000 out of a total of 20,000 training steps, with each evaluation consisting of 50 rollouts in the simulator. During each rollout, we randomize the position and/or rotation of the object(s). The policy receives a binary score indicating whether it fulfilled the intended task according to an environment-specific success criterion, e.g.~opening the door by a wide enough angle. The score of the encoder on the task is then the best average success rate across all 20 evaluations. Following prior work R3M~\cite{R3M}, we calculate scores of encoders on tasks by additionally averaging the scores across three camera views, and three random seeds per view used to vary the policy network's initialization, for a total of nine experiments per encoder and task. This averaging procedure helps improve the statistical robustness of the evaluation.   
For our custom-designed tasks, we employ a temporal horizon of 750 time steps. Gaussian noise is additionally added to the generated action to simulate imperfect execution and increase the tasks' difficulty. See the Supp.~Mat.~for success definitions. 

\myparagraph{Results}
\label{sec:results}
\autoref{tab:eval_dexterous} compares \methodname with various baselines and SotA methods across eight dexterous manipulation tasks. 
\methodname achieves the best overall performance among all methods, exceeding the second-best approach by more than two standard deviations. 
While other baselines occasionally outperform our method on individual tasks, \methodname exhibits strong generalization across all 8 tasks, achieving the highest success rate in four tasks and ranking a close second-best in a fifth, whereas no other considered method leads in more than one. 

Except for \methodname, encoders explicitly designed for robotic manipulation struggle to outperform the generalist DINO and CLIP models in terms of mean success rates, indicating a lack of generalizability beyond their training domains. 
In contrast, \methodname excels in fine-grained manipulation tasks such as the in-hand pen reorientation, brush and drill tasks, all of which require precise grasping. 

Finally, the high success rates of \methodname in our newly introduced environments, despite heavily randomized initial tool positions, further validate its robustness and efficacy for object localization and manipulation.

\begin{table}[t!]
\centering
{\footnotesize
\caption{\textbf{Ablation of Loss Terms.} We report the effect of training with and without the contact loss, as well as different types of hand pose losses on the Door, Hammer, Pen and Relocate tasks. Checkmarks (\checkmark) indicate the inclusion and crosses (\xmark) indicate the exclusion of a loss term. The use of the hand pose loss without the hand tokenizer is indicated by $\bar{\mathrm{T}}$. Success rates are averages across all seeds and camera views of the respective tasks. 
}
\label{tab:loss_ablations_main}
}
\vspace{-0.2cm}
\footnotesize
\begin{tabular}{ccccccc}
\toprule
$\mathcal{L}_{ct}$ &  $\mathcal{L}_{hand}$ & Door & Hammer & Pen & Relocate & Mean \\ \midrule
\xmark & \checkmark  &  29.3   & 10.0  &   72.0     &    32.7 & 36.0      \\  
\checkmark & \xmark  &  \textbf{32.4}  & 9.8 & 68.9 & 31.1 & 35.6      \\  
\checkmark & $\bar{\mathrm{T}}$ & 31.8 & \textbf{13.3} &  70.2 &  31.6 & 36.7  \\  
\cmidrule(lr){1-7}
\checkmark & \checkmark  &  31.8 & \textbf{13.3}  & \textbf{75.1}  & \textbf{32.9} &    \textbf{38.3} \\
\bottomrule 
\end{tabular}

\vspace{-0.5cm}
\end{table}

\myparagraph{Analysis}
We ablate the contribution of our contact and hand pose losses on the four DAPG dexterous tasks~\cite{DAPG} in \autoref{tab:loss_ablations_main}. Performance on the in-place, yet highly dexterous pen task suffers from excluding $\mathcal{L}_{hand}$ while being less sensitive to the exclusion of the localization-focused $\mathcal{L}_{ct}$, as expected from the task's in-place nature. In all settings, training the model with a hand pose tokenizer leads to a higher performance than letting it reconstruct hand poses from scratch. The less dexterous but localization-focused door task, requiring the policy to locate the handle and push it down in any feasible way, suffers most from the exclusion of $\mathcal{L}_{ct}$. This shows the benefits of using hand pose tokenization and demonstrates the usefulness of $\mathcal{L}_{hand}$ for dexterous tasks as well as $\mathcal{L}_{ct}$ for tasks emphasizing localization.

\subsection{Evaluation in Real-World Settings} 
\label{sec:evaluation_realworld}
\begin{figure}
    \centering
    \includegraphics[width=0.9\linewidth]{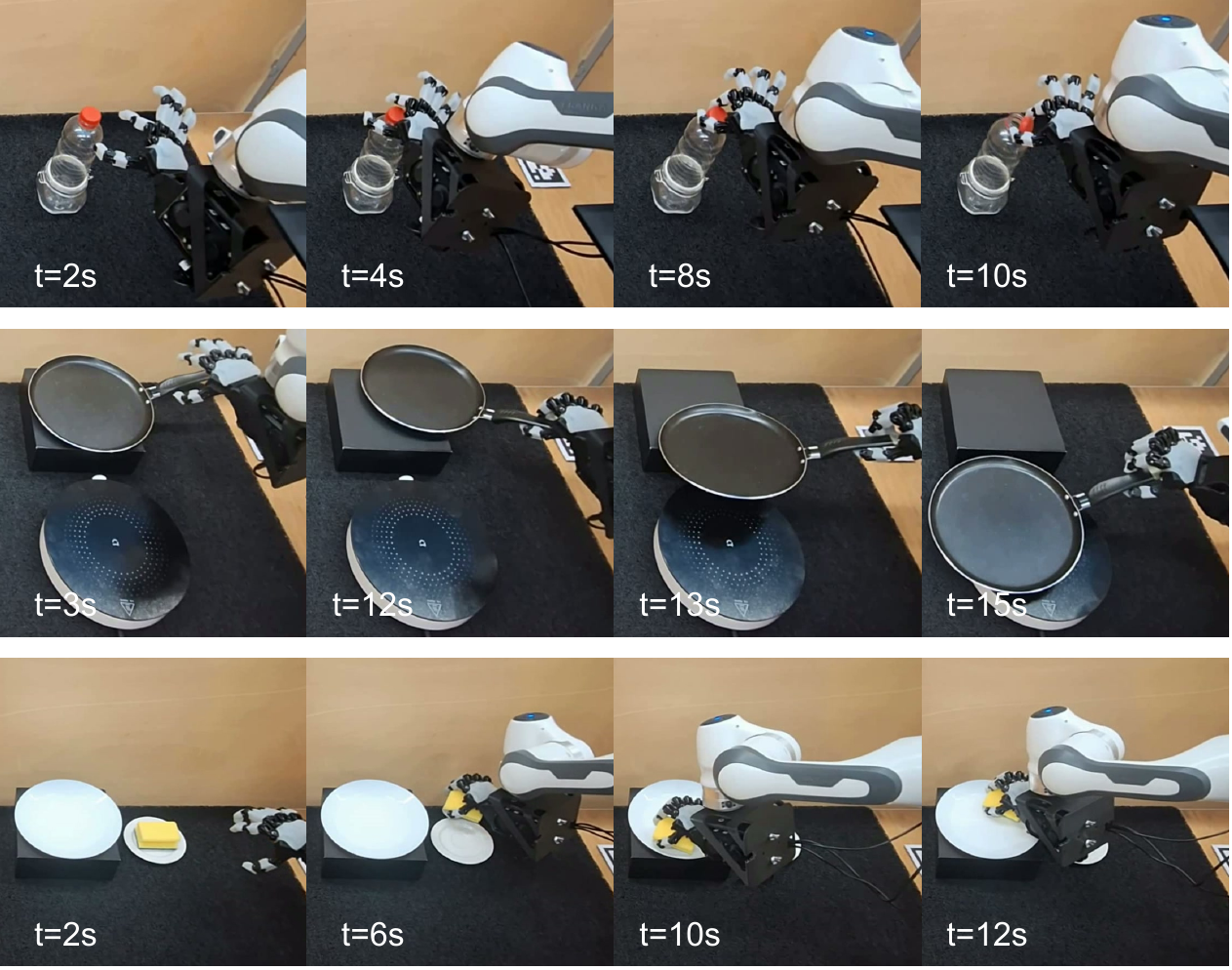}
    \vspace{-0.3cm}
    \caption{\textbf{Real World Sequences of the Evaluated Tasks.} Example rollouts of \methodname. From top to bottom: `Open the bottle cap' (bottle), `Place the pan` (pan) and `Wash the dish` (sponge) tasks using the ORCA hand and the Franka Emika Panda manipulator. 
    }
    \label{fig:real_world_sequences}
    \vspace{-0.3cm}
\end{figure}

\myparagraph{Task Designs}
Our real-world evaluation features one task derived from those used in simulation, as well as two novel tasks:
(1) \textbf{Pan}: The robot grasps the handle of a pan, lifts it, and places it onto a burner. After placement, the robot releases the handle. This task features an object commonly found in the Ego4D dataset.
(2) \textbf{Bottle}: The robot needs to open the cap of a bottle. This requires executing a twisting motion after a correct pinching grasp of the cap.\\
(3) \textbf{Sponge}: The robot picks up a sponge, moves it to an adjacent plate, sets it down to make contact with the plate's surface, and circles the sponge around the plate while maintaining the grasp. 
Additional task descriptions and data collection details can be found in Supp.~Mat. 

\myparagraph{Hardware Setup}
Our setup comprises an ORCA robotic hand~\cite{ORCA} mounted on a Franka Emika Panda arm~\cite{FrankaEmika}, along with an external RGB camera and a wrist-mounted RGB camera located below the hand. Human demonstrations are collected using Rokoko Smartgloves and the Coil Pro motion capture system~\cite{Smartglove}, which provide proprioceptive location data from the demonstrator’s fingers and wrist.

\myparagraph{Experiment Protocol} 
After training on egocentric videos, we freeze our encoder to ensure stable feature extraction and use the extracted features for manipulation policy training. 
We employ the Diffusion Policy framework~\cite{chi2023diffusion} to predict robot actions using the Diffusion Transformer (DiT)~\cite{dasari2024DiTpolicy}. 
Visual inputs are processed by our trained encoders, and their embeddings are combined with proprioceptive data.
We use DDiM~\cite{Song2021ddim} as our noise scheduler, with 100 steps during training and 8 steps during inference. 
Further architectural and training details are in Supp.~Mat. 

\begin{figure}
    \vspace{-0.1cm}
    \centering
    \includegraphics[width=1.0\linewidth]{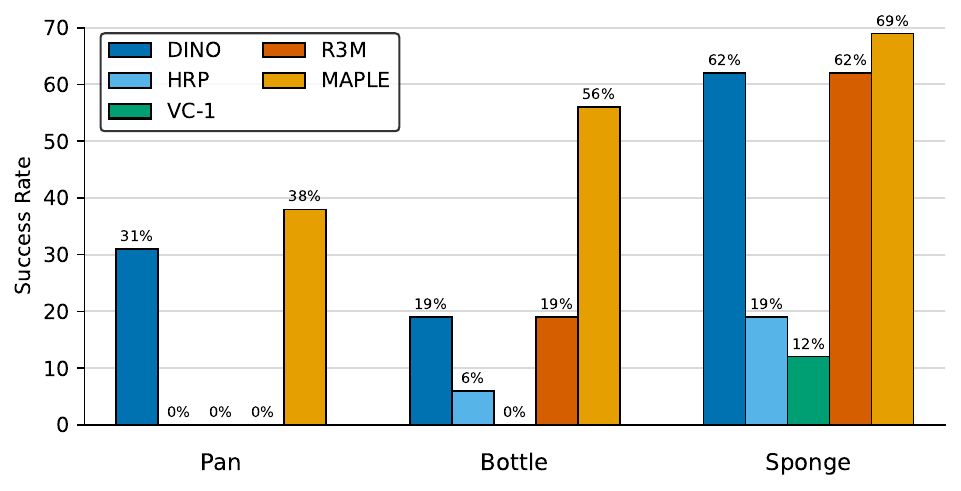}
    \vspace{-0.8cm}
    \caption{\textbf{Results on Real-World Environments.} We evaluate different encoders, using the extracted features for imitation learning. Missing bars indicate zero success rates.}
    \label{fig:real_world_successrate}
    \vspace{-0.5cm}
\end{figure}

\myparagraph{Results} 
For each model and task, we conduct trials using 16 settings to evaluate success rates, each with a different position and/or rotation of the objects in the scene. \autoref{fig:real_world_successrate} summarizes the performance across all models.
Our results indicate that our encoder consistently outperforms alternative approaches. \methodname shows the highest overall performance, benefiting from its stability and strong generalizability across varying task conditions.
In contrast, other encoder-based methods exhibit behaviors that make them less suitable for our tasks.
We encourage the reader to visit the project webpage for videos of our real-world rollouts.

\begin{figure}
    \vspace{0.3cm}
    \centering
    \includegraphics[width=0.95\linewidth]{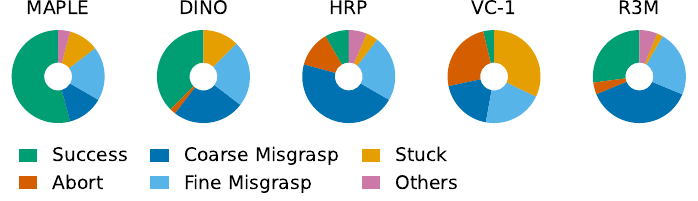}
    \vspace{-0.2cm}
    \caption{\textbf{Failure Analysis.} We report the failure case analysis for each method, aggregated across all real-world experiments.}
    \label{fig:failure_analysis}
    \vspace{-0.5cm}
\end{figure}

\myparagraph{Failure Analysis} We analyze the failure modes of the methods evaluated and present the results in \autoref{fig:failure_analysis}. 
Failures are categorized as: runs aborted automatically or manually for hardware safety (``abort"), lack of motion (``stuck"), incorrect object localization (``coarse misgrasp"), or failed manipulation (``fine misgrasp"), alongside other cases (``others").
Notably, \methodname is the only method that never caused an abort and showed the least localization errors (counting coarse and fine misgrasps). 
The analysis confirms the superior efficacy and safety of our encoder against baselines for dexterous manipulation tasks in the real world.

\section{Conclusion}
In this work, we studied how to infuse dexterous manipulation priors into visual representations by training on diverse human videos, aiming to improve performance on downstream robotic tasks in human-centric environments. 
%
%
We introduced \methodname, a novel approach that learns manipulation-specific priors from egocentric videos. \methodname trains an encoder to predict hand-object contact points and detailed 3D hand poses from a single image, producing features useful for downstream dexterous manipulation tasks. For the benefit of the community and a more rigorous evaluation, we further proposed new dexterous simulation environments.
Our strong results on both new tasks and existing benchmark tasks in simulation, as well as our superior real-world performance demonstrated promising manipulation prior learning and excellent generalization ability with \methodname. 
Future directions of research include combining \methodname with language conditioning and incorporating more diverse, possibly exocentric human manipulation datasets during its training. 
\label{sec:conclusion}

\section{Acknowledgements}
This work was supported as part of the Swiss AI Initiative by a grant from the Swiss National Supercomputing Centre (CSCS) under project ID a03 on Alps, as well as the Swiss National Science Foundation Advanced Grant 216260: ``Beyond Frozen Worlds: Capturing Functional 3D Digital Twins from the Real World".

{
   \small
   \bibliographystyle{ieeenat_fullname} 
   \bibliography{main}
}

\unhidefromtoc

\addtocounter{part}{1}  

\clearpage

\maketitlesupplementary

\renewcommand{\thesection}{S\arabic{section}}  
\renewcommand{\thetable}{S\arabic{table}}  
\renewcommand{\thefigure}{S\arabic{figure}} 

\renewcommand{\cftsecnumwidth}{20pt}

\setcounter{section}{0}

\begin{center}
\begin{minipage}{0.48\textwidth}  
\setlength{\parskip}{0pt}         
\makeatletter
\makeatother
{\small                
\setcounter{tocdepth}{3}
\tableofcontents
}
\end{minipage}
\end{center}
\vspace{5pt}

\newcommand{\citesupp}[1]{\cite{#1}}  

\section{Evaluation Details in Real World}
\label{sec:supp_eval_real}

\begin{figure}[!ht]
    \centering
    \includegraphics[width=0.99\linewidth]{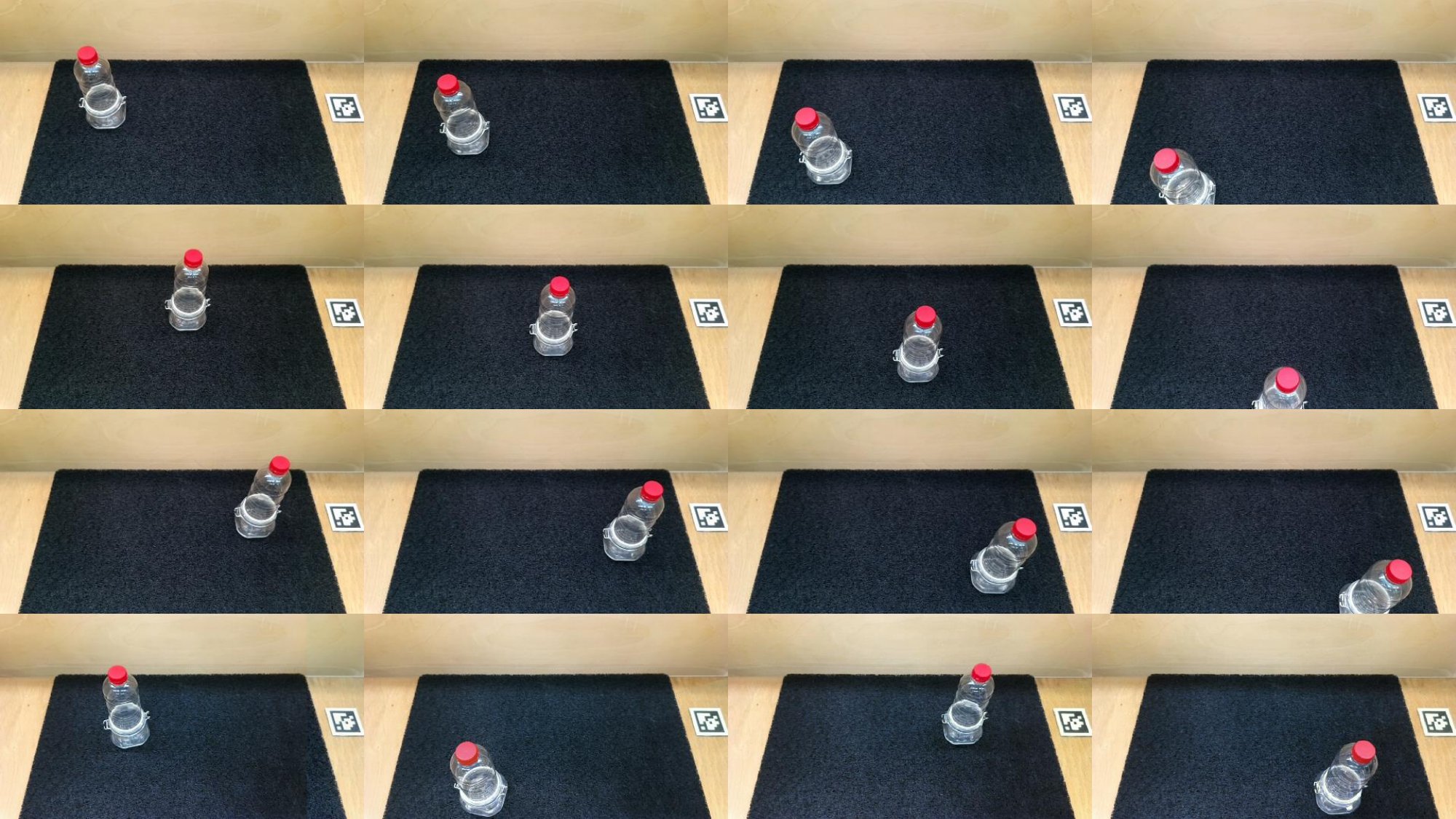}\\[0.15cm]
    \includegraphics[width=0.99\linewidth]{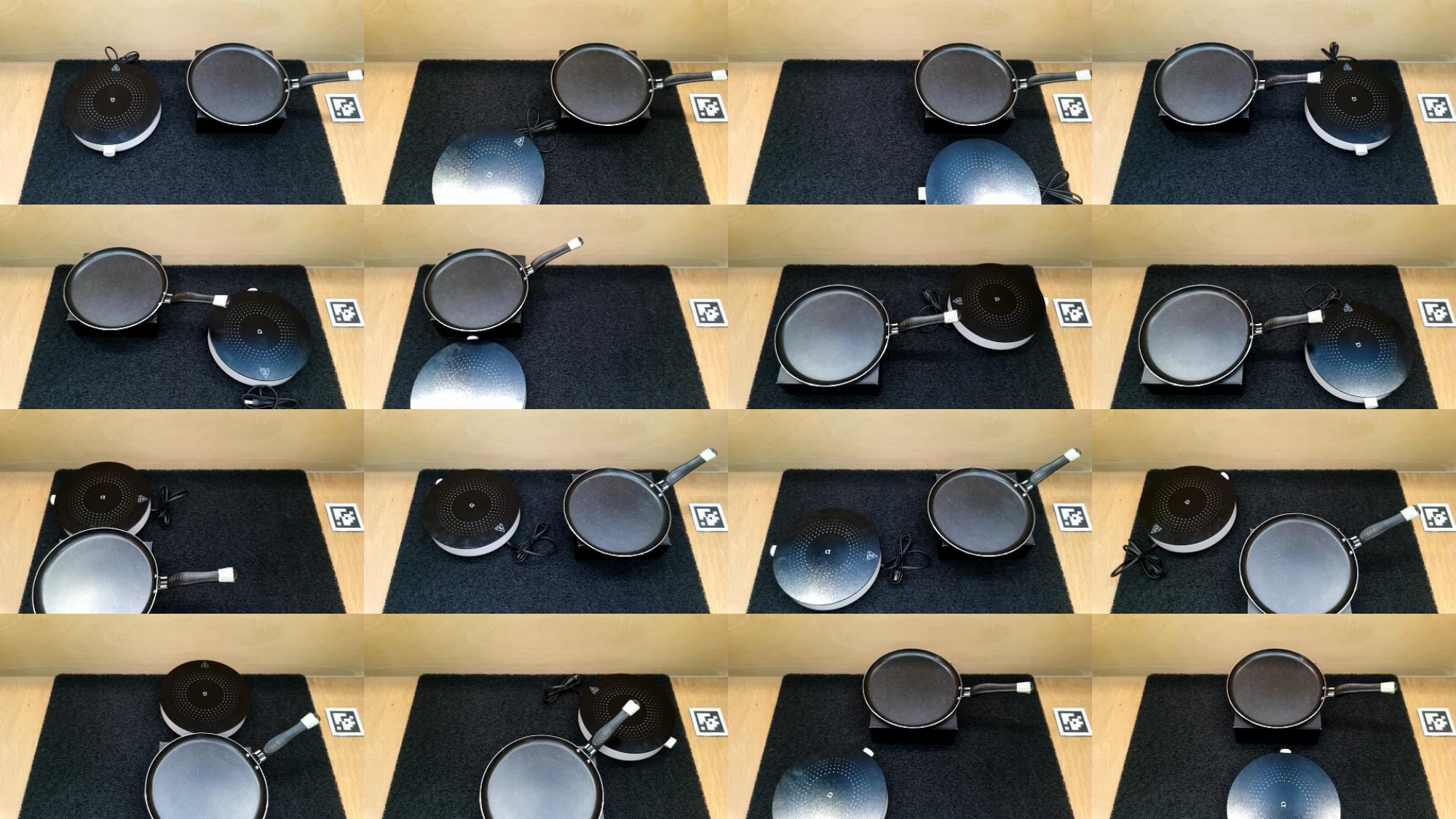}\\[0.15cm]  
    \includegraphics[width=0.99\linewidth]{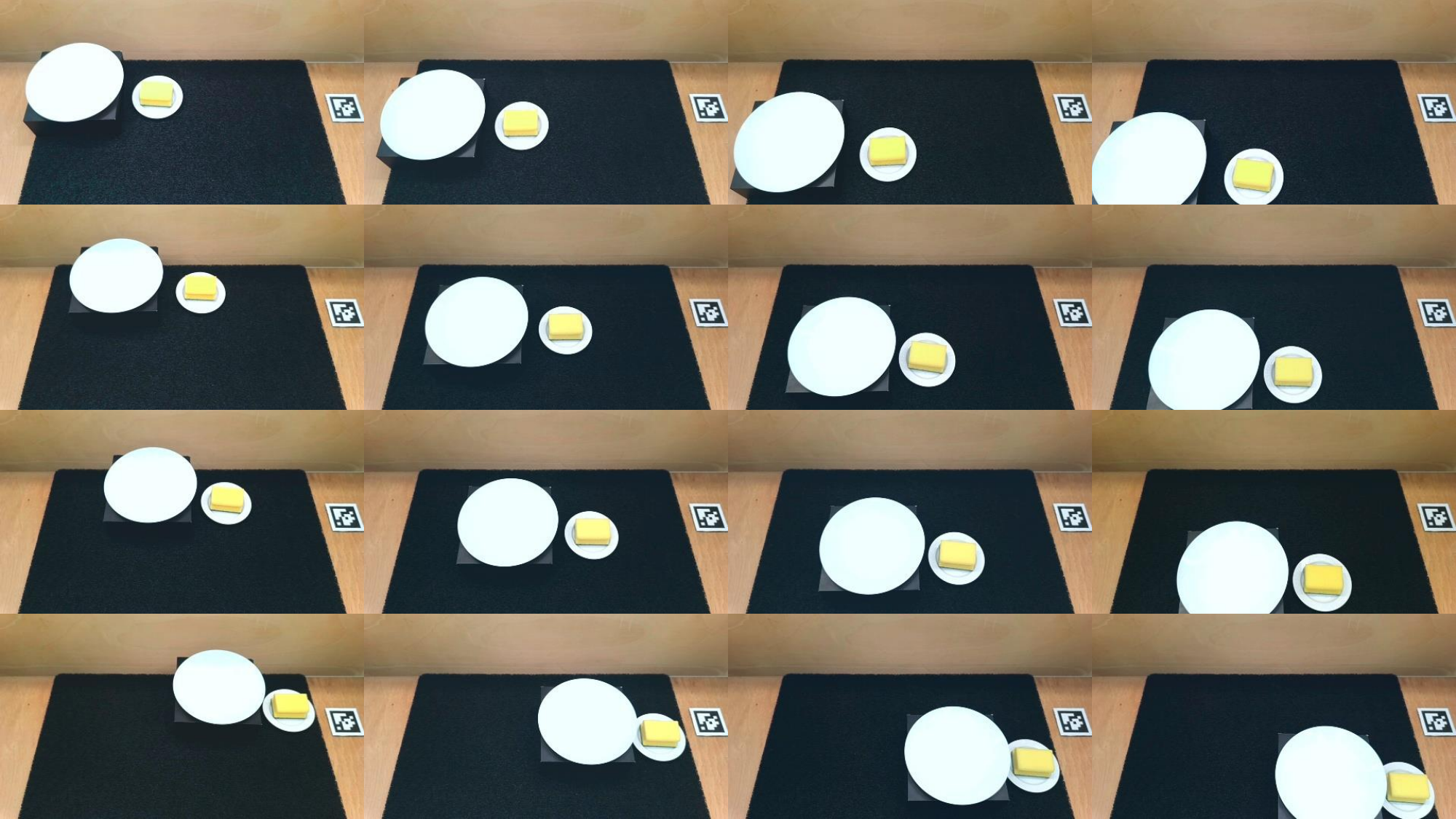}
    \caption{The initial setups of all real-world evaluations: open the bottle cap (bottle), place the pan (pan), and wash the dish (sponge). Each method is evaluated for 16 episodes, starting from each of the states shown in the figure.
}
    \vspace{-0.5cm}
\label{fig:real_init}
\end{figure}

In this section, we provide more details on the imitation learning evaluations introduced in Sec.~4.3. 
Each method is evaluated for 16 episodes for every task, with predefined initial setups as visualized in Fig.~\ref{fig:real_init}.

\subsection{Robot Platform and Policy Execution}
Our system consists of an ORCA robotic hand~\citesupp{ORCA} mounted on a Franka Emika Panda robotic arm~\citesupp{FrankaEmika}. One external and one wrist-mounted OAK-D camera \cite{OakD} capture the scene, and the arm and hand are controlled at 10 Hz. The arm’s action is defined by a 7-dimensional end-effector pose (x, y, z, plus a quaternion), and the hand’s action is represented by a 17-dimensional vector corresponding to its joint positions. All camera images are cropped and resized to $480\times 270$ pixels. Our framework uses an observation window spanning two time steps, while control actions are executed in chunks spanning 10 time steps.

\subsection{Failure Cases}

\myparagraph{Coarse/Fine Misgrasp} In Fig.~7, we define ``coarse misgrasp'' as an episode in which the policy moves the robot arm toward a direction not leading to the target object, closes the hand without grasping anything, then either stops moving the arm in a significant manner for over 15 seconds or fails to touch the object at least 6 times after attempting to grasp the object again. In case the policy moves the arm within the hand's grasping range, or the hand touches the object but fails to lift it, a ``fine misgrasp'' classification follows instead.

\myparagraph{Stuck} An episode in which the hand never closes its grasp yet the arm stops significant movement for at least 15 seconds is classified as ``stuck''.

\myparagraph{Abort} Additionally, the policy execution framework automatically aborts the episode in case the arm moves outside the defined joint limits, a maximum joint velocity is exceeded, or the difference between the measured and expected joint position grows too large (e.g.~as a result of the arm colliding with an obstacle). Alternatively, the human operator may stop the execution manually if they detect a dangerous deformation of the hand's fingers due to an incorrect interaction with an object, or in case of an imminent collision of the hand with a wall. In such an event, the episode is classified as ``abort''.

\myparagraph{Other} Rarely observed reasons for failure classified as ``other'' may include knocking over the object, placing it in an unintended target location, inability to open the cap of a bottle despite grasping it due to not strong enough movements, or similar.

\subsection{Imitation Learning Details}
\label{sec:supp_il_detail}
To accelerate training and enable larger batch sizes, we cache embeddings from the observation encoders directly within the dataset. This approach significantly speeds up training. To increase the policy's robustness to visual perturbations, we create 10 cached copies for each observation, each augmented differently.

Our DiT~\citesupp{dasari2024DiTpolicy} backbone has a model dimension of 512, consists of 6 blocks, and employs 8 attention heads. The network is trained using the Adam \cite{Adam} optimizer with a cosine warmup learning rate schedule. All policies working in the real world are evaluated after 35,000 gradient steps.

\subsection{Detailed Task Descriptions}

We report details regarding the demonstration collection and evaluation of each of our three real-world tasks.

\subsubsection{``Open the Bottle Cap'' Task}

\myparagraph{Demonstrations} 285 demonstrations are collected and
used for training. 80\% of them are used as training set, 20\% as evaluation set. The starting position of the bottle is randomized. Demonstrations are collected by an experienced operator wearing Rokoko Smartgloves and the Coil Pro motion capture setup \citesupp{Smartglove}, which provides absolute proprioceptive data of the demonstrator's fingers and wrist positions. Images from two cameras are collected and fed during policy rollout: front camera and wrist camera.

\myparagraph{Evaluation} Each method is evaluated for 16 episodes, all starting from the same set of initial conditions as shown in Fig.~\ref{fig:real_init}.a. 
Before each trial, the arm and hand are moved to a normative pose.
An episode is counted as successful only if the bottle cap is opened using the fingers. Due to the difficulty of the task, the policy is allowed to reattempt opening the cap up to 6 times if it misses the bottle.

\subsubsection{``Place the Pan'' Task}

\myparagraph{Demonstrations} 450 demonstrations are collected and used for training. 80\% of them are used as training set, 20\% as evaluation set. The position and orientation of the pan and stove on the table is randomized. The demonstrations are collected with the same Rokoko solution as described before. Images from two cameras are collected and fed during policy rollout: front camera and wrist camera.

\myparagraph{Evaluation} Each method is evaluated for 16 episodes, all starting from the same set of initial conditions as shown in Fig.~\ref{fig:real_init}.b. 
Before each trial, the arm and hand are moved to a normative pose.
An episode is counted as successful only if the pan is grasped, lifted and placed stably on the induction stove.

\subsubsection{``Wash the Dish'' Task}

\myparagraph{Demonstrations} 227 demonstrations are collected and used for training. 80\% of them are used as training set, 20\% as evaluation set. The starting position of the sponge and plate together is randomized. The demonstrations are collected using the same Rokoko solution as described above. Images from two cameras are collected and fed during the policy rollout: front camera and wrist camera.

\myparagraph{Evaluation} Each method is evaluated for 16 episodes, all starting from the same set of initial conditions as shown in Fig.~\ref{fig:real_init}.c.
Before each trial, the arm and hand are moved to a normative pose.
An episode is counted as successful only if the sponge is grasped, lifted, placed on the plate and rotated around the plate's interior at least three times.
\hfill

\section{Zero-Shot Real World Experiments}

\begin{figure}[t!]
    \centering
     \begin{subfigure}[t]{.45\linewidth}
     \centering
    \includegraphics[width=1\textwidth]{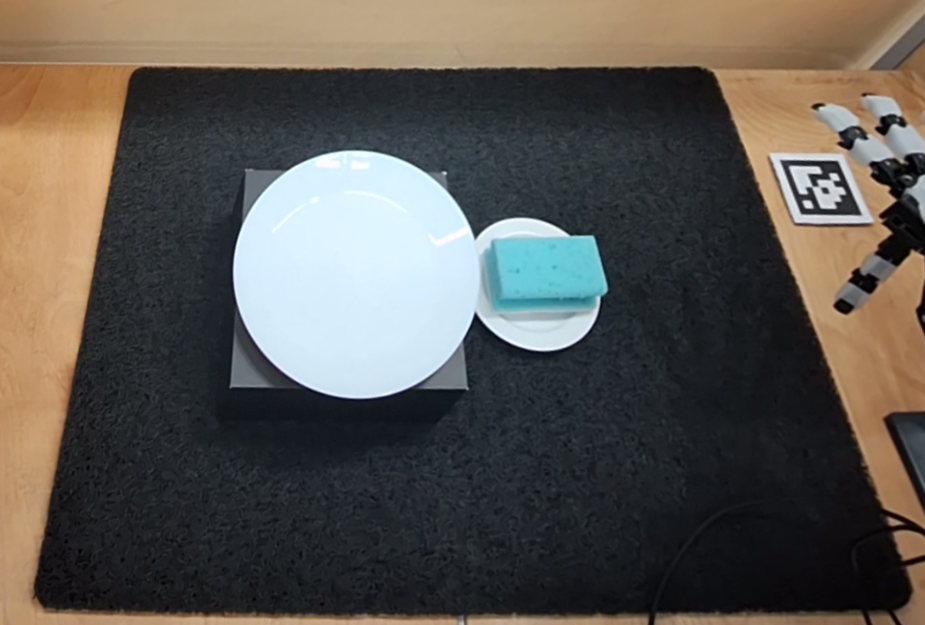}

    \end{subfigure}   
    \begin{subfigure}[t]{.45\linewidth}    
     \centering
     \includegraphics[width=1\textwidth]{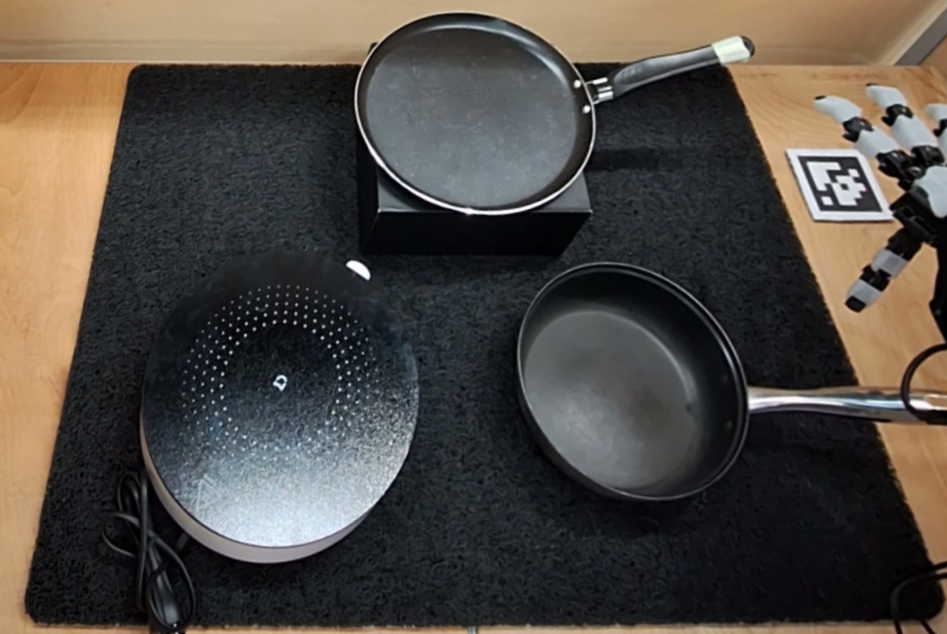} 
    \end{subfigure}
    \caption{\textbf{Visualization of Zero-Shot Initial Configurations.} Exemplary initial configurations for our zero-shot task variations. We modify the ``Wash the Dish'' task to use a sponge of a different color and shape than seen during training (``Unseen Object''). We further conduct evaluations on the ``Place the Pan'' task when an additional pan is present in the image as a ``Distractor Object''.}
    \label{fig:zeroshot}   
\end{figure}

To test the robustness of policies trained using our features against those using features of baseline methods, we devise two zero-shot experiments in which the policies are exposed to out-of-domain visual inputs without corresponding training data and measure the resulting success rates. Specifically, we evaluate the methods on a version of the ``Wash the Dish'' task featuring a sponge with a different color and shape. Moreover, we add an additional pan to the ``Place the Pan'' task and evaluate the policies. We conduct 16 trials for each method and task, following the positions visualized in \autoref{fig:real_init}. Exemplary configurations are visualized in \autoref{fig:zeroshot}.

We summarize the result in \autoref{tab:zeroshot}. The policies using our features outperform those using competing encoders, demonstrating the promising zero-shot capabilities and robustness of MAPLE.

\begin{table}[t!]
\small
\centering
\setlength{\tabcolsep}{3pt}
\begin{tabular}{l@{\hskip 2pt}cccccccc}
\toprule
Experiment & DINO & HRP & VC-1 & R3M & MAPLE \\
\midrule
Unseen Object & 25.0\% & 0\% & 0\% & 25.0\% & \textbf{31.3\%} \\
Distractor Object & 0\% & 0\% & 0\% & 0\% & \textbf{18.8\%} \\
\bottomrule
\end{tabular}
\caption{\textbf{Performance on Zero-Shot Experiments.} We report success rates for MAPLE and four baseline methods on our zero-shot task variations. See \autoref{fig:zeroshot} for visualizations of some initial configurations.}
\label{tab:zeroshot}
\vspace{-0.5cm}
\end{table}

\section{Demonstration Data and Benchmark Environments for Simulation}

\begin{figure*}
    \centering
   
  \renewcommand{\arraystretch}{2.0}

    \resizebox{0.8\textwidth}{!}{
            \begin{tabular}{cccccc} 
\includegraphics[width=.90\linewidth]{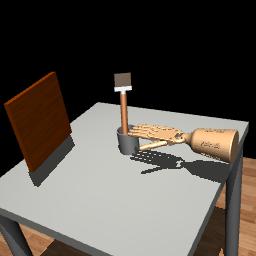}
\includegraphics[width=.90\linewidth]{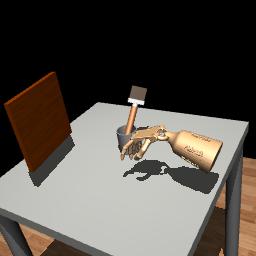}
\includegraphics[width=.90\linewidth]{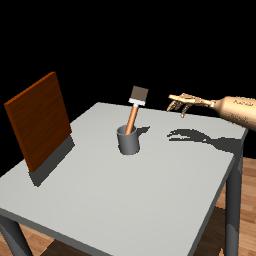}
\includegraphics[width=.90\linewidth]{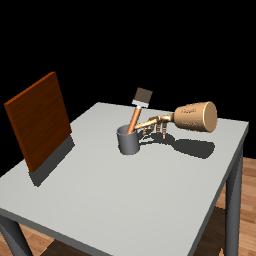}
\includegraphics[width=.90\linewidth]{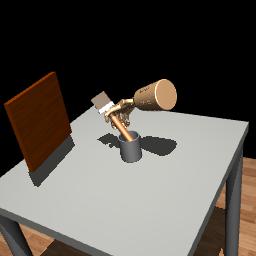}
\includegraphics[width=.90\linewidth]{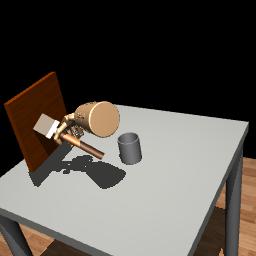}
\includegraphics[width=.90\linewidth]{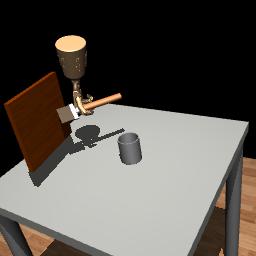} \\ \vspace{2mm} 
\includegraphics[width=.90\linewidth]{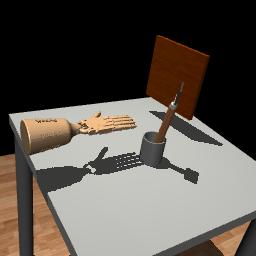}
\includegraphics[width=.90\linewidth]{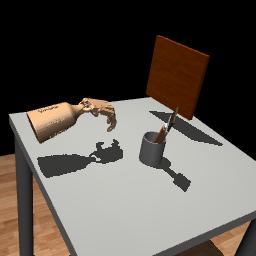}
\includegraphics[width=.90\linewidth]{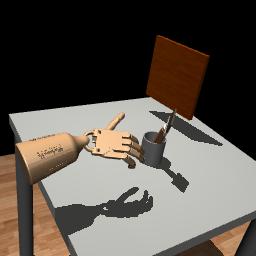}
\includegraphics[width=.90\linewidth]{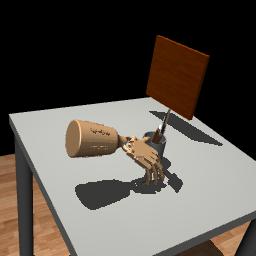}
\includegraphics[width=.90\linewidth]{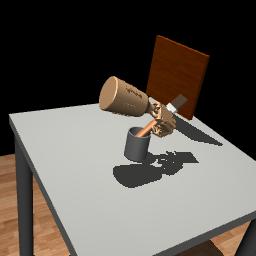}
\includegraphics[width=.90\linewidth]{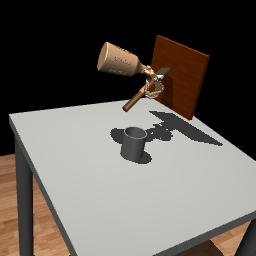}
\includegraphics[width=.90\linewidth]{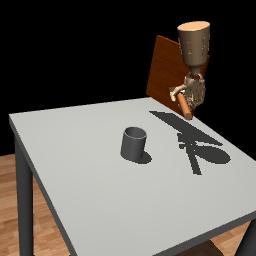} \\ \vspace{2mm} 
\includegraphics[width=.90\linewidth]{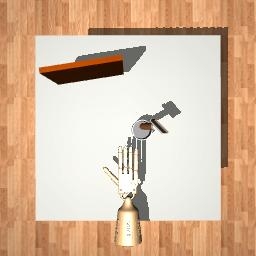}
\includegraphics[width=.90\linewidth]{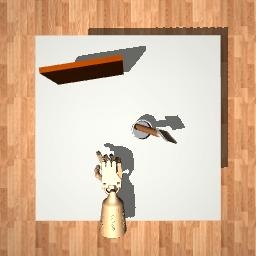}
\includegraphics[width=.90\linewidth]{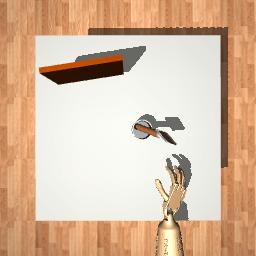}
\includegraphics[width=.90\linewidth]{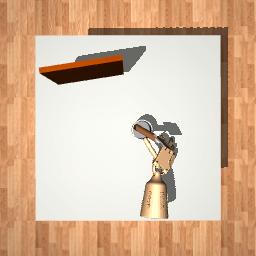}
\includegraphics[width=.90\linewidth]{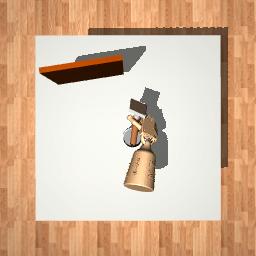}
\includegraphics[width=.90\linewidth]{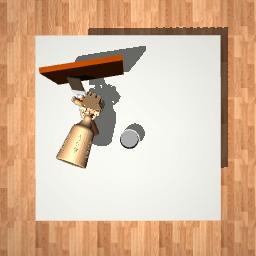}
\includegraphics[width=.90\linewidth]{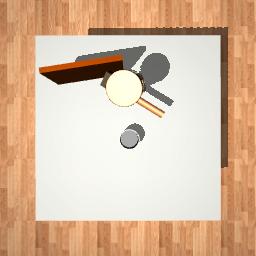} \\ \vspace{2mm} 

            \end{tabular}}
            
    \caption{\textbf{Example Demonstration Trajectory:} Camera observations of a demonstration for the 'Brush' task. The top row displays the left camera view, the middle row shows the right camera view, and the bottom row presents the top camera view.}
    \label{fig:rollout_paint}
\end{figure*}

\begin{figure*}[t!]
    \centering
   
  \renewcommand{\arraystretch}{2.0}

    \resizebox{0.8\textwidth}{!}{
            \begin{tabular}{cccccc} 
\includegraphics[width=.90\linewidth]{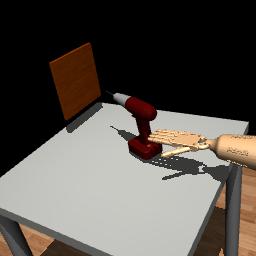}
\includegraphics[width=.90\linewidth]{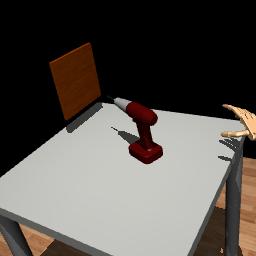}
\includegraphics[width=.90\linewidth]{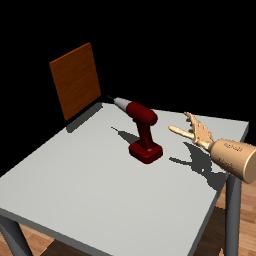}
\includegraphics[width=.90\linewidth]{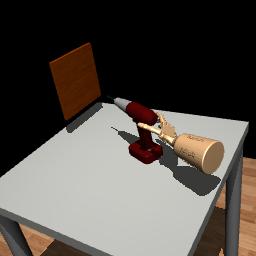}
\includegraphics[width=.90\linewidth]{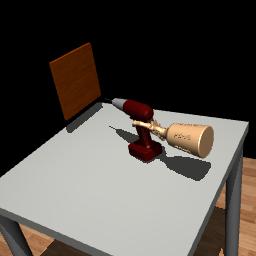}
\includegraphics[width=.90\linewidth]{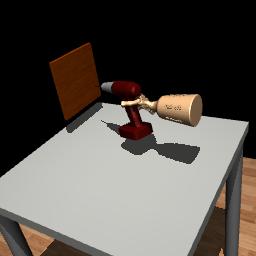}
\includegraphics[width=.90\linewidth]{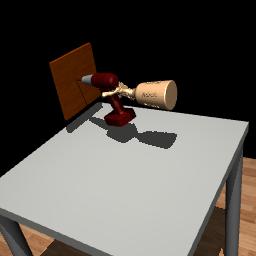} \\ \vspace{2mm} 
\includegraphics[width=.90\linewidth]{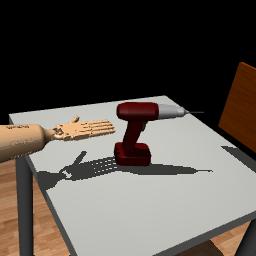}
\includegraphics[width=.90\linewidth]{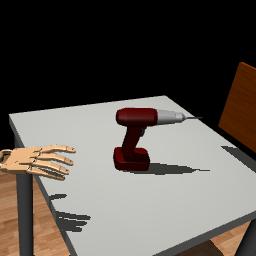}
\includegraphics[width=.90\linewidth]{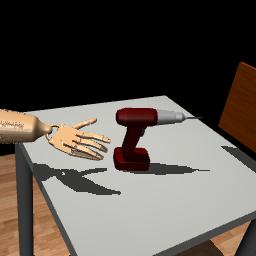}
\includegraphics[width=.90\linewidth]{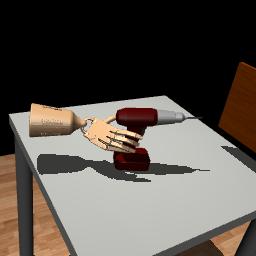}
\includegraphics[width=.90\linewidth]{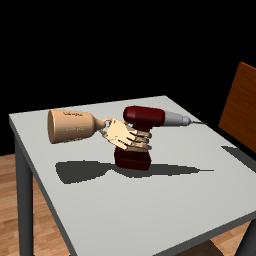}
\includegraphics[width=.90\linewidth]{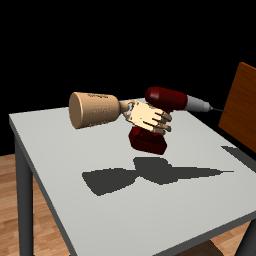}
\includegraphics[width=.90\linewidth]{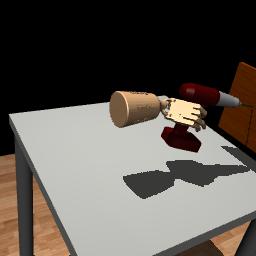} \\ \vspace{2mm} 
\includegraphics[width=.90\linewidth]{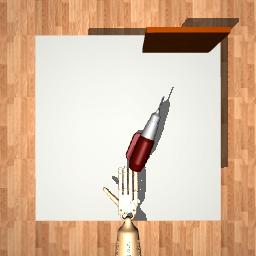}
\includegraphics[width=.90\linewidth]{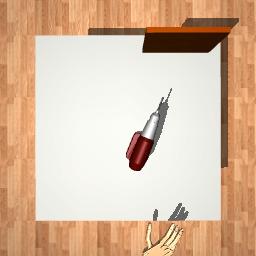}
\includegraphics[width=.90\linewidth]{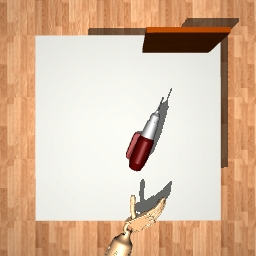}
\includegraphics[width=.90\linewidth]{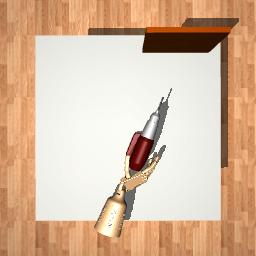}
\includegraphics[width=.90\linewidth]{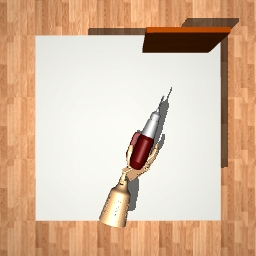}
\includegraphics[width=.90\linewidth]{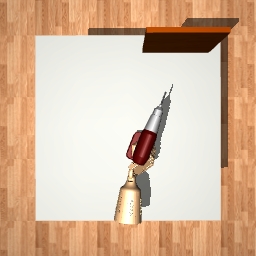}
\includegraphics[width=.90\linewidth]{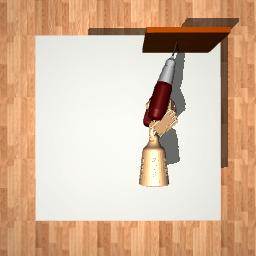} \\ \vspace{2mm} 

            \end{tabular}}
            
    \caption{\textbf{Example Demonstration Trajectory:} Camera observations of a demonstration for the 'Drill' task. The top row displays the left camera view, the middle row shows the right camera view, and the bottom row presents the top camera view.}
    \label{fig:rollout_drill}
\end{figure*}

\begin{figure*}[ht]
    \centering
   
  \renewcommand{\arraystretch}{2.0}

    \resizebox{0.8\textwidth}{!}{
            \begin{tabular}{cccccc} 
\includegraphics[width=.90\linewidth]{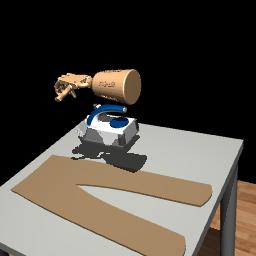}
\includegraphics[width=.90\linewidth]{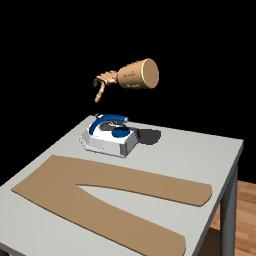}
\includegraphics[width=.90\linewidth]{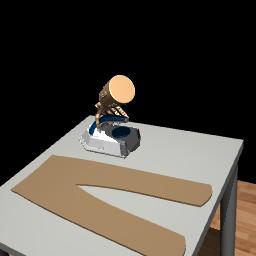}
\includegraphics[width=.90\linewidth]{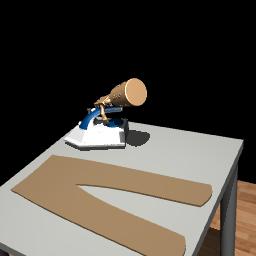}
\includegraphics[width=.90\linewidth]{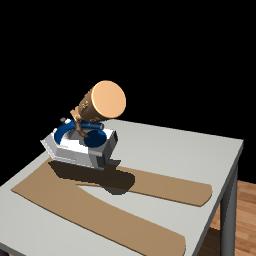}
\includegraphics[width=.90\linewidth]{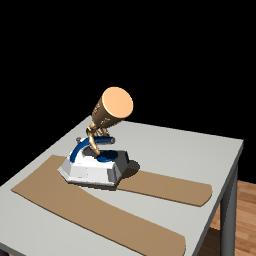}
\includegraphics[width=.90\linewidth]{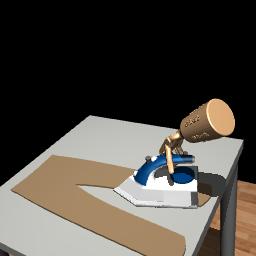} \\ \vspace{2mm} 
\includegraphics[width=.90\linewidth]{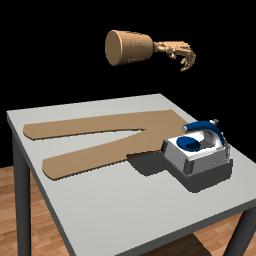}
\includegraphics[width=.90\linewidth]{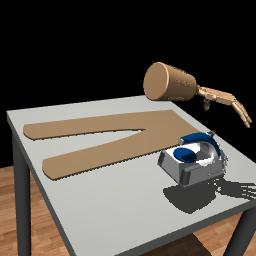}
\includegraphics[width=.90\linewidth]{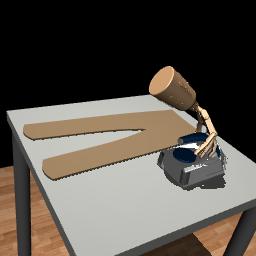}
\includegraphics[width=.90\linewidth]{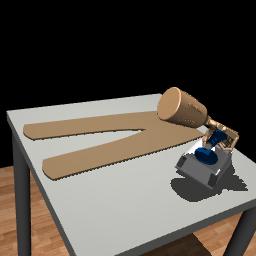}
\includegraphics[width=.90\linewidth]{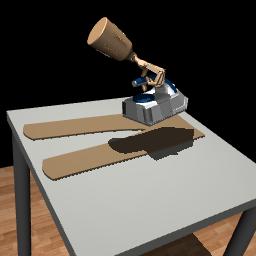}
\includegraphics[width=.90\linewidth]{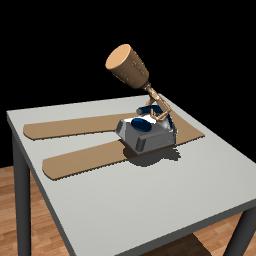}
\includegraphics[width=.90\linewidth]{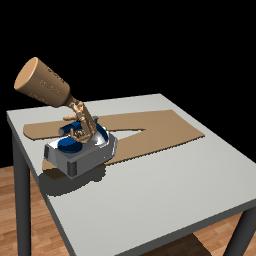} \\ \vspace{2mm} 
\includegraphics[width=.90\linewidth]{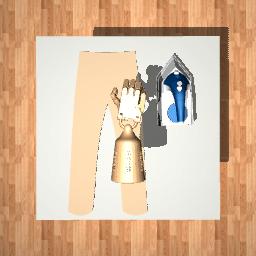}
\includegraphics[width=.90\linewidth]{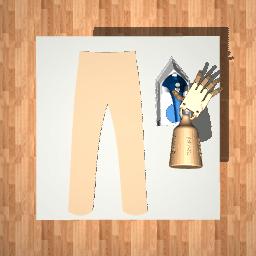}
\includegraphics[width=.90\linewidth]{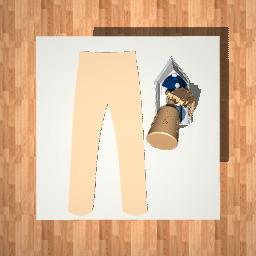}
\includegraphics[width=.90\linewidth]{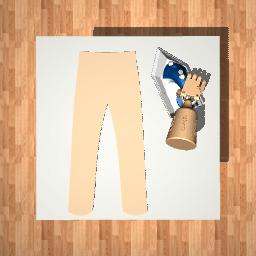}
\includegraphics[width=.90\linewidth]{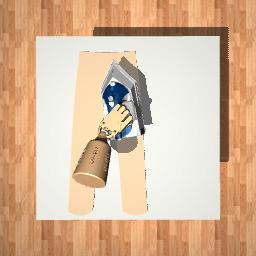}
\includegraphics[width=.90\linewidth]{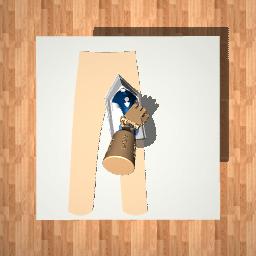}
\includegraphics[width=.90\linewidth]{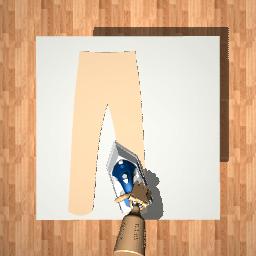} \\ \vspace{2mm} 

            \end{tabular}}
            
    \caption{\textbf{Example Demonstration Trajectory:} Camera observations of a demonstration for the 'Iron' task. The top row displays the left camera view, the middle row shows the right camera view, and the bottom row presents the top camera view.}
    \label{fig:rollout_iron}
\end{figure*}

\begin{figure*}[ht]
    \centering
   
  \renewcommand{\arraystretch}{2.0}
    \resizebox{0.8\textwidth}{!}{
            \begin{tabular}{cccccc} 
\includegraphics[width=.90\linewidth]{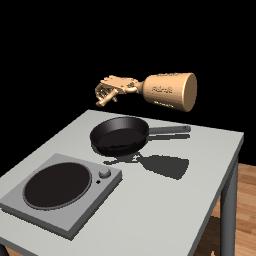}
\includegraphics[width=.90\linewidth]{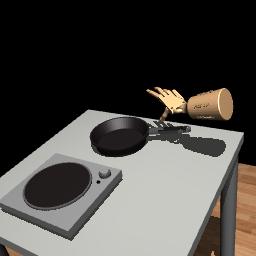}
\includegraphics[width=.90\linewidth]{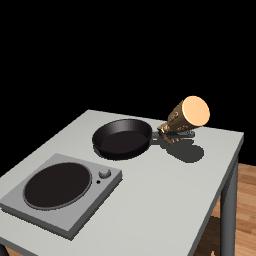}
\includegraphics[width=.90\linewidth]{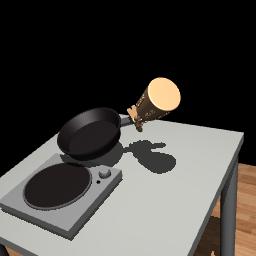}
\includegraphics[width=.90\linewidth]{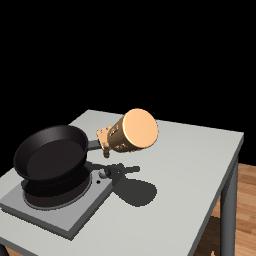}
\includegraphics[width=.90\linewidth]{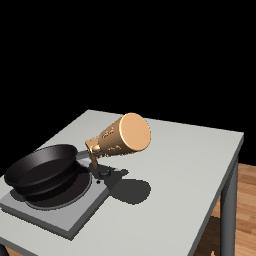}
\includegraphics[width=.90\linewidth]{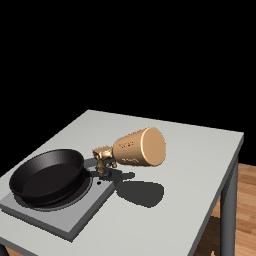} \\ \vspace{2mm} 
\includegraphics[width=.90\linewidth]{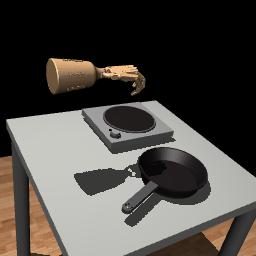}
\includegraphics[width=.90\linewidth]{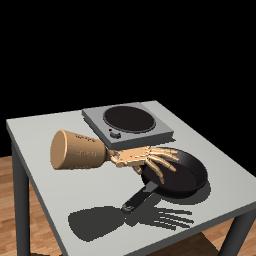}
\includegraphics[width=.90\linewidth]{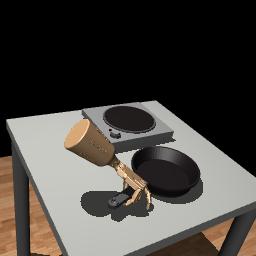}
\includegraphics[width=.90\linewidth]{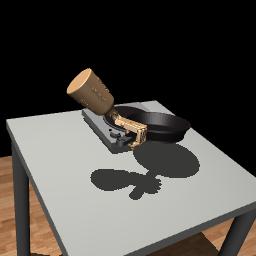}
\includegraphics[width=.90\linewidth]{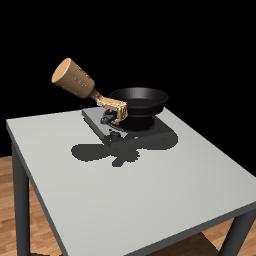}
\includegraphics[width=.90\linewidth]{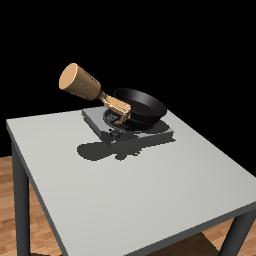}
\includegraphics[width=.90\linewidth]{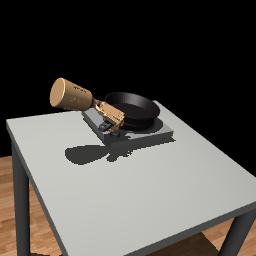} \\ \vspace{2mm} 
\includegraphics[width=.90\linewidth]{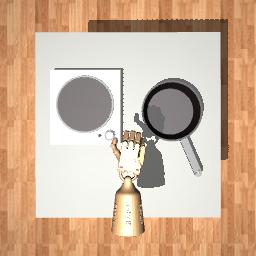}
\includegraphics[width=.90\linewidth]{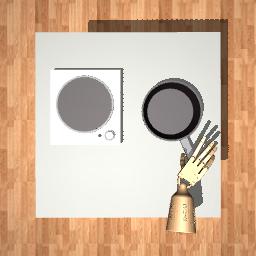}
\includegraphics[width=.90\linewidth]{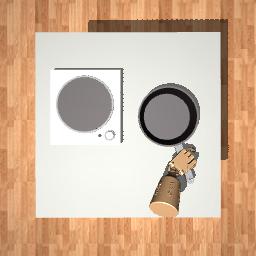}
\includegraphics[width=.90\linewidth]{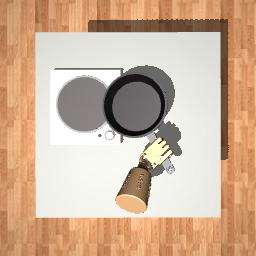}
\includegraphics[width=.90\linewidth]{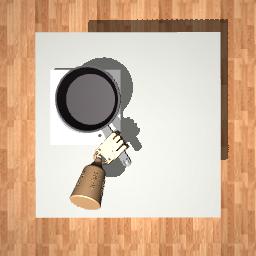}
\includegraphics[width=.90\linewidth]{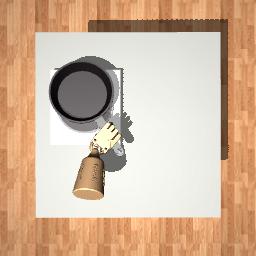}
\includegraphics[width=.90\linewidth]{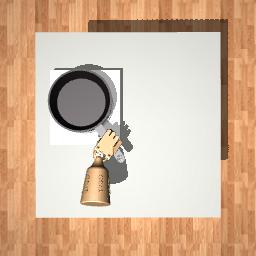} \\ \vspace{2mm} 

            \end{tabular}}
            
    \caption{\textbf{Example Demonstration Trajectory:} Camera observations of a demonstration for the 'Pan' task. The top row displays the left camera view, the middle row shows the right camera view, and the bottom row presents the top camera view.}
    \label{fig:rollout_pan}
\end{figure*}
\subsection{Hand Retargeting Algorithm}
The algorithm from \citesupp{DexMV} works by optimizing over the vector of the Adroit joint actuations, minimizing a Huber loss on the difference between pairs of an inter-fingertip or finger-to-wrist vector on the human hand and a corresponding vector measuring the same distance on the robotic hand. The hand pose is inferred by the aforementioned retargeting procedure, while the global position of the wrist joint is obtained using the Rokoko Coil Pro, an EMF-based tracking device. We use virtual PID controllers to drive each joint.

\subsection{Visualization of Demonstration Trajectories}
We show selected frames from our demonstration trajectories for the brush (\autoref{fig:rollout_paint}), the drill (\autoref{fig:rollout_drill}), the iron (\autoref{fig:rollout_iron}) and the pan (\autoref{fig:rollout_pan}) tasks for each camera view.

\label{sec:rollouts}

\section{Evaluation Details in Simulation}
\label{sec:evaluation_details}

To evaluate the suitability of a given visual encoder for downstream dexterous manipulation tasks, we use a simulator in which a policy network consuming features from a visual encoder is trained to operate an observed robotic manipulator. As usual in a real-world deployment, the policy network additionally receives proprioceptive features describing the current joint configurations of the robot as input. Position and orientation information related to the object(s) to be manipulated is only provided through visual cues, so as to increase realism and encourage the use of visual features to grasp and manipulate the object. All considered environments feature the Adroit \citesupp{AdroitHand} dexterous hand with 30 DoF in total, which is allowed to move freely within a predefined operating range above the table holding the object(s) to manipulate.

The policy networks are trained to perform the respective tasks using demonstrations recorded by human participants, as described in Sec.~4.2. While training the policy networks, we periodically perform rollouts (i.e.~simulation runs where the performance of the policy is assessed). The policies are trained for 20,000 training iterations. Rollouts are performed every 1,000 training iterations, and the highest success rate achieved over a given run is chosen for the computation of the final average success rates reported in the tables. 50 rollouts are used during each evaluation, and we calculate the success rate for a given policy and task by averaging over the success rates obtained using 3 different viewpoints and 3 policy seeds per viewpoint. We randomize the position and orientation of the object(s) to manipulate upon each evaluation start to make the tasks more challenging and realistic.

During the training of the policy, the checkpoint of the respective visual encoder is used as a frozen feature extractor, and only the parameters of the policy are trained. For the MAPLE model and all ablations, we use the checkpoint obtained after training 3,000 iterations on our Ego4D-based dataset. This iteration count was chosen to match the approximate contact prediction performance saturation point on a small validation dataset, manually constructed from prediction frames produced by our data extraction pipeline. In all experiments, the policy is a 2-layer MLP with 256 units for each hidden layer. The MLP receives the visual features produced by the respective encoder from a rendered RGB view of the input scene. Examples of rendered simulator views are visualized in Figure 3. The MLP further receives a number $p$ of proprioceptive features encoding the current joint configuration of the jaw gripper resp.\ robotic hand. Specifically, due to the hand being fixed, we use $p=24$ for the DAPG pen environment, and $p=30$ for all other environments due to the hand's movement adding 6 additional DoF. All these values are chosen to match the setup in previous work \citesupp{R3M}. The number of proprioceptive features generally corresponds to the DoF of the robotic manipulator. At each simulation timestep, given these environment observations, the MLP networks output actuation vectors which determine the force applied at each actuator according to the conversion outlined in \citesupp{MujocoXMLReference}. To learn from human demonstrations, the policy is provided observation inputs recorded during human teleoperation of the robot manipulator, and is trained to predict the actuation vector corresponding to the action that the human performed at that timestep. As is standard in literature \citesupp{DAPG, R3M}, we train a separate policy per environment and use the same base seed for all object position/orientation randomizations to ensure reproducibility.

\subsection{Environment Success Criteria}

Here, we provide detailed criteria that must be fulfilled for a task to count as successfully completed, for each of our dexterous environments. For distance measurements, MuJoCo \texttt{site} markers \citesupp{MujocoXMLReference} attached to the robotic hand,  manipulated object (\textit{brush, drill, iron, pan})
and target object (\textit{wooden board, clothes, induction stove})  are used.

\begin{itemize}
    \item \textbf{Brush:} The tip of the brush must contact the wooden board's front side while the palm of the hand is in the close vicinity of the handle of the brush.
    \item \textbf{Drill:} The index finger must be in close vicinity of the drill's trigger, and the tip of the drill bit must be in close vicinity of the wooden board's front side.
    \item \textbf{Iron:} The bottom of the clothes iron must slide along at least three of eight markers placed equidistantly along the right leg of the pants laid out on the table, while the palm of the hand is in close vicinity of the handle of the clothes iron.  
    \item \textbf{Pan:} The bottom of the pan must be in close vicinity of the induction stove. Here, we explicitly do not require the hand to contact the pan's handle to permit a (slight) dropping of the pan to the induction stove to achieve the defined goal. 
\end{itemize}

\section{Contact Point Prediction Details}

Here, we provide definitions for the SIM and NSS metrics used in Sec.~4.1, training details, as well as qualitative examples of predictions obtained using our method and a representative baseline.

\subsection{Metric Definitions}

Following the evaluation protocol of \cite{liu2022joint}, for Tab.~1 we train a conditional variational autoencoder \cite{cVAE} working with features from various encoders and predicting $c=5$ contact points per given input image showing a person about to interact with some object of a scene. Setting $c>1$ helps account for the ambiguity of this contact point prediction, arising from multiple possible actions. To evaluate the performance of the resulting contact point predictor, each predicted contact point for the given image is rendered as an isotropic Gaussian with $\sigma=3$ on a heatmap $M \in \mathbb{R}^{H\times W}$ of height $H$ and width $M$. The ground-truth contact point is rendered in an identical manner on a ground-truth heatmap $\hat{M} \in \mathbb{R}^{H\times W}$.
Given $M$ and $\hat{M}$, we define $SIM$ and $NSS$ for the corresponding image as follows:
\begin{equation*}
    SIM(M, \hat{M}) = \sum_{y=1}^{H} \sum_{x=1}^{W} \min\left(\frac{M_{y,x}}{n_M}, \frac{\hat{M}_{y,x}}{n_{\hat{M}}}\right),
\end{equation*}
\begin{equation*}
    NSS(M, \hat{M}) = \sum_{y=1}^{H} \sum_{x=1}^{W}  \mathbb{I} ({\hat{M}_{y,x} > 0}) \cdot \frac{M_{y,x} - \mu_M}{\sigma_M},
\end{equation*}
where
\begin{equation*}
    n_K = \sum_{y=1}^{H} \sum_{x=1}^{W} K_{y,x},\ \mu_K = \frac{n_K}{H\times W},
\end{equation*}
\begin{equation*}
    \sigma_K = \sqrt{\frac{\sum_{y=1}^{H} \sum_{x=1}^{W} (K_{y,x} - \mu_K)^2}{H\times W}},
\end{equation*}
and $\mathbb{I}$ is the indicator function. The final $SIM$ and $NSS$ metrics are averages over their image-wise values for the dataset used in \cite{liu2022joint}. Following the official implementation of \cite{liu2022joint}, we set $H = 32,\ W = 32$.

\subsection{Training Details}

We train the cVAE contact predictors for 3000 iterations, evaluating them every 150 iterations. A method's score in Tab.~1 is determined based on the checkpoint that reached the highest SIM score during all evaluations of the method. The losses used are identical to those in \cite{liu2022joint}. We use AdamW \cite{AdamW} with a learning rate of $5\times10^{-5}$ and a batch size of 128.

\subsection{Qualitative Examples}

\begin{figure}[t]
    \centering
    
    \begin{subfigure}[b]{0.32\linewidth}
        \centering
        \includegraphics[width=\linewidth]{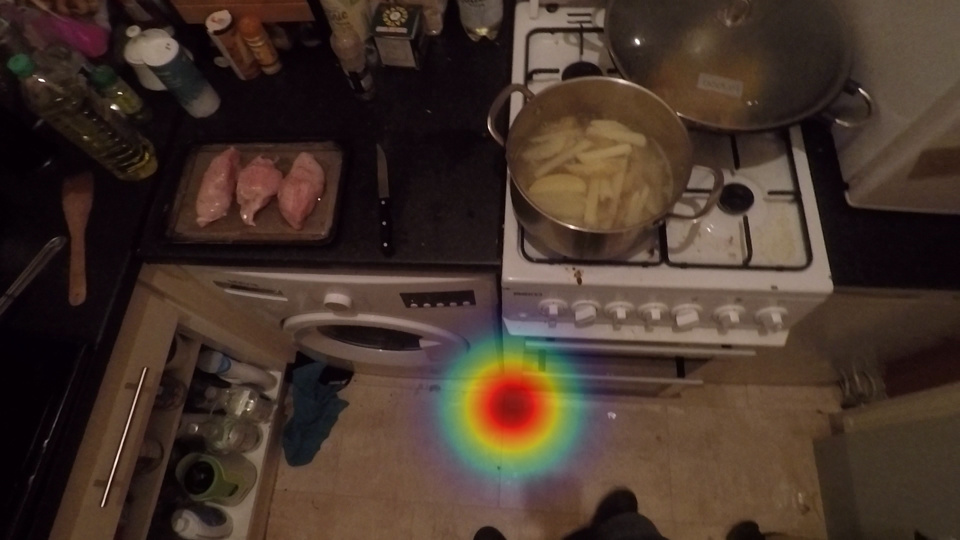}
        \caption{DINO}
    \end{subfigure}
    \hfill
    \begin{subfigure}[b]{0.32\linewidth}
        \centering
        \includegraphics[width=\linewidth]{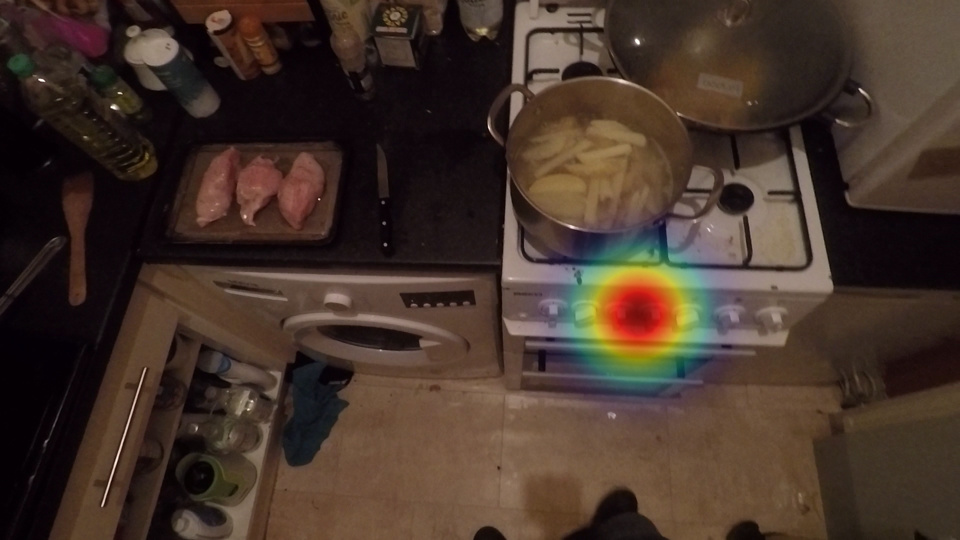}
        \caption{MAPLE}
    \end{subfigure}
    \hfill
    \begin{subfigure}[b]{0.32\linewidth}
        \centering
        \includegraphics[width=\linewidth]{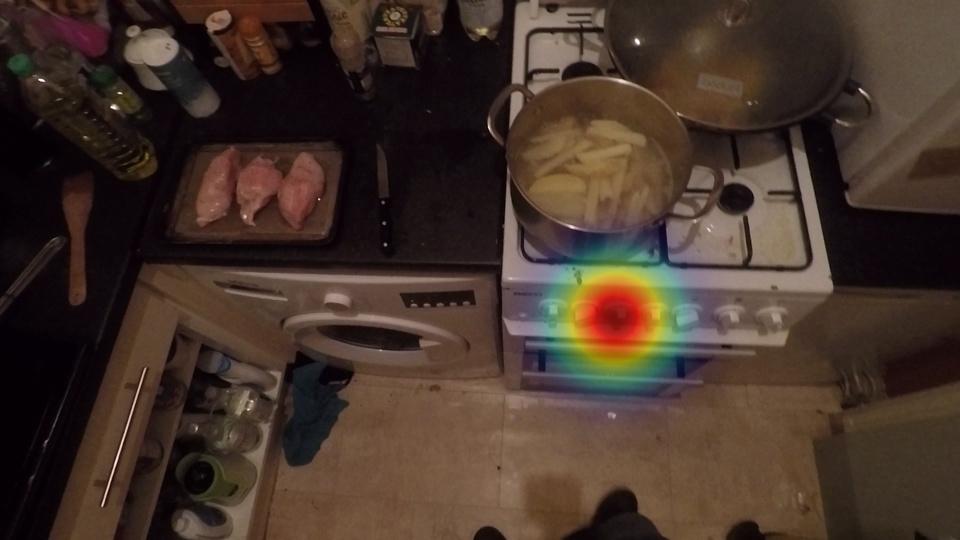}
        \caption{Ground-Truth}
    \end{subfigure}

    \vspace{0.3em}
    
    \begin{subfigure}[b]{0.32\linewidth}
        \centering
        \includegraphics[width=\linewidth]{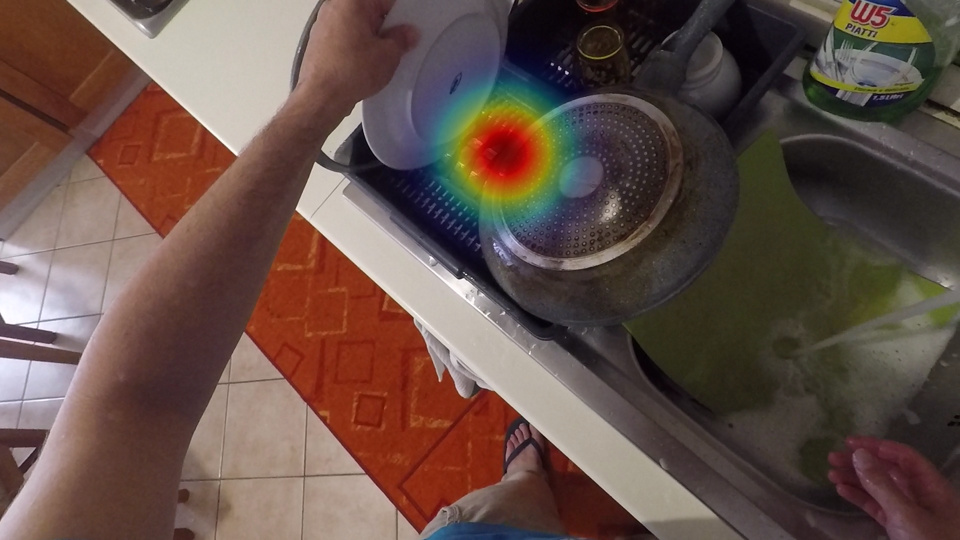}
        \caption{DINO}
    \end{subfigure}
    \hfill
    \begin{subfigure}[b]{0.32\linewidth}
        \centering
        \includegraphics[width=\linewidth]{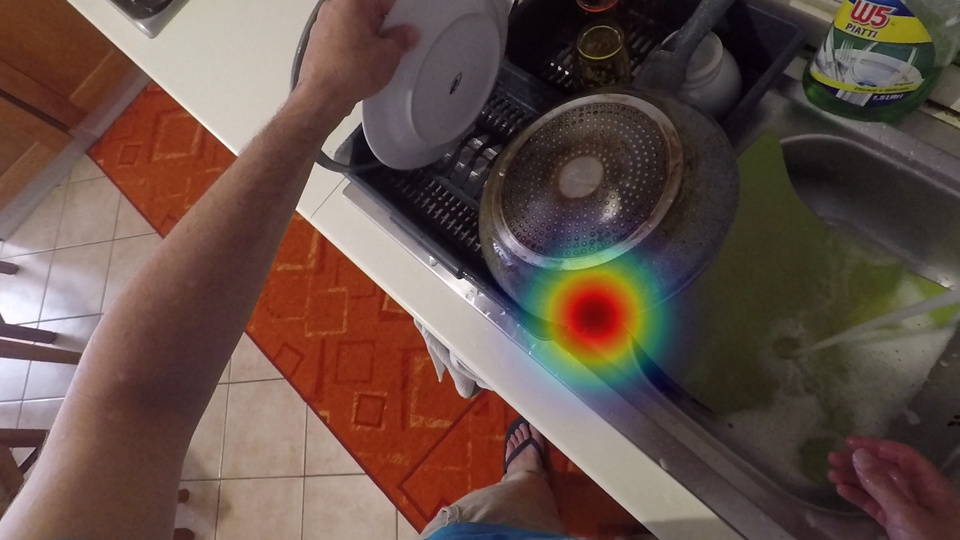}
        \caption{MAPLE}
    \end{subfigure}
    \hfill
    \begin{subfigure}[b]{0.32\linewidth}
        \centering
        \includegraphics[width=\linewidth]{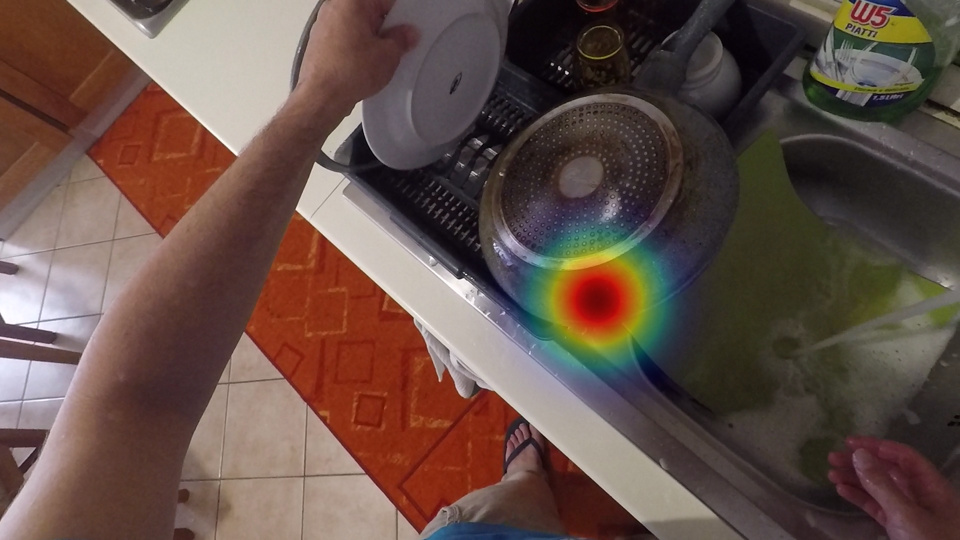}
        \caption{Ground-Truth}
    \end{subfigure}
    
    \caption{\textbf{Qualitative Comparison of Contact Predictions.} We showcase the output of cVAE contact predictor heads working with DINO vs.~MAPLE features for two samples from \cite{liu2022joint}.}
    \label{fig:contact_pred_baselines}
\end{figure}

\begin{figure}[t]
    \centering

    \begin{subfigure}[b]{0.32\linewidth}
        \centering
        \includegraphics[width=\linewidth]{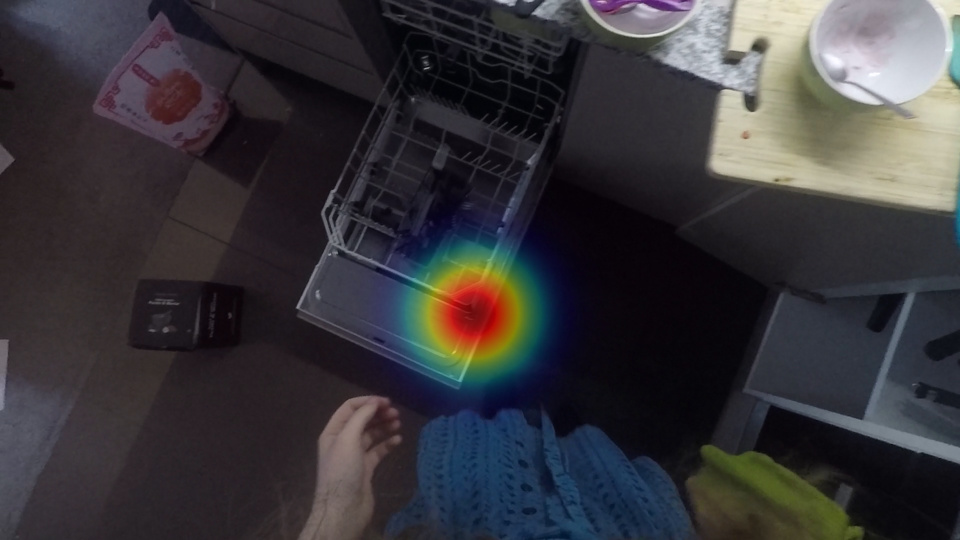}
    \end{subfigure}
    \hfill
    \begin{subfigure}[b]{0.32\linewidth}
        \centering
        \includegraphics[width=\linewidth]{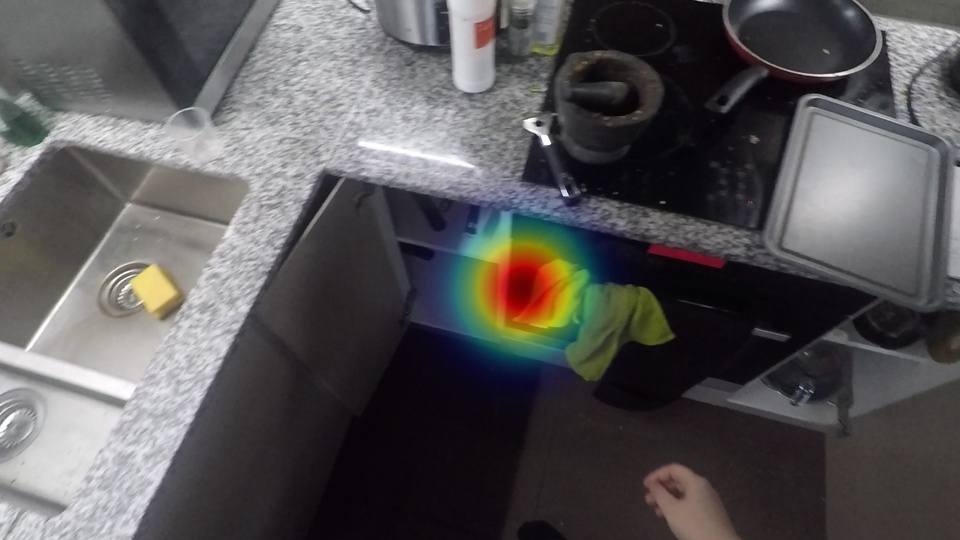}
    \end{subfigure}
    \hfill
    \begin{subfigure}[b]{0.32\linewidth}
        \centering
        \includegraphics[width=\linewidth]{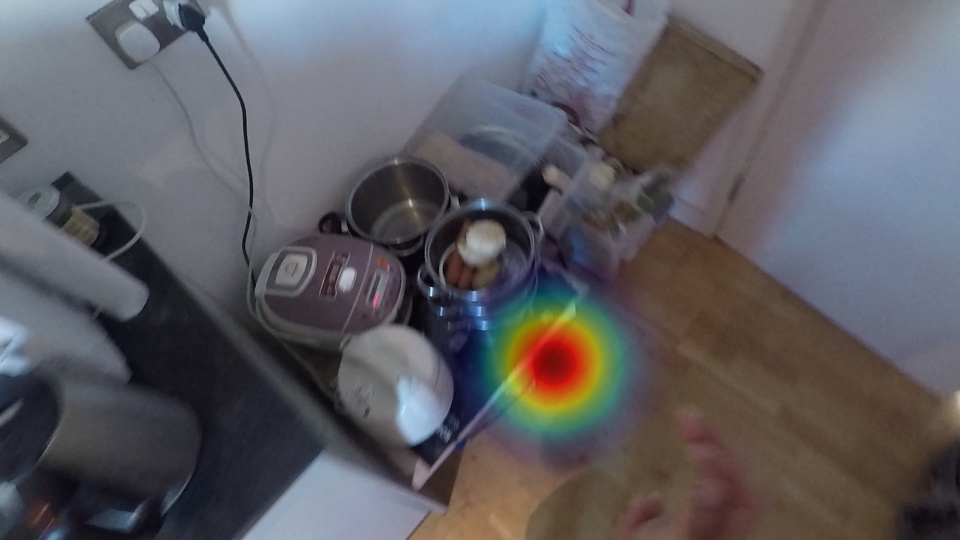}
    \end{subfigure}

    \vspace{0.3em}

    \begin{subfigure}[b]{0.32\linewidth}
        \centering
        \includegraphics[width=\linewidth]{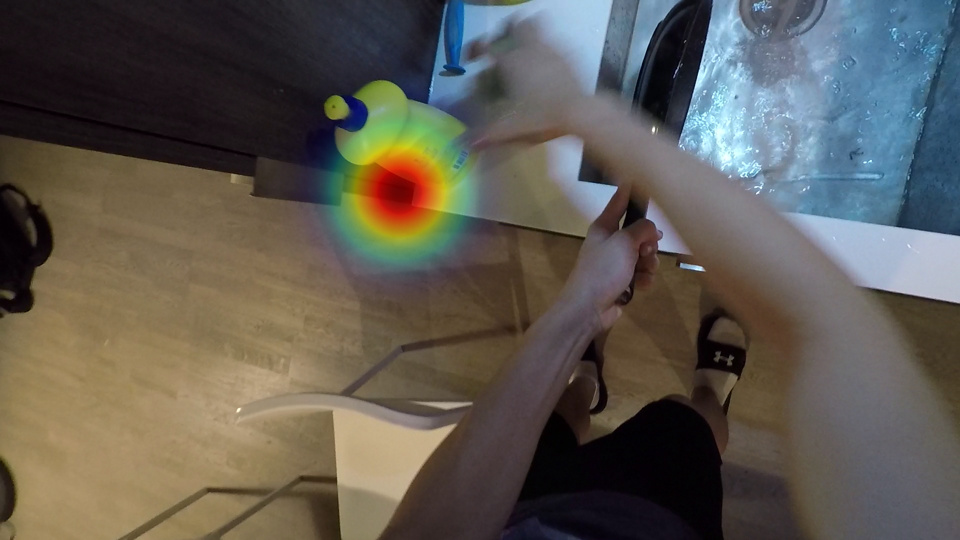}
    \end{subfigure}
    \hfill
    \begin{subfigure}[b]{0.32\linewidth}
        \centering
        \includegraphics[width=\linewidth]{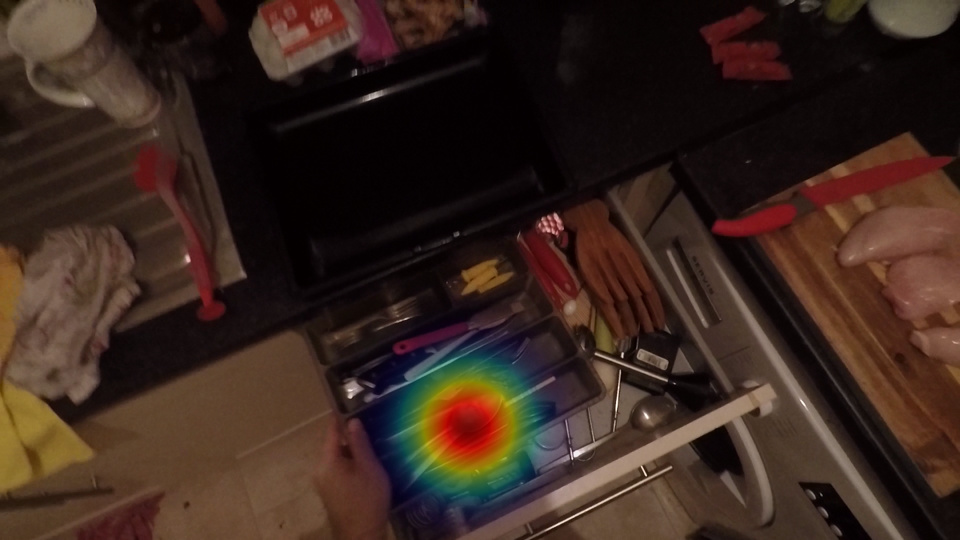}
    \end{subfigure}
    \hfill
    \begin{subfigure}[b]{0.32\linewidth}
        \centering
        \includegraphics[width=\linewidth]{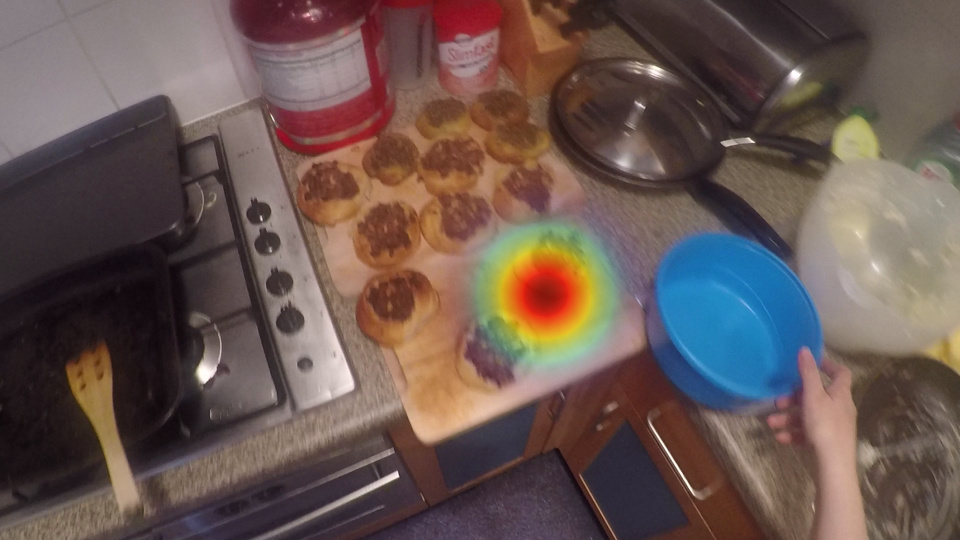}
    \end{subfigure}

    \vspace{0.3em}
    
    \begin{subfigure}[b]{0.32\linewidth}
        \centering
        \includegraphics[width=\linewidth]{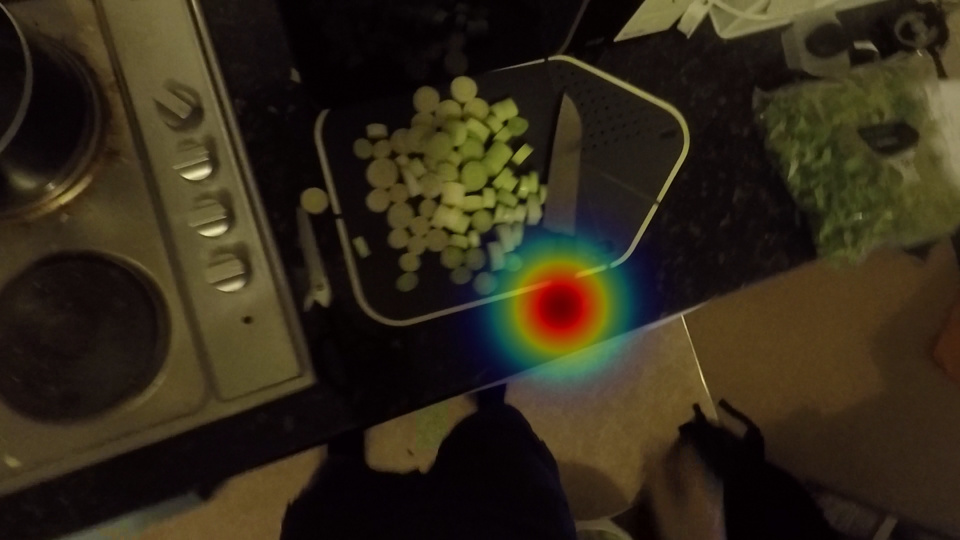}
    \end{subfigure}
    \hfill
    \begin{subfigure}[b]{0.32\linewidth}
        \centering
        \includegraphics[width=\linewidth]{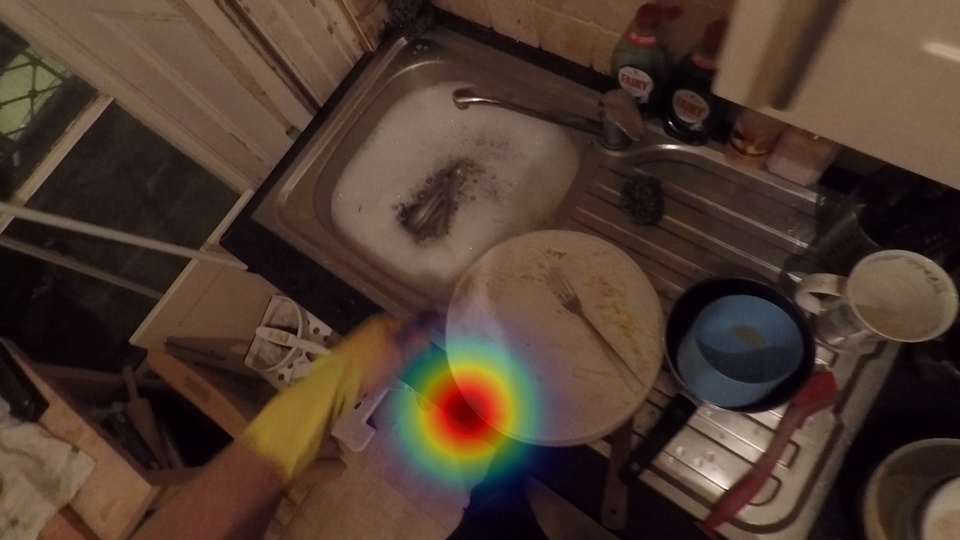}
    \end{subfigure}
    \hfill
    \begin{subfigure}[b]{0.32\linewidth}
        \centering
        \includegraphics[width=\linewidth]{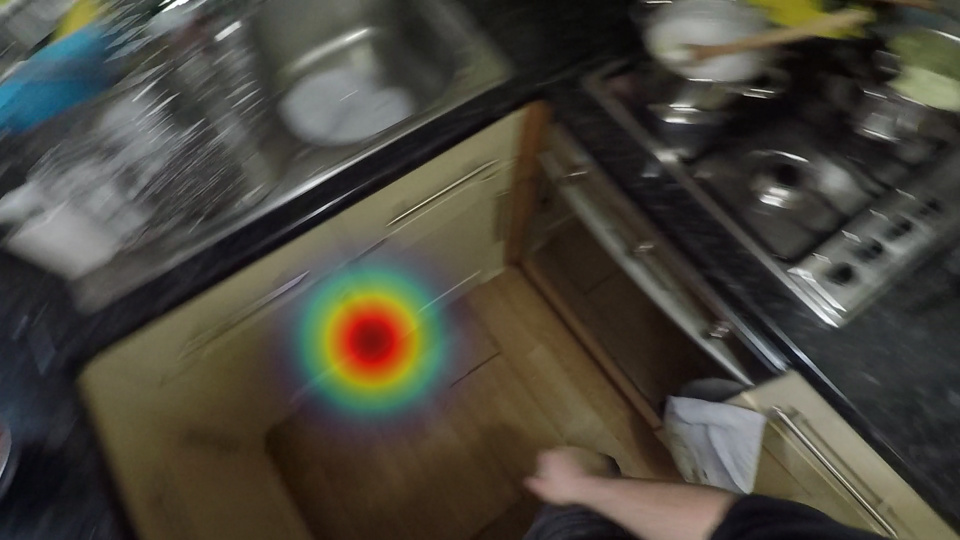}
    \end{subfigure}
    
    \caption{\textbf{Further Qualitative Examples of Contact Predictions.} Our encoder allows for high-quality contact point predictions on egocentric videos when combined with a cVAE predictor head.}
    \label{fig:contact_pred_more}
\end{figure}

We visualize predictions for two samples of our contact point predictor trained using MAPLE features vis-à-vis those of the DINO-based contact predictor in \autoref{fig:contact_pred_baselines}. Here, we only incorporate the top-scoring prediction in the heatmap for visual clarity. We further show additional high-quality predictions by the MAPLE-based contact predictor in \autoref{fig:contact_pred_more}.

\section{Training Supervision Extraction}

To obtain data for training our visual encoder, we start by processing every frame in the \texttt{clips} subset of the Ego4D dataset \citesupp{Ego4D} using the VISOR-HOS \citesupp{EKVisor} hand-object interaction segmenter, and retain frames where exactly one right hand is contacting an object with an HOS confidence at least $c \geq 0.9$, and no left hand is contacting an object with a similarly high confidence. Afterwards, we obtain frame-wise HaMeR \citesupp{HaMeR} hand reconstructions for each contact frame. These hand reconstructions are used to initialize the prediction-frame--seeking contact point tracking with the thumb and index fingertips: for each contact frame, we first perform a binary erosion operation (12 iterations) on the object mask to remove the boundary, as motivated in Section 3.2. Then, we project the thumb and index fingertip points of the right hand in the contact frame to the eroded object mask. In case the ratio of the projected points' Euclidean distance to their original distance falls outside the range between 0.3 and 1.7, the contact frame is not used. This helps eliminate objects with degenerate HOS masks. Otherwise, we initialize SAM-PT \citesupp{SamPT} with the projected fingertip points, as well as 10 other points sampled from the eroded object mask according to the query point sampling algorithm described in \citesupp{SamPT} and used to track the object. We continue tracking the points backward through the video until we encounter a frame where the HOS hand mask dilated using 75 binary dilation iterations (i.e., the expanded hand mask) no longer contains the tracked projected fingertip points. In case no such frame is encountered after 45 frames (i.e.~1.5s), the given contact frame is discarded. Otherwise, we use this frame as the prediction frame, and the backtracked points in that frame as training supervision for the contact loss. For the hand pose loss, we note that there are little constraints on the hand pose as it is moving in the direction of an object to grasp it. Hence, we always let our method reconstruct the hand pose at the contact frame, to promote learning the more informative grasping pose expected to be seen when the hand is already in contact with the object.

\begin{figure}[!ht]
    \centering
    \begin{subfigure}[t]{\linewidth}
        \centering
        \begin{subfigure}[t]{.35\linewidth}
            \centering
            \includegraphics[width=1\textwidth]{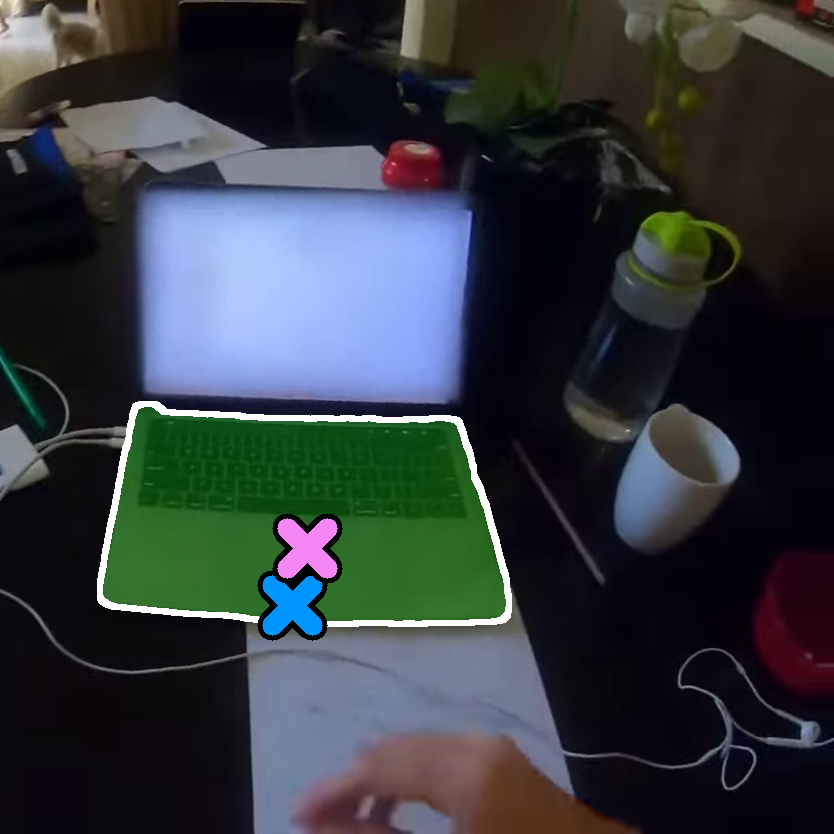}
        \end{subfigure}   
        \begin{subfigure}[t]{.35\linewidth}    
            \centering
            \includegraphics[width=1\textwidth]{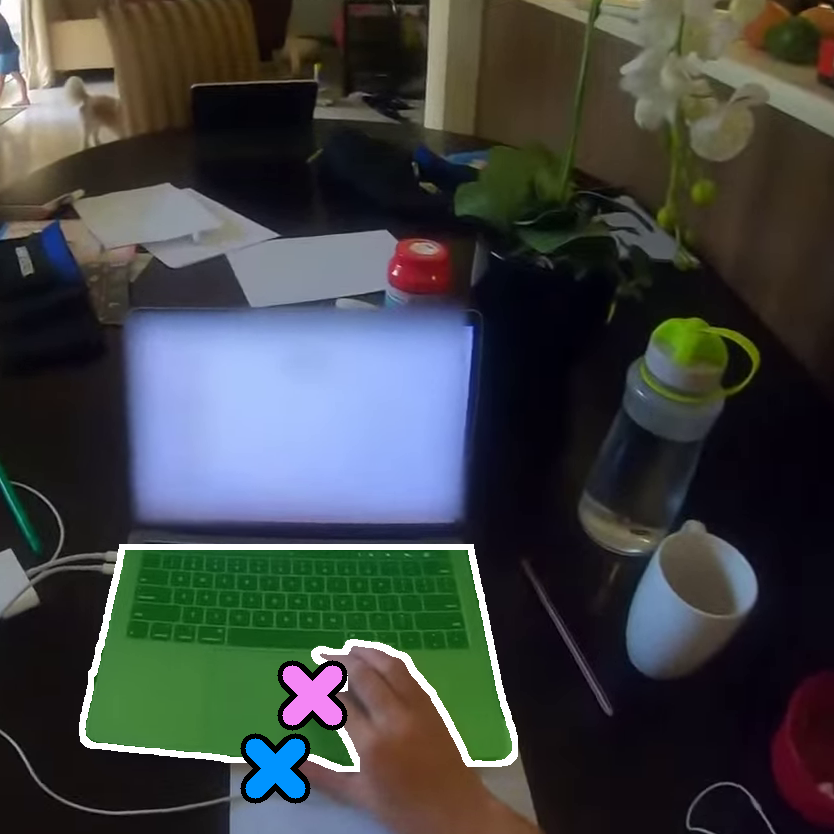} 
        \end{subfigure}  
        \caption{Laptop}
        \label{fig:sample1}
    \end{subfigure}
    
    \vspace{0.5em}
    \begin{subfigure}[t]{\linewidth}
        \centering
        \begin{subfigure}[t]{.35\linewidth}
            \centering
            \includegraphics[width=1\textwidth]{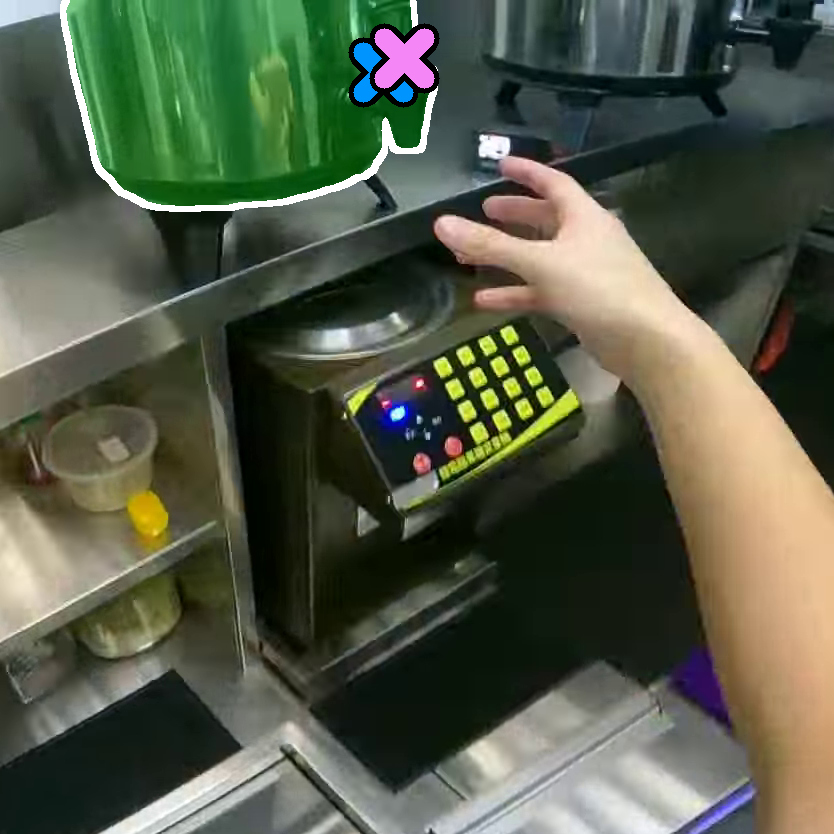}
        \end{subfigure}   
        \begin{subfigure}[t]{.35\linewidth}    
            \centering
            \includegraphics[width=1\textwidth]{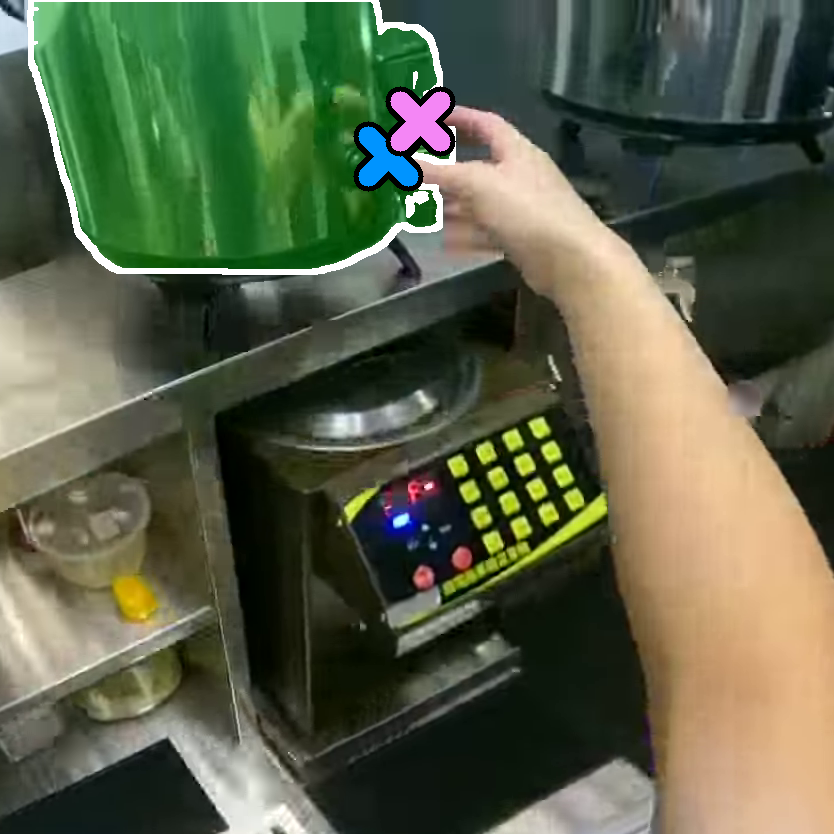} 
        \end{subfigure}  
        \caption{Oil Can}
        \label{fig:sample2}
    \end{subfigure}

    \caption{\textbf{Visualization of Object Mask and Contact Points.} The left column displays the prediction frames, while the right column shows the corresponding contact frames automatically extracted during our dataset acquisition step. The \textcolor{cyan}{cyan} cross mark represents the contact point of the thumb, and the \textcolor{pinkpurple}{pink} cross mark indicates the contact point of the index finger. }
    \label{fig:object_mask_contact_point_visualizations}
\end{figure}

Examples of prediction frames and contact point labels extracted using our pipeline are provided in \autoref{fig:object_mask_contact_point_visualizations}. 

\subsection{Error Analysis of Contact Supervision Extraction}

Although MAPLE achieves strong performance using dexterous manipulation priors learned from egocentric videos, occasionally we still notice some cases where our contact pseudo-labels are not accurate. Here, we analyze several failure modes of our automatic supervision extraction pipeline, as described in Sec.~3.2. The elimination of these errors has the potential to improve the effectiveness of the performance further and is left as a direction for future work.

\begin{figure}
    \centering
     \begin{subfigure}[t]{.35\linewidth}
     \centering
    \includegraphics[width=1\textwidth]{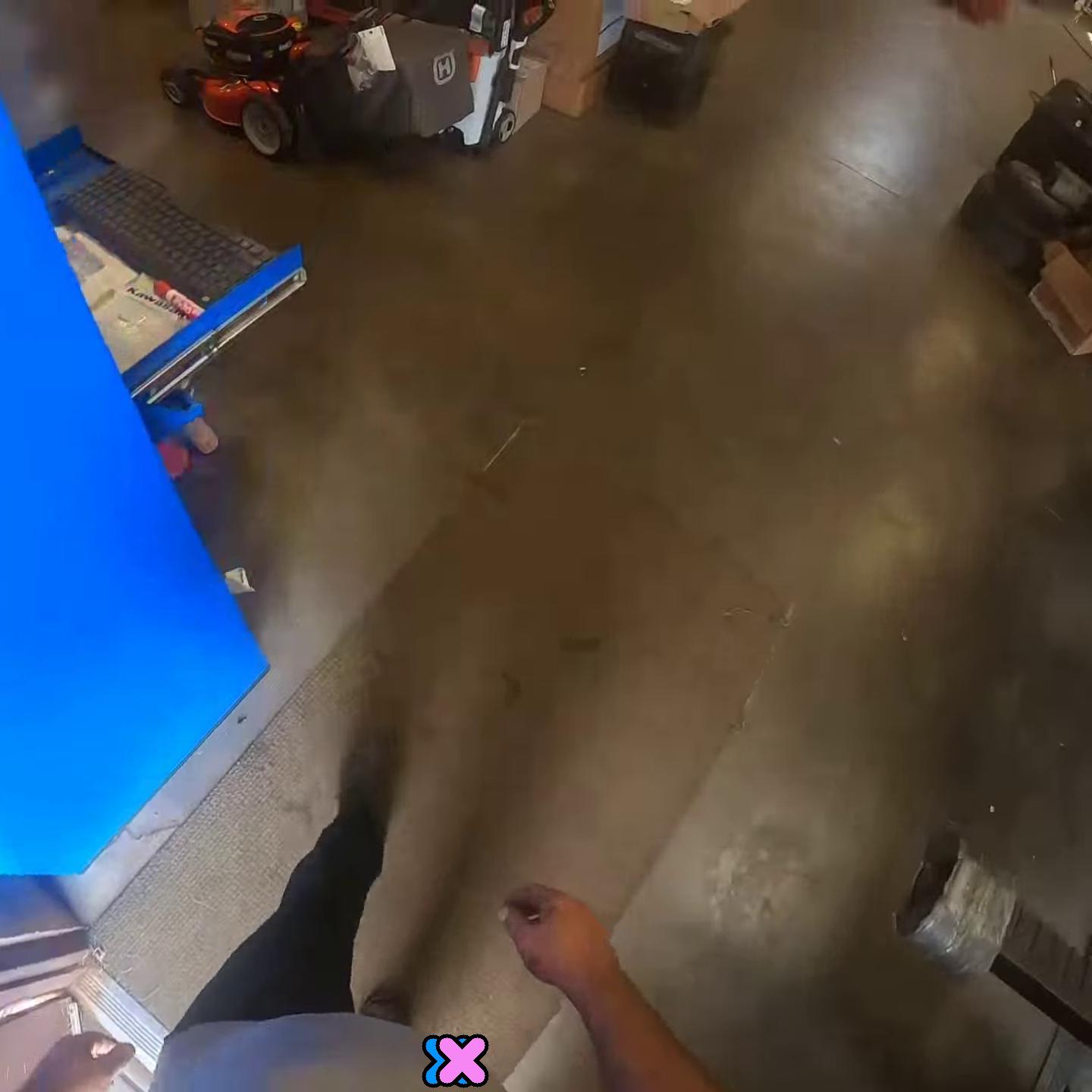}
    
    \caption{Prediction Frame}
    \end{subfigure}   
    \begin{subfigure}[t]{.35\linewidth}    
     \centering
     \includegraphics[width=1\textwidth]{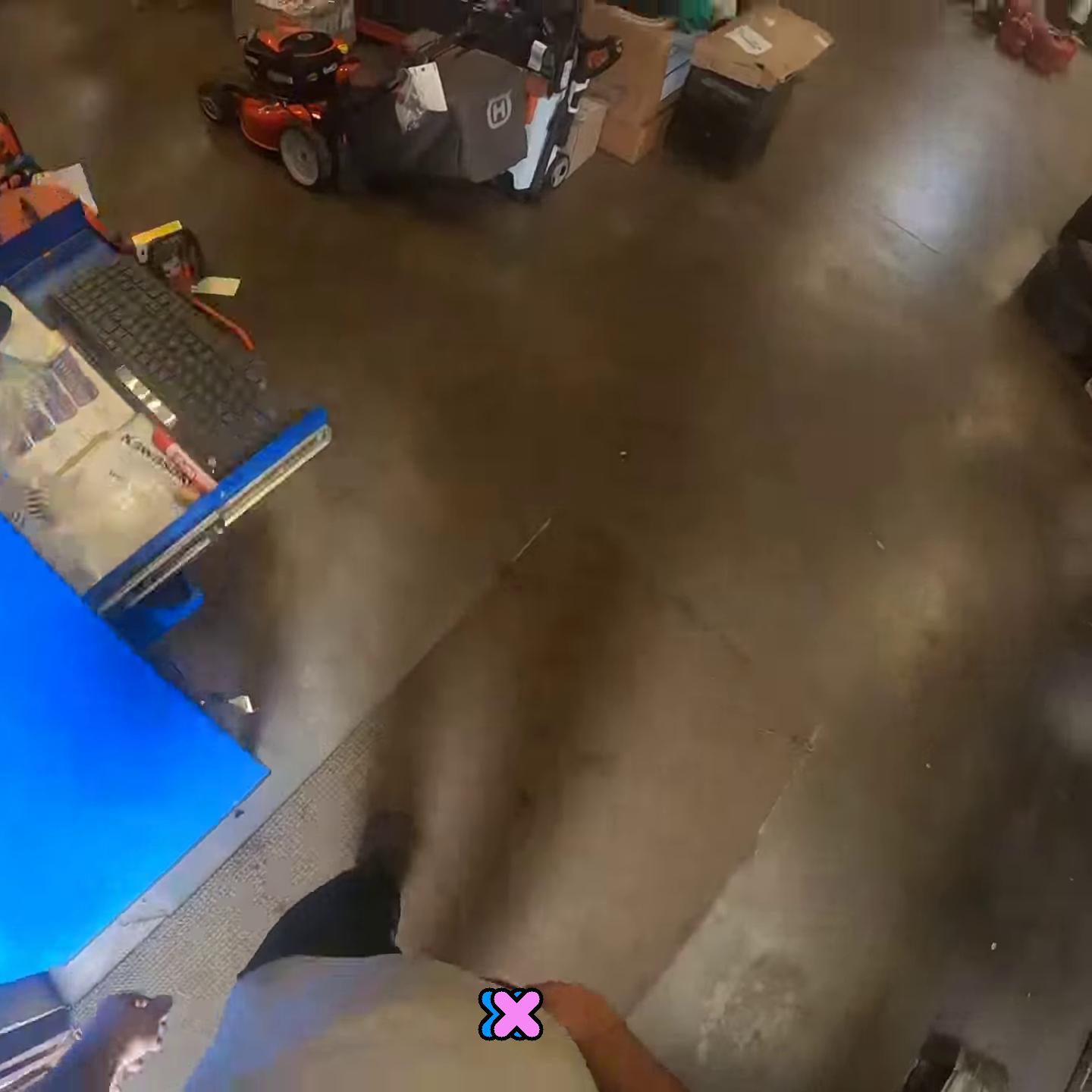} 
    \caption{Contact Frame}
    \end{subfigure}   
    \caption{\textbf{Visualization of Self-Contact Case.} The right hand is hovering above the actor's shirt in (b), which is detected as a contact frame by the HOS segmenter.}
    \label{fig:self-detection}
\end{figure}

\myparagraph{Self-Contact} The first type of categorized error is \textit{self-contact}, which occurs when the hand makes or appears to make contact with the human body or worn clothing rather than an interacted object. In such cases, since the hand still appears to establish contact with a surface in the input image, the hand-object interaction segmenter (HOS \citesupp{EKVisor}) may misclassify these frames as contact frames. An example of a self-contact error is illustrated in \autoref{fig:self-detection}. A large proportion of training data exhibiting self-contact errors may induce an undesirable bias towards human garments in the encoder's features. We hypothesize that this type of error can be tackled effectively by prompting visual grounding models \citesupp{GLIP, GroundingDINO} for masks of people in the prediction frame and eliminating samples with contact points inside such masks.

\begin{figure}[t!]
    \centering
     \begin{subfigure}[t]{.35\linewidth}
     \centering
    \includegraphics[width=1\textwidth]{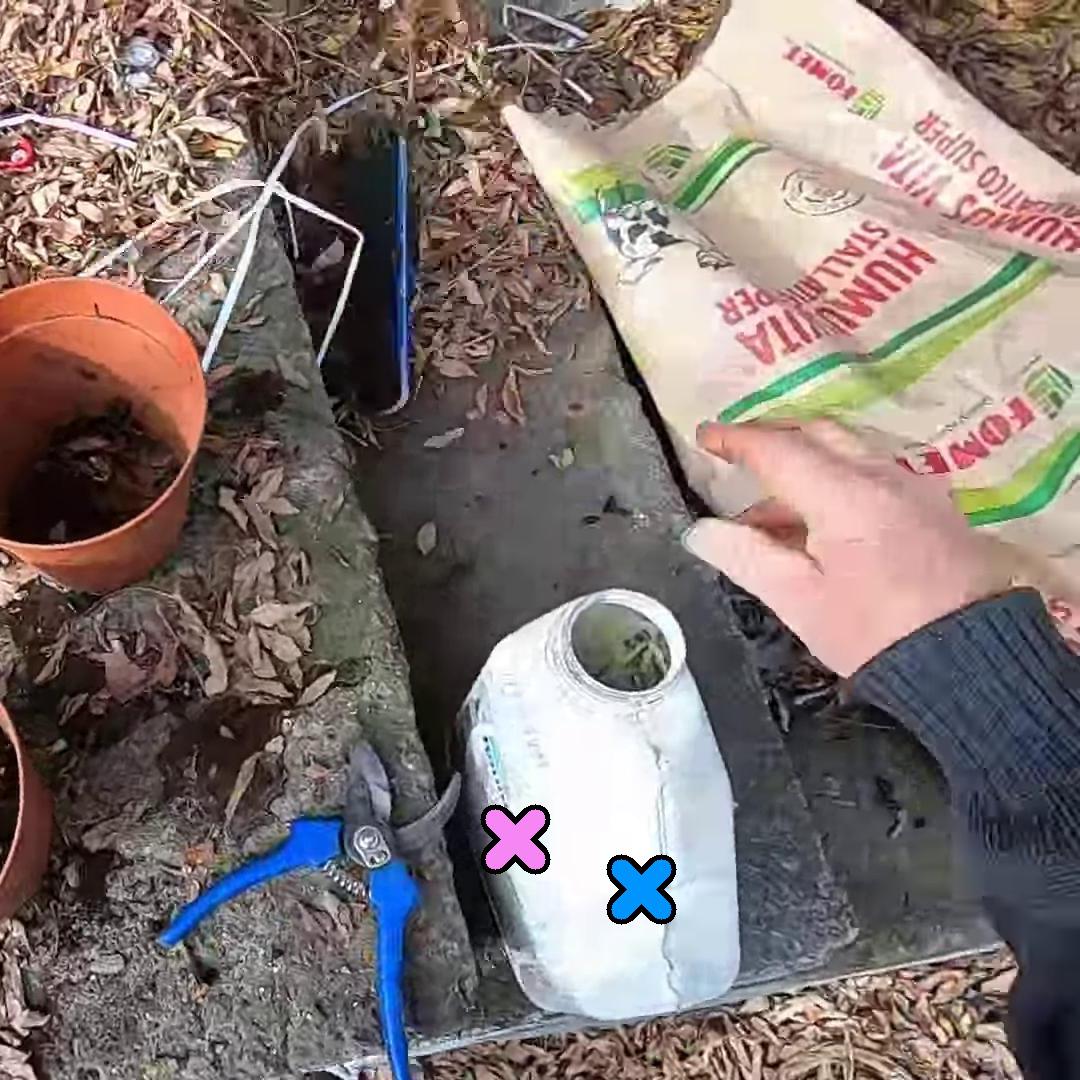}
    
    \caption{Prediction Frame}
    \end{subfigure}   
    \begin{subfigure}[t]{.35\linewidth}    
     \centering
     \includegraphics[width=1\textwidth]{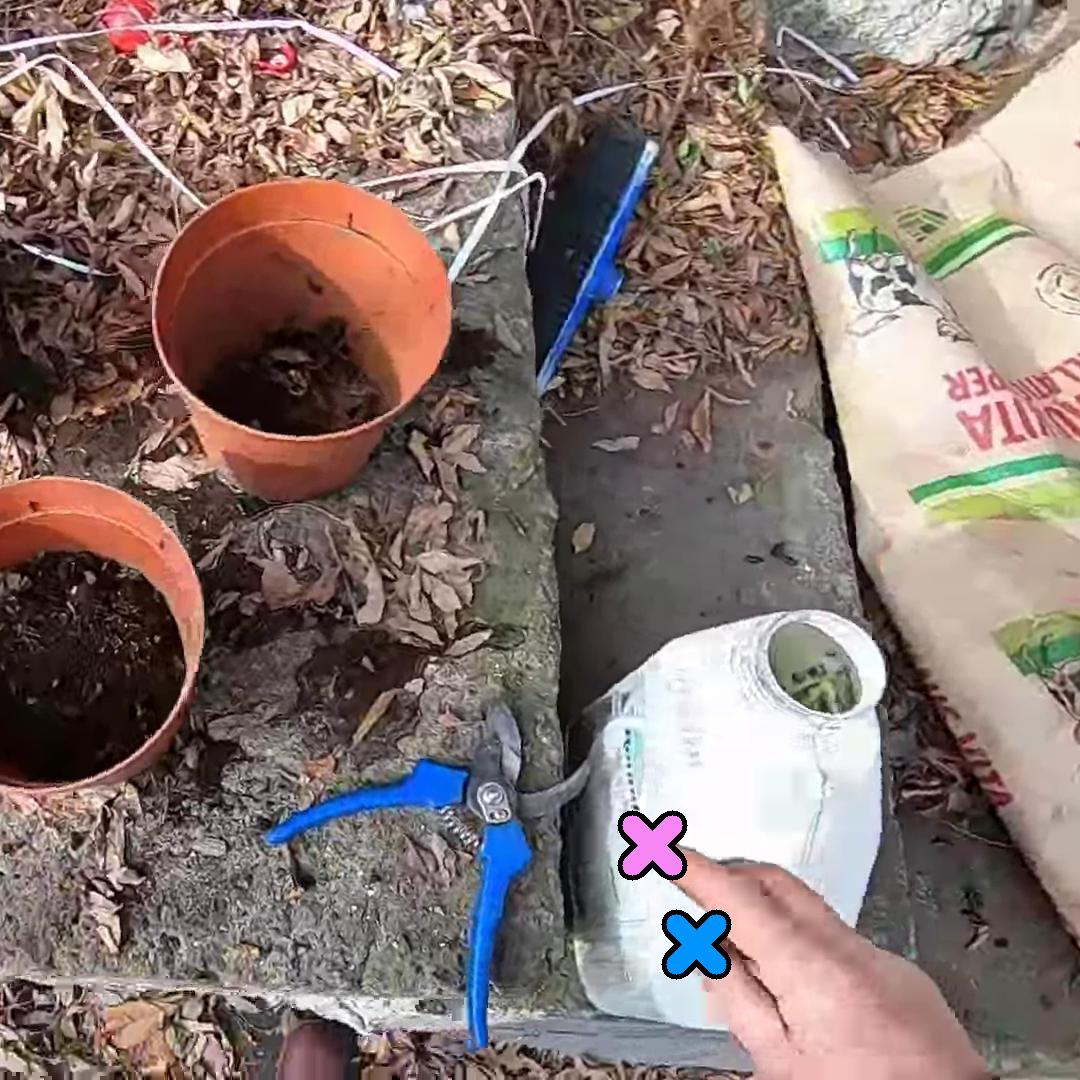} 
    \caption{Contact Frame}
    \end{subfigure}   
    \caption{\textbf{Visualization of Premature Contact Case.} The actor is reaching out for the garden scissors in (a), yet the hand is already detected as being in contact with the jug by the HOS segmenter, resulting in erroneous contact points in the prediction frame.}
    \label{fig:wrong-contact}
\end{figure}

\myparagraph{Premature Contact} Another type of error associated with contact frames occurs when the hand is still approaching an object rather than making contact, yet the HOS segmenter identifies the presence of a hand-object interaction with either a background object or a border region of the target object. We classify these instances as \textit{premature contact}. \autoref{fig:wrong-contact} illustrates an example of this error type. Eliminating premature contact errors may prove nontrivial, as we are not provided depth information in our setting, and as the human touch itself can often be fleeting, such as when pressing a button. A large proportion of training data exhibiting premature contact errors may shift the encoder's focus away from regions of interaction towards background objects and irrelevant regions of the interacted object.

\subsection{Error Analysis of Prediction Frame Search}
\label{sec:error_analysis_pred}

\begin{figure}[t!]
    \centering
     \begin{subfigure}[t]{.35\linewidth}
     \centering
    \includegraphics[width=1\textwidth]{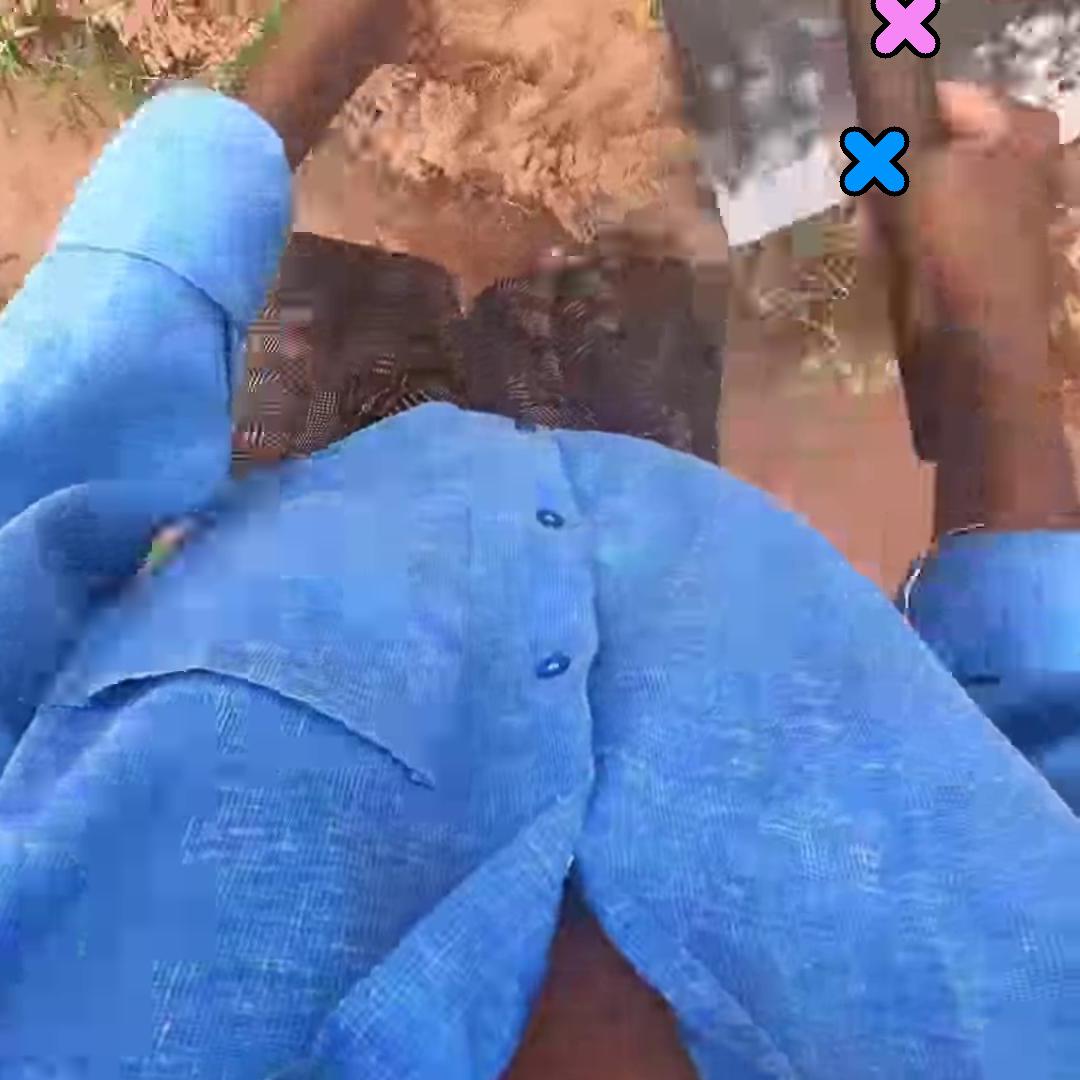}
    
    \caption{Prediction Frame}
    \end{subfigure}   
    \begin{subfigure}[t]{.35\linewidth}    
     \centering
     \includegraphics[width=1\textwidth]{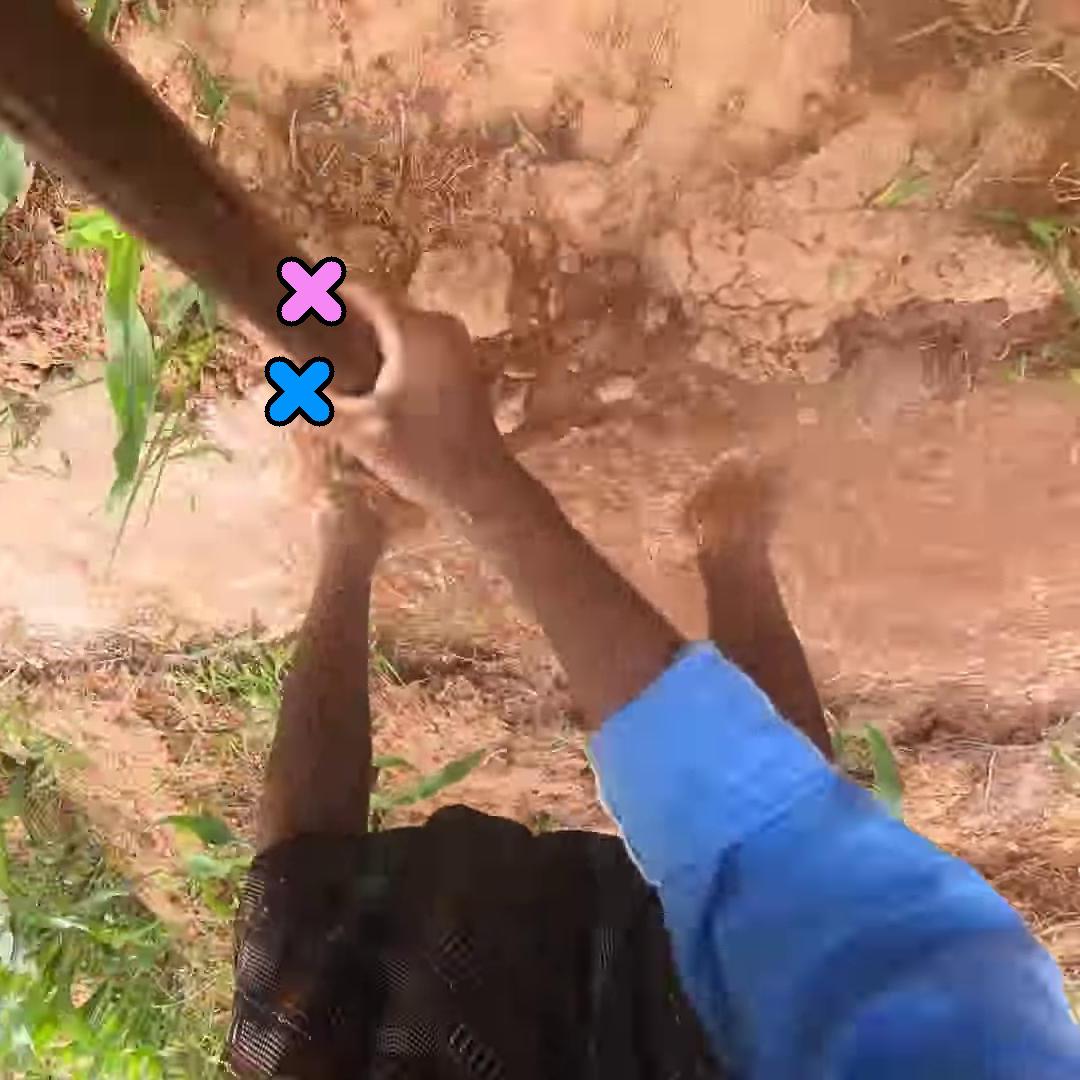}
     
    \caption{Contact Frame}
    \end{subfigure}   
    \caption{\textbf{Visualization of Premature Termination Case.} In the prediction frame, the hand is still in contact with the stick.}
    \label{fig:preliminary-termination}
\end{figure}

\myparagraph{Premature Tracking Termination} When tracking the contact points back in time from a contact frame in search for a prediction frame, the hand is sometimes not accurately detected due to a failure of the HOS segmenter, where it may be mistakenly identified as not being near the object region even though it is still contacting the object. For such samples, the prediction frame will still feature the hand interacting with the object and predicting the contact points will most frequently reduce to regressing the position of the thumb and index finger in the frame. A large proportion of training data exhibiting premature tracking termination errors may train the encoder and decoder to simply regress fingertip positions for the contact points, which will result in less informative features when a manipulator in a downstream application is presented with an image of a yet ungrasped object. An illustration of this error type is provided in \autoref{fig:preliminary-termination}.

\subsection{Tokenizer Visualizations}

\begin{figure}
    \centering
    \includegraphics[width=0.98\linewidth]{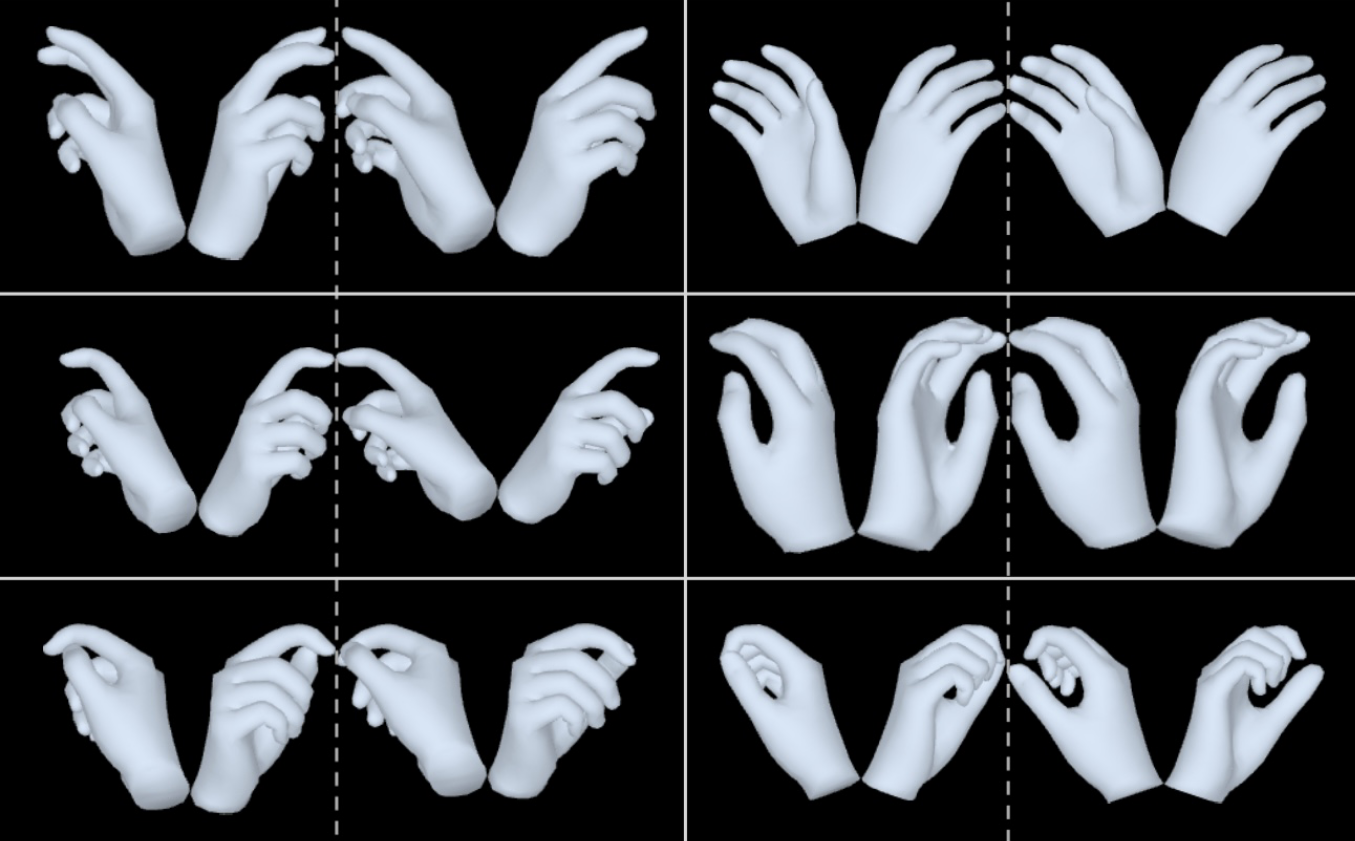}  
    \caption{\textbf{Visualization of Original and Tokenized Hand Meshes.}  The reconstructions differ only slightly from the original input, preserving the original grasp.
    }
\label{fig:tokenizer}
\end{figure}

We present examples of direct comparisons between original hand poses as regressed from the Ego4D, and the hand poses after tokenization, in \autoref{fig:tokenizer}. We find that our tokenizer learns to tokenize the hand poses well in most cases, with only small deviations observable between the original and the data reconstructed after tokenization.

\end{document}